# Chinese Discourse Annotation Reference Manual

Siyao Peng, Yang Janet Liu, Amir Zeldes
Department of Linguistics, Georgetown University
sp1184, yl879, amir.zeldes@georgetown.edu
Last updated on October 11, 2022

## Abstract

This document provides extensive guidelines and examples for Rhetorical Structure Theory (RST) annotation in Mandarin Chinese. The guideline is divided into three sections. We first introduce preprocessing steps to prepare data for RST annotation. Secondly, we discuss syntactic criteria to segment texts into Elementary Discourse Units (EDUs). Lastly, we provide examples to define and distinguish discourse relations in different genres. We hope that this reference manual can facilitate RST annotations in Chinese and accelerate the development of the RST framework across languages.

# 0 Preprocessing

Before going into the EDU segmentation and RST annotation guidelines, we illustrate how metadata, tokenization, and document structure were annotated in the corpus.

## 0.1 Gold metadata information

Gold metadata annotation assembles [the Georgetown University Multilayer (GUM) Corpus](#). We include the following metadata at the beginning of *raw/*.txt* documents:

- ***title***: the Chinese title of each document displayed in the source data. e.g., 老龄化对言语感知的影响
- ***shortTitle***: a one-word English short title for each document, *e.g., aging*
- ***type***: the genre of each document, e.g., *academic*
- ***text id***: a unique text id per document, which is the combination of the corpus name, genre, and short title of the document, e.g., *gcdt_academic_aging*
- ***author***: the author of the document, *e.g., Meijuan Ning* for academic articles or *Wikipedia, The Free Encyclopedia*
- ***dateCreated, dateModified & dateCollected***: dates when the document is first created, last modified in the source, and collected by this corpus. The dates follow YYYY-MM-DD format and XXXX-XX-XX if such information is unknown
- ***sourceURL***: the source URL where the document is retrieved, e.g., *https://www.hanspub.org/journal/PaperInformation.aspx?paperID=28037*
- ***speakerCount:*** the number of speakers in the document, e.g., *0*
- ***speakerList:*** the list of speakers in the document, e.g., *none*

## 0.2 Gold section, paragraph, and sentence split

Gold section, paragraph, and sentence splits are annotated in the *raw/*.txt* documents for future macro- versus micro- level RST analyses:
- Section and subsection breaks are marked by: *<section/>, <subsection/>, <subsub…>*
- paragraph breaks are marked by two line feeds: *\n\n*
- sentence breaks are marked by one line feed: *\n*

## 0.3 Gold tokenization

This corpus follows the tokenization guidelines for [The Segmentation Guidelines for the Penn Chinese Treebank (3.0)](#) and mirrors decisions in the [Chinese Treebank (CTB) 9.0 corpus](#).

## 0.4 Automatic dependency parsing

We use the Python stanza package for automatic dependency parsing. Instead of the default *gsdsimp* model trained on [UD_Chinese_GSDSimp](#), we convert the [Chinese Treebank 9.0](#) constituent trees to dependencies using [Stanford CoreNLP](#) and use them for training stanza. Though not natively annotated in dependencies, the CTB corpus is more consistently annotated for tokenization, POS tags and syntactic relations.

The java conversion command is the following:
*java -mx1024m -cp "*:"*
*edu.stanford.nlp.trees.international.pennchinese.ChineseGrammaticalStructure -treeFile <FILE> -basic -conllx*

# 1 EDU Segmentation

*Chinese examples are shown in italics with gold tokenization and double pipelines (||) indicating EDU boundaries,* **EDUs under discussion are highlighted in bold.**
For each EDU, we provide token-level glossing; for each example, we provide an overall translation (in quotes).

## 1.1 Segmented Units

### 1.1.1 Purpose clauses

**Purpose clauses are segmented. Most of these purpose clauses are examples of Serial Verb Constructions (SVC) in Chinese.**

1. *他　　于 1724年 前往 圣彼得堡　 || 出任 数学 教授，　　　|| 但 不 喜欢 那里 。*
   3SG.M in 1724 go-to St-Petersburg || take-office math professor || but NEG like there
   "He went to St. Petersburg in 1724 to be a professor of mathematics but didn't like it there."
   source: gcdt_bio_bernoulli

2. *于是 和 母亲 移居 到 诺丁汉郡　　　 的 世袭 领地　 || 生活 。|| [ 6 ]*
   so and mother move to Nottinghamshire DE hereditary territory || live || [6]
   "So he and his mother moved to the hereditary land of Nottinghamshire to live. || [ 6 ]"
   source: gcdt_bio_byron

### 1.1.2 Relative clauses

**Relative clauses marked by de (的) are segmented, forming a SAME-UNIT + ELAB-ATTR relation.** The relative clauses must be required to show an overt predicate structure, for example, verb+object or subject+verb.

3. *约翰 因为　　 不 能 承受　 || 和 他　 的　　 后代　　　 做 比较 的 ||*
   John because NEG can endure || with 3SG.M DE descendants make comparison DE ||
   *"羞耻"，||　　 把 丹尼尔 逐出 他　 的 家族 。*
   shame || BA Daniel expel 3SG.M DE family
   "John expelled Daniel from his clan because he could not bear the "shame" of comparing with his descendants."
   source: gcdt_bio_bernoulli

4. *2月，上议院　　　通过了 || 毁坏 机器 的 ||*
   February Upper-House pass LE || destroy machine DE ||
   *工人 必须　　判处　死刑 的 || 　　　　法案*
   worker must sentence death-penalty DE || bill
   "In February, the House of Lords passed the bill announcing that workers who destroy machines must be sentenced to death."
   source: gcdt_bio_byron

5. 在 || 他　　　还 只有 十几 岁 的 ||　　　时候 , 他　　就 发现 了 ||
   at || 3SG.M still only teens year-old DE || time, 3SG.M already discover LE ||
   ***n次***　　多项式 可以 用 根式 解 的 ||　　　充要 条件
   n-degree polynomial can use radical solve DE || necessary-sufficient condition
   "When he was only a teenager, he discovered the necessary and sufficient conditions when radicals can solve polynomials of degree n."
   source: gcdt_bio_galois

Relative clauses can also be let by 之, a more formal alternative to 的.

6. 而 它们 的 || 逝去 、|| 　　消亡 之 || 　处　　正　　　　　　是 ||
   And 3PL.IN DE || passing-away || dying-out DE || place exactly COP ||
   生出　　它们　　的 || 本根 之 道 。
   give-birth　3PL.IN DE || *root DE way*
   "And the place of their passing away and dying out is the fundamental way of giving birth to them."
   source: gcdt_academic_taoist

In rare cases, relative clauses can omit the overt DE. In these situations, we ensure the pre-nominal modifier is clausal (by running 了-insertion, 着-insertion, or adverb insertion tests) and segment these DE-less relative clauses.

7. 进入 仙桃 || 　人员　必须 进行　　　体温　　　检测
   enter Xiantao || people must conduct temperature check
   "People that enter Xiantao must undergo temperature checks."
   source: gcdt_news_hubei

8. *24日 22时 , (潜江市)　　　　关闭 潜江站　　　　　　　　||*
   24th 22:00, (Qianjiang-city)　close Qianjiang-station　　　||
   离开 潜江　　|| 通道 。
   leave Qianjiang || pathway
   "They will close Qianjiang-station's pathway to leave Qianjiang. "
   source: gcdt_news_hubei

*9.* 尽管　　有　　一些 对于 || 含有　　　致命 病毒 || 疫苗 安全性 的 抨击
Although EXIST some toward || **contain deadly viruses || vaccine** safety DE criticism
"Although there are some criticisms towards vaccines that contain a deadly virus."
source: gcdt_academic_rabies

However, 的(DE)-marked clauses can directly function as complement clauses. In these situations, they are not segmented.
In the following example, "非 标记性 主位 传达 的" is the subject of "是 旧 信息." Thus, they belong to the same EDU.

*10.* 与 主语　　　重合　的 || 非 标记性　　　主位 传达　　　的 是　旧 信息 , **||**
with subject coincide DE || non- marking theme convey DE  COP old information , ||
而 述位　　　传达 的 是　　新 信息 。　　　　**||**
but rheme convey DE COP new information　　　　||
"those conveyed by the non-marking theme that coincides with the subject are old information, but those conveyed by rheme are new information."
source: gcdt_academic_iconicity

*11.* 对方　　　　在 这 段　　　对话 中　　得到 的 只 有 乐趣 和 充满 趣味 回忆 ,
the-other-side at this CL conversation middle get DE only have fun and full joy memory
"those the other side gets in this conversation are only fun and joyful memories, "
source: gcdt_whow_flirt

## 1.1.3 Manners and Means

Manner and means adverbial clauses usually occur in the middle of a sentence. Here are some examples:

*12.* 而且 他 试图 || 用 这 一 方式 ||　　解释 波义耳 定律
and 3SG.M try **||** use this one method || explain Boyle's law
"He tries to use this method to explain Boyle's law."
source: gcdt_bio_bernoulli
Note that 用 is equivalent to 利用 in this example and is used as a verb.

*13.* 往往 由 公安　机关 ||　　　以　　寻衅滋事　　　　为 由 ||
often by security department || taking trouble-provoking as reason ||
处以 行政 拘留
sentence admininistrative detention
"(they) were often sentenced as administrative detention because of trouble-providing by the security departments."
source: gcdt_academic_supervision

*14.* 有　　一 种 业余　　　玩法　　　是 将 边上 的 球 ||
EXIST one CL amateur method COP BA edge DE ball ||
按照　　　　　半色 、实色 、半色 、实色　　　　　的 顺序 || 摆放 。
according-to　half-color , solid-color, half-color , solid-color DE order || place
"An amateur way of playing is to place the balls on the side in the order of half color,
solid color, half color, solid color."
source: gcdt_whow_pool

*15.* 然后　　　再 || 看 情况 ||　　采取 进一步 行动 。
after-that then ||see situation || take further action
"Then take further action based on the situation."
source: gcdt_news_tiktok

***16.*** 读 起来 || 挺 拗口　　 的
read start || very mouthful DE
"It's a mouthful when you read it."
source: gcdt_whow_glowstick

## 1.1.4 Reported speeches and cognitive predicates

**Reported speeches and cognitive predicates suggest segmentation of the complement
when two conditions are simultaneously met.**
- (a) The main predicate belongs to one of the following categories:
    - saying verbs;
    - cognitive verbs (feelings, thoughts, hopes);
    - perception verbs (see, feel, hear, sense).
- (b) The complement is by itself an entire clause

The following verbs are attested to introduce reported speeches in Chinese:
- 说 say
- 声称 claim
- 要求 request
- 宣称 claim
- 宣布 announce
- 建议 suggest
- 询问 inquire
- 提到 mention
- 称 claim
- 描述 describe
- 否认 deny
- 显示 indicate
- 主张 assert
- 建议 suggest
- 规定 prescribe

- 讨论 discuss
- 综述 sum up
- 提出 put forward
- 标志 signify
- 表示 express
- 明确 clarify
- 强调 emphasize
- 指责 accuse
- 确定 affirm
- 说明 explain
- 认为 think
- 以为 think
- 想像 imagine
- 相信 believe
- 知道 know
- 懂得 understand
- 明白 understand
- 推荐 recommend
- 写 着 (that) writes
- 感觉 feel
- 想 着 think
- 约定 agree
- 觉得 feel
- 希望 hope
- 表明 indicate
- 意味 imply
- 提示 suggest
- 倡导 advocate
- 期望 expect
- 赞成 approve
- 计划 plan
- 打算 plan
- 决定 decide
- 盼 hope
- 鼓励 encourage
- 透露 reveal
- 探讨 discuss
- 看到 see
- 见到 see
- 想到 think of
- 思考 think
- 发现 discover
- 考虑 consider
- 记住 remember

- 记得 remember
- 谨记 remember
- 评价 comment

For reference, the following verbs are included as reported speech and cognitive verbs in English RST-DT (Carlson & Marcu 2003):
- say, tell, state, announce, declare, suggest, advise, report, indicate, point out, explain, ask
- think, believe, know, imagine, suppose, conjecture, wish, hope, predict, fear, estimate, calculate, anticipate, expect, dream
- see, feel, hear, sense

Here are some examples from GCDT:

*17.* 他 自己 说 ：‖" 在 应用文　方面， 英文 、德文 、法文 没有　 问题 。
3SG.M self say : ‖ " in formal-writing aspect, English, German, French NEG-have problem .
"He said: as for formal writing, there is no problem with English, German, and French.'"
source: gcdt_bio_chao

*18.* 即　 援引 他 棺材 上 的　 银盘 刻印 ，‖
that-is cite 3SG.M coffin on-top-of DE silver-place engraving ‖
认为 ‖ 他　 是 " 年 约　 65 岁 。 "
think ‖ 3SG.M COP " age about 65 years-old . "
"That is, citing the inscription on the silver plate on his coffin and believing he was 'about 65 years old.'"
source: gcdt_bio_emperor

Moreover, the subject of the reported speech can be implied or inherited from the previous context.

*19.* 至少　 让 其他 人　 知道 ‖ 你 要　 去　 哪儿 ，
at-least let other people know ‖ 2SG want go-to where
"At least (you should) let other people know where you're going,"
source: gcdt_whow_hiking

Such attribution of reported speech can as well be negative.

*20.* 但是，如果　 不　 知道 ‖ 怎幺　 正确　 地 驱逐 老鼠
But if NEG know ‖ How-to correct ADV expel mice
"However, if you don't know how to expel mice"
source: gcdt_whow_mice

One sentence can also contain multiple occurrences of combinations of speech verb + content. In the following example, 提到 (mention), 称 (state), and 希望 (hope) introduce new EDUs. However, the EDU separation after 说 (say) is due to the relative clause after instead of a complete clausal complement of saying.

21. 他　　　提到 || 较早前 接受　　电视台 节目 采访　　　　时 || 　说 ||
    3SG.M mentioned || earlier receiving TV　　　show interview at-time-of || say ||
    "二元　优惠　计划 可能 要　　调高　金额　至 三元 "　　的 || 讲法 ， ||
    " 2-yuan discount plan may have-to increase amount to 3-yuan "　DE|| statement ||
    称 ||　这　不　是　　政府　　的 立场 ， ||
    state || This NEG COP government　DE position, ||
    希望 || 不　　要 引起 一些　不 必要　　的 误会 。||
    hope || NEG will cause some NEG necessary DE misunderstandings.
    "In an earlier interview with a TV program, he said that 'the two-yuan discount plan may have to increase the amount to three yuan;' and stated that this is not the government's position and hoped that won't cause some unnecessary misunderstandings."
    source: gcdt_news_unemployment

Here is a counter-example where 宣称 (claim) does not introduce a new EDU because the following portion is not a clause but a fixed expression 宣称 A 为 B (claim A to be B).

22. 卡美哈梅哈 五世 宣称　　　诺顿 一世 为　"全 美国　　唯一 的 统治者 "。
    Kamehameha V declared　　Norton I　to-be "all America only　DE ruler".
    "Kamehameha V declared Norton I 'the sole ruler of all America.'"
    source: gcdt_bio_emperor

## 1.1.5 Coordinations

**Coordinated copula clauses are separated.**

23. 他　　是 欧拉　　的 同 时代 人 ， ||　也　　是 密友 。
    3SG.M COP Euler DE same era people , || also COP close-friend
    "He was Euler's contemporary, and a close friend."
    source: gcdt_bio_bernoulli

**Subordinated coordinating conjunctions are also separated.**

24. 每天　忙着　为　　希腊 军队 筹集 物资 ， ||　　购买　先进　武器 ， ||
    everyday be-busy for Greek army raise supply , || purchase modern weapon , ||
    调节 内部 纠纷
    resolve internal conflict
    "Busy daily raising supplies for the Greek army, buying advanced weapons, mediating internal disputes."
    source: gcdt_bio_byron

## 1.1.6 Predicative adjectives

**Predicative adjectives in Chinese do not require overt copula and can be segmented from other clauses.**

**25.** 拜伦　先天性 的 跛足，‖　　而 他 的　　　母亲 性情 乖戾 、　　喜怒 无常
Byron congenital DE lame, ‖ but 3SG.M DE mother temper grumpy, happy-sad unstable
"Byron was born lame, and his mother was surly and moody."
source: gcdt_bio_byron

Similar to English, when multiple predictive adjectives are conjoined with the same subject, they jointly form one EDU. **We DO NOT segment these coordinated predicative adjectives.**

26. 我 知道 ‖ 这 周　　　你 很　辛苦　、很　不 容易
1SG　know ‖ this week 2SG very hard-working 、 very NEG easy
"I know that you have been working hard and not easy this week."
source: gcdt_whow_procrastinating

However, exceptions apply when a predicative adjective is conjoined by a strong discourse marker, for example, 而且 (but also) in the following example.

**27.** 由于 证据 含糊不清 、矛盾　　‖　　而且 寥寥无几
since evidence ambiguous , contradictory ,‖ but-also rare
"Because the evidence is vague, contradictory, but also scant."
source: gcdt_bio_byron

**28.** 虽然 安全 ，‖ 但　　不 方便
although safe ‖ but NEG convenient
"Although safe, but inconvenient."
source: gcdt_interview_wimax

Moreover, when some conjoined predicative adjectives take PP or NP complements, they are separated from other adjectives and form their EDU.

**29.** 突发　疫情　　　是　　指　　突发 的 、　　　　　群发 的 、‖
sudden epidemic COP refer-to  suddenly-happened DE　grouply-happened DE ‖
对　　　公共 健康　或 经济 、 政治 、 社会　　等 影响　　　大 的 ‖
toward public health or  economy politics society　etc. influence big DE ‖
( 已　　造成 危害 ‖ 或 可能 造成 危害 ) ，‖
( already cause harm ‖ or may cause harm ) , ‖
需要 紧急　　采取 控制 措施，‖ 与 传染病　　　　有关 的 ‖
need urgently take control measure ‖ to infectious-disease related DE ‖
公共 卫生　　事件
public health event

"A sudden outbreak refers to a sudden, mass public health event that has a great impact – has caused harm or may cause harm – on public health or the economy, politics, and society; and that needs urgent control measures and relevant to infectious diseases."
source: gcdt_academic_governance

## 1.1.7 Nominal predication

Nominal predicate structures can occur in Chinese without an overt copula verb.
The following example states that the area of China is 9.6M km^2 without a copula 是.

> ***30.*** 中国 国土 面积 ***960***万 平方 公里
> China land area 9.6M squared kilo
> "China's land area is 9.6 million square kilometers"

Here is an example from GCDT where no overt copula occurs between 原名 (original-name) and 樋口奈津 或 樋口夏子 (Higuchi-Najin or Higuchi-Natsuko):

> ***31.*** 樋口一叶 || （ *1872*年 *5*月 *2*日 － *1896*年 *11*月 *23*日 ）, ||
> Higuchi Ichiyo || ( 1872 May 2 - 1896 November 23 ) , ||
> 生于 东京 , ||原名 樋口奈津 或 樋口夏子 , ||
> born-in Tokyo, || original-name Higuchi-Najin or Higuchi-Natsuko, ||
> 是 日本 明治 初期 主要 的 女性 小说家 。
> COP Japan Meiji early-period leading DE female novelist .
> "Higuchi Ichiyo (May 2, 1872 - November 23, 1896), born-in Tokyo, formerly known as Higuchi Najin or Higuchi Natsuko, was Japan's leading female novelist in the early Meiji period."
> source: gcdt_bio_higuchi

## 1.1.8 Parentheticals and references

**Parentheses are separated,** including round "(" ")", square "[" "]" and curly "{" "}" brackets. However, "《" "》" mark book titles in Chinese and does not create EDU boundaries.

> ***32.*** 约翰 还 曾 试图 盗窃 丹尼尔 的 著作 《 ***Hydrodynamica***》**||**（ 流体 力学 ）**||**
> John also once try steal Daniel DE piece Hydrodynamica || ( fluid mechanics ) ||
> 并 把 它 重新 命名 为《 ***Hydraulica***》。
> also BA 3SG.IN anew name to-be Hydraulica .
> "John also tried to steal Daniel's book Hydrodynamica (Fluid Mechanics) and renamed it Hydraulica."
> source: gcdt_bio_bernoulli

Following RST-DT and GUM guidelines, **supporting references are separated from the contents.**

33. *希腊　政府　　　　为 拜伦 举行　　　　了 隆重 的 国葬　　　　仪式 。||*
    Greek government　for Byron take-place LE grand DE state-funeral ceremony . ||
    *[ 1 ] [ 2 ]*
    [ 1 ] [ 2 ]
    "The Greek government held a grand state funeral for Byron. [ 1 ] [ 2 ]"
    source: gcdt_bio_byron

However, **exceptions apply when square brackets denote International Phonetic Alphabet (IPA) or when the brackets denote mathematical equations.**

34. *参加 了 || 需要　　　识别 ||*
    attend LE ||need-to identify ||
    *音节　　[ ba ] 或 [ pa ] 和 [ ba ] 、[ da ] 或 [ ga ] 合成 的 ||*
    syllable [ ba ] or [ pa ] and [ ba ] , [ da ] or [ ga ] synthesize DE ||
    *连续续 的 || 实验 。*
    continuum DE || experiment
    "Participated in an experiment that required the identification of the continuum of syllables [ ba ] or [ pa ] and [ ba ] , [ da ] or [ ga ] ."
    source: gcdt_academic_aging

35. *当 且 仅 当 $p=2^{2^{k}}+1$*
    if and only if　$p=2^{2^{k}}+1$
    "if and only if $p=2^{2^{k}}+1$."
    source: gcdt_bio_galois

36. *你 将 获得 $2^{(X-1)}$ 元 。*
    2SG will win $2^{(X-1)}$ yuan 。
    "You will get $2^{(X-1)}$ dollars."
    source: gcdt_bio_bernoulli

In addition, **inserted core arguments are not separated, whereas optional modifiers are** in the following examples.

37. *在此之前 , （ 我们 ）都 密集　地 和 秘书长　　　与*
    prior-to-this ,, ( 1PL ) all intensive ADV with secretary-general and
    *副秘书长　　　　　进行 联系 ,　||*
    deputy-secretary-general　conduct connect , ||
    "Prior to this, ( we ) were in intensive contact with the secretary general and the deputy secretary general,"
    source: gcdt_interview_cycle

38. 基本上　是　必须　要　　与　||
basically COP must　must-be with ||
（自由车）||协会　　　　进行　　多　　次 协调　　　　的
( bicycle )　|| association　　conduct multiple CL coordination　　DE
"Basically we must conduct multiple coordinations with the bicycle association."
source: gcdt_interview_cycle

39. 我们 ||　（德懋 国际 ）||非常 荣幸，||
1PL　||（Demao International. ) || very honored , ||
能　　赞助 这 次 的　　　　环台赛 。
can sponsor this CL DE Ring-Taiwan-Tour.
"We( Demao International ) are very honored to sponsor this Tour of Taiwan ."
source: gcdt_interview_cycle

Moreover, Note that **parenthetical dates in article citations are not EDUs, but parenthetical dates describing dated events, birth years, etc. are EDUs:**

In English, we see the following:
[We read Smith (2000)]
[Jane Smith] [(1901-1974)] [was a paleontologist]

Similarly in Chinese:

40. 另外　　　樋口　　的 一些 作品，林文月　　　　翻译 并 发表 至
in-addition Higuchi de some works , Lin-Wenyue translate and publish in
《联合 文学 》杂志　中，　如　　《比肩 》（*1998年 1月* ）、
 " United Literature" magazines in , for-example　" Bijian " ( January 1998 ) ,
《浊江 》（*1998年 9月* ）。
" Zhuojiang " ( September 1998 ) .
"In addition, some of Higuchi's works were translated and published by Lin Wenyue in "United Literature" magazines, such as 'Bijian' (January 1998) and 'Zhuojiang' (September 1998)."
source: gcdt_bio_higuchi

## 1.1.9 Dashes, hyphens, and colons

Like RST-DT, when dashes and multi-hyphens introduce parenthetical information or subtitles, we break the sentence and include the dashes and hyphens in the embedded EDU.

41. 德沃夏克 在 纽约　　遇到 了　　　他 后来　 的　 学生 哈里·布雷 ||
    Dvorak  at New-York  met PERF 3SG.M future DE student Harry-Bray ||
    —— 最 早　 的 美国　　黑人 作曲家 之一 。
    -- most early DE American black composers one-of .
    "Dvorak met his future student Harry Bray in New York – one of the first African-American composers."
    source: gcdt_bio_dvorak

Single hyphens commonly denote a combined meaning between words and thus do not create new EDUs.

42. 菱形　　　　　　球框 里 球 的　　　摆放 方式 是　　　按照
    rhombus-shaped bracket inside ball DE placement way COP according to
    1 - 2 - 3 - 2 - 1 的 顺序 来 的 。
    1 - 2 - 3 - 2 - 1 DE order place DE .
    "The balls are placed in the rhombus in the order 1 - 2 - 3 - 2 - 1."
    source: gcdt_whow_pool

43. 1933年 至 1936年 年间 ，|| 横跨　　　旧金山湾 的 || 旧金山 - 奥克兰 海湾 大桥 ||
    1933 to 1936 between-years , ||span-across SF-bay DE || SF - Oakland bay bridge ||
    （ 又 称　　海湾 大桥 ）|| 建成 。
    ( also name bay bridge ) || build .
    "Between 1933 and 1936,  the San Francisco-Oakland Bay Bridge (also known as the Bay Bridge) across the San Francisco Bay was completed."
    source: gcdt_bio_emperor

**Exception:** when a multi-hyphen or dash functions as a nominal combinator (similar to a single hyphen), we do not segment it.

44. 国家　 文字　 改革　 委员会 ，　　适时 推出 一 种 ||
    national writing reform committee , timely launch one CL ||
    简　 ---　　繁　 之间 十分 容易 相互　　转换　 的 || 软件
    simplified - traditional between very easy each-other convert DE || software
    "The National Character Reform Committee will launch a software that is very easy to convert between simplified --- traditional characters. "
    source: gcdt_news_simplified

Similar to dashes, colons introduce new EDU segments even if the fragment occurs after the colon is a word or phrase, "as long as the text that follows the colon provides further elaboration on the topic introduced by the colon" (Carlson et al. 2003). In other words, when it is not adnominal, we segment them.

45. 又 翻译　　作 :‖ 雅可比
    also translated as : ‖ Jacobi
    "Also translated as: Jacobi"
    source: gcdt_bio_galois

46. 英语　中 主要　分为　　三个　"态"：‖
    English in mainly divide-into three CL "voices": ‖
    主动态 ， 　中动态　　和 被动态 。
    active-voice , middle-voice and passive-voice .
    "English is mainly divided into three "voices": active voice , middle voice and passive voice."
    source: gcdt_academic_iconicity

Exceptions apply when the nominal phrase after the colon is adnominal:

47. 也 有些　　学者 认为 是 骨骼　的　 发育不良 ‖ *[ 19 ] : pp. 3–4* 。
    also some scholar think COP skeleton DE　dysplasia ‖ [ 19 ] : pp. 3–4 。
    "Some scholars also believe that it is the dysplasia of the bones ‖ [19]: pp. 3–4."
    source: gcdt_bio_byron

However, phrases separated by semicolons are not separate EDUs. The following example is a single long EDU.

48. *1月　25日 14*时 ， 封闭　 沪渝　　　　　　高速　　 黄石 ；
    January 25th 14：00 , close Shanghai-Chongqing Expressway Huangshi;
    大广　　　　高速　黄石西 、　大冶　金湖 、阳新　龙港 ；
    Daguang Expressway Huangshi West, Dayu Jinhu, Yangxin Longgang;
    杭瑞　高速　　阳新 枫林 、 木港 、 排市 ；
    Hangrui Expressway Yangxin Fenglin, Mugang, Paishi;
    黄咸　　　　高速 大冶 陈贵 、　　灵乡 、 金牛 共 *10* 个 出口
    Huangxian Expressway Daye Chengui, Lingxiang, Jinniu total 10 CL exits
    "By 2:00 pm on Jan 25th, ten exits are closed: Huangshi exit on Shanghai-Chongqing Expressway; Huangshi West, Dayu Jinhu, Yangxin Longgang exits on Daguang Expressway; Yangxin Fenglin, Mugang, Paishi exits on Hangrui Expressway; and Daye Chengui, Lingxiang, Jinniu exits on Huangxian Expressway."
    source: gcdt_news_hubei

In the following example, semicolons with enumerations also do not create new EDUs as long as they graphically reside in the same sentence.

*49.* *英语 中动态　　　　具有 如下　　特点：||*
English middle-voice have following characteristics : ||
***1)*** *非 事件性；　　***2)*** *泛指性；　***3)*** *施动性　；　***4)*** *情态 概念　**||**　　*[ 2 ]*。*
1) non eventuality；　2) generality；　3) agency；　　4) modal concept || 　[ 2 ]。
"English middle-voice has the following characteristics : 1) non eventuality；  2) generality；  3) agency；  4) modal concept [ 2 ]."
source: gcdt_academic_iconicity

## 1.1.10 Strong discourse cues

The RST-DT manual states that "phrasal expressions that occur with strong discourse cues are marked as EDU." In this Chinese corpus, we categorize and exemplify the following Chinese tokens or phrases as strong discourse cues. When making decisions regarding whether specific tokens are discourse cues or not, we refer Explicit Connectives annotated in the PDTB-styled Chinese Discourse Treebank (CDTB) and TED Chinese Discourse Treebank (TED-CDB).

**Adversarial Discourse Markers**
- 尽管 although
- 虽然 although
- 不管是 no matter what/how
- 除了 apart from
- 除 apart from
- 但 but
- 但是 but
- 可是 instead
- 此外 besides
- 然而 however

**Attributional Discourse Markers**
- 根据 according to
- 据 according to
- 按照 according to
- 按 according to
- 依照 according to
- 基于 based on

**Causal Discourse Markers**
- 因为 because (of)
- 所以 so
- 因 because (of)
- 由于 due to

- 基于 because of
- 经过 as a result of

**Circumstantial Discourse Markers**
- 如果 if
- … 的话 in the case of …
- 随着 along with
- 通过 by means of
- 透过 through
- 经过 through

**Coordinating Discourse Markers**
- 不论 regardless of
- 无论 regardless of
- 不但 not only
- 不仅 not only
- 而且 but also
- 还是 instead
- 并且 in addition
- 并 at the same time
- 越...越... the more… the more…

**Elaborating Discourse Markers**
- 针对 regarding

**Topic Discourse Markers**
- 对（于）… 来说 as far as…concerned (when taking an complement)
- 对（于）… 而言 as far as…concerned (when taking an complement)
- 从 ... 来看 from the view of … (when taking an complement)

Here are some examples from GCDT:

*50.* 而且 越 多 **||** 越 适得其反 。
and the-more more || the-more backfire
"And the more || the more counterproductive it is."
source: gcdt_whow_flirt

*51.* 薄荷油　　　是 天然　　的 驱逐剂 ，**||**对 啮齿类 动物 来说 **||** 太 刺激 ，
peppermint COP nature DE repellant || to rodent animal regard || too irritating
"Peppermint oil is a natural repellant, too irritating for rodents."
source: gcdt_whow_mice

*52.* 过　　一会 再　　　想　一个　　好 玩笑 || 发 过去 ，||
after while again think-of one CL good joke || send to-there , ||
总　　比　　你 弄得　　　对方　不　　自在，**||**
at-least than 2SG make the-other-one NEG comfortable , ||
然后 又　试图 给 自己 解释　**||**　　要　　容易 得 多 。
then again try GEI self explain || COP easy DER much . ||
"after a while, think of another good joke and send it over; at least that would be much
easier than making the other one uncomfortable and then trying to explain yourself."
source: gcdt_whow_flirt

For reference, the followings are strong discourse markers in English RST-DT:
because, despite, despite, regardless, irrespective, without, according to, as a result of, not only
... but also.

## 1.1.11 Translanguaging

When translanguaging happens, especially between English and Chinese, we take the English
portions as a fixed block and merge them into Chinese syntax. As a result, we DO NOT
segment the English phrases.

**53. Max hit Harry and Harry hit Max** 表示 **|| Max hit Harry** 在先，**|| Harry hit Max** 在后 。
　　　　　　　　　　　　　　　indicate　　　　　　　　at-first　　　　　　　at-last
"Max hit Harry and Harry hit Max means that Max hit Harry first, then Harry hit Max."
source: gcdt_academic_iconicity

**54. Sammy 's mad and I 'm glad** 和 **He comes , I will stay** 。
　　　　　　　　　　　　　　　　and
"Sammy 's mad and I 'm glad and He comes , I will stay."
source: gcdt_academic_iconicity

Similarly, foreign book titles should be blocked, thus not segmented.

*55.* "珍娜·玛柏 ”　　　这 频道　　成名　　　　于《 **How to trick people**
" Jenna Marber " this channel become-famous at
**into thinking you 're good looking** 》和《 **How To Avoid Talking To People**
　　　　　　　　　　　　　　　　　　　　and
**You Do n't Want To Talk To** 》这 **2** 部 视频　，**||**
　　　　　　　　　　　　　this 2 CL video , ||
其中《 **How to trick people into thinking  you're good looking** 》在 **||**
among
上传 后 的 **||**　　第一 周 便　　录得　　超级 **530万** 次 的 观看数 **|| [ 13 ] [ 14 ] ||**
upload after DE || 　first week already accept super 5.3M CL DE view || [ 13 ] [ 14 ] ||
而《 **How To Avoid Talking To People You Do n't Want To Talk To** 》则 于

and                                                    then at
**2011**年 **8**月 分别 被 《 纽约 时报 》 *‖15 ]* ‖ 和《 *ABC* 新闻 》‖ *[ 16 ]* ‖ 报导 。
2011 Aug separately BEI New York Times ‖ [ 15 ] ‖ and ABC News ‖ [ 16 ] ‖ reported .

"'Jenna Marber' became famous for two videos: 'how to trick people into thinking you 're good looking' and 'how to avoid talking to people you don't want to talk to;' among then, 'how to trick people into thinking you 're good looking' received 5.3M views one week after upload [13][14] and 'how to avoid talking to people you don't want to talk to' was reported by New York Times [15] and ABC News [16] in August 2011."

source: gcdt_bio_marble

## 1.1.12    Stranded left-side tokens

Due to pre-verbal modification in Mandarin Chinese, we often see some small segments stranded on the left side of a sentence due to the intervening strong discourse markers. We segment all these stranded spans, and form a *same-unit* relation with the discontinuous right-side span.

These are some made-up examples:

- **Adverb ‖  adjunct clause ‖ main clause**
- e.g., "However, ‖ because he likes CS, ‖ John went to CMU."

- **Subject ‖ adjunct clause ‖ main clause**
- e.g., John, ‖ because he likes CS, ‖ went to CMU.

- **PP ‖ adjunct (PP or clause) ‖ main clause**
- e.g., In the summer, ‖ because he likes CS, ‖ John will go to CMU.

## 1.2 Not Segmented Unit

In contrary to previous criterions for EDU swegmentation, we exemplify situations where a clause or sentence is not segmented into separate EDUs.
The same double-pipe symbol || is still used to indicate segmented EDU boundaries.

### 1.2.1 Complement clauses

**Complement clauses are not segmented,** for example, clausal subjects and objects.

*56.* 甚至 更　　 让　　 考官　 恼怒　 的 是，　　　　　 他　　 将 **||**
even more make examiner angry DE COP ,　　　　　 3SG.M BA　　 **||**
擦 黑板 的 **||**　　　 抹布 扔在 了 考官　　　 的 脑袋 上
erase black-board DE **||** rag throw-at LE examiner DE head on-top-of
"What annoyed the examiner even more was that he threw the rag for erasing the chalkboard on the examiner's head."
source: gcdt_bio_galois

*57.* 求解　 复合　 运动　　　 经常 需要 把 运动　　　　 分解
solve compound movement　 usually need　BA movement decompose
为 平移 和 转动 。
to-be translation and rotation .
"Solving compound motions often requires decomposing the motion into translations and rotations."
source: gcdt_bio_bernoulli

*58.* 这 两　 方面　 的 原因 使得 他 形成 了 孤僻　　 和　　 忧郁　 的 性格 。
this two aspect DE reason make 3SG.M form LE solitary and melancholy DE personality.
source: gcdt_bio_byron

**Subject clauses are not segmented in Chinese, not even for coordinated subject clauses.**

*59.* 因此，　　 研究 老年人 言语　　　　 感知 特点
therefore , study　elderly speech perception characteristics
和 探索　　　 老年人 言语 感知 策略 , 能 **||**
and explore elderly speech perception strategy , can **||**
为 提高 老年人　　　 言语 感知　　 能力 **||**　　 提供 参考 , **||**
for improve　elderly　speech perception ability **||** provide reference , **||**
也 能　对 老年人　　　 言语 感知　　 障碍　　 的 临床 诊断　　 治疗 、
also can for elderly speech perception disorder DE clinical diagnosis treatment ,
老年人 助听器　　 的 研发　　　 提供 新 思路 , **||**
elderly　hearing-aids DE research provide new idea , **||**
对于 促进 老年人 与 他人　　　　 之间 的 交流　　　　　　　 沟通 **||**
for promoting elderly with others in-between DE communication communication **||**

有 着 　　　重 要 作 用 。
have PART important role .

"Therefore , studying the characteristics of speech perception in the elderly and exploring speech perception strategies in older adults can provide a reference for improving the speech perception ability of the elderly; it can also provide new ideas for the clinical diagnosis and treatment of speech perception disorders in the elderly and the research of hearing aids for the elderly; it also has an important role in promoting communication between the elderly and others."
source: gcdt_academic_aging

60. *Halliday & Matthiessen* || *[ 1 ]* || 认为 ||
　　　　　　　　　　　　　　　　believe ||

主 位 表 达 　　旧 信 息 ,
theme express old information ,

述 位 表 达 新 信 息 　　是 　　非 　　标记性 　　信 息 匹配 　　结构 ; ||
rheme express new information  COP non- marking information match structure ||

"Halliday & Matthiessen [ 1  believe that themes expressing old information and rhemes expressing new information is a non-marking information matching structure.
source: gcdt_academic_iconicity

On the other hand, **coordinated object clauses are segmented under two conditions: the verb is an attribution verb, and the objective clauses do not share the same subordinated object.**

## 1.2.2 Prepositional phrases

**Prepositional phrases are not segmented.**

61. 在 流体 力学 和 　　空气 动力学 　　中 　　有 关键性 的 作用 。
in fluid mechanics and aero dynamics within have critical DE effect .

"It plays a key role in fluid mechanics and aerodynamics."
source: gcdt_bio_bernoulli

However, **when a preposition heads a clausal complement, the phrase is separated from others.**

62. 他 　　|| 对 　修改 　版权法 , ||使 　　文件 　共享 　　合法化 ||
3SG.M || towards amending copyright-law || make document sharing legalize ||

持 开放 态度
hold open mind

"He is open to changing copyright laws and to legalizing file sharing."
source: gcdt_interview_falkvinge

63. 这样 能 避免 你 ‖ 　　被 忙碌 的 　日程 　压 得 ‖ 　喘 　不 过 气 。
this can avoid 2SG ‖ BEI busy DE schedule squeeze DER ‖ breath NEG PAST breath .
"This will prevent you from being overwhelmed by a busy schedule and out of breath."
source: gcdt_whow_procrastinating

In the following example, we see that 时 is a localizer, so the rules about prepositional phrases in this section apply. However, in the second example, 时候 is a noun; thus 你一个人远足的 is a relative clause that modifies 时候. In these two cases, even though 时候 and 时 share the same meaning, we segment them differently according to their part-of-speech.

64. 当 　　你 在 户外 远足 　　时 ,‖ 势必 　　　会 碰到 交叉 路口 。
when 2SG at outdoor hiking time , ‖ be-bound-to will hit cross road .
"When hiking outdoors, you are bound to hit an intersection."
source: gum_whow_hiking

65. 如果 ‖ 你 一 个 人 　远足 的 ‖ 时候 　　发生 了 什么 事故 , ‖
if ‖ 2SG one CL person hiking DE ‖ time happen PERF any accident , ‖
你 将 　更 难 　　　获得 帮助。
2SG will more hard retain help .
"If you are hiking alone and something goes wrong, you will have a harder time getting help."
source: gum_whow_hiking

## 1.2.3 Dislocated NPs

Topicalization happens quite commonly in Chinese. However, **dislocated NPs are not segmented.**

66. 资讯 　　　安全 , 有 无意 　　　　与 恶意 　　的 攻击者 ,
information security , has unintentional and malicious DE attackers ,
要 怎么 去 阻止
need how go stop
"How can information security stop unintentional and malicious attackers."
source: gcdt_interview_wimax

## 1.2.4 MSP

MSP is a unique part-of-speech label in Chinese treebank, reserved for a small set of "other particles." In our segmentation task, when nominals outside the MSP phrase are arguments of the predicate with the MSP phrase, we treat them as a single segment. Among such MSP particles, the most common ones are 所 (suo), 而 (er), and 来 (lai).

67. *而 没有 明确 意识 到　　|| 人 的 本质　　　所应　　　有 的 || 丰富 内涵*
but no clear realize ASP || human DEC nature **MSP** should have DEC || rich connotation
"without clearly realizing the rich connotation that human nature should have"
source: gcdt_academic_socialized

68. *许多 乌克兰人 都 能 讲述　　||*
many Ukrainians all can tell ||
*自己 祖辈　在　　大 饥荒 中 所　　　经历 的 ||　　血泪　　　历史 。||*
selves' ancestor at great famine in **MSP** experience DEC || blood-and-tear history . ||
"Many Ukrainians can tell the Blood and Tears History that their ancestors experienced in the great famine."
source: gcdt_news_famine

69. *但 很 可能 所有 他 的　　　　声明 和 行为 都　 是　　||*
but very possible all 3SG DEC statement and behavior all COP ||
*对于　 贫穷 的　　压力 而 产生 的 ||*
toward poverty DEC pressure **MSP** arise DEC ||
*富于　　　　创意 的　　|| 反应 。*
be-full-of creation DEC || reaction .
"But it's likely that all his statements and actions are reactions that arose from the pressure of poverty and are creative."
source: gcdt_bio_emperor

70. *我们 如何 从　象似性　　　角度 来　　分析 它　　呢 ？*
1PL how from iconographic angle MSP analyze 3SG.IN PART ?
"How can we analyze it from an iconographic point of view ?"
source: gcdt_academic_iconicity

On the other hand, two other MSP tokens – 以 (yi) and 去 (qu) – connect two clauses where the latter expresses the purpose of the former. In these cases, we segment them and draw a backward purpose-goal relation.

71. *获得 更 多 的 选票 , ||　　 以　　确保 我们 不　会 在 选举日　　用完 。*
get more many DEC vote , || **MSP** ensure 1PL NEG will at election-day run-out .
" to get more votes to ensure we don't run out on election day."
source: gcdt_interview_falkvinge

72. *伽罗瓦 使用 群论 的 想法　　|| 去　讨论 方程式　 的 可解性 , ||*
Galois use group DEC idea || **MSP** discuss equation DEC solvability , ||
"Galois uses the idea of the group to discuss the solvability of equations,"
source: gcdt_bio_galois

## 1.2.5 Coordinations

**Multiple verbs with the same explicit object or prepositional complement are not segmented.**

*73.* 他们 同时　　　　参加　并　　试图　获得巴黎　　大学 的
3PL simultaneously participate-in and attempt-to win Paris University DE
科学　竞赛　　　的第一 名
science competition DE first place
"They both participated and tried to win first place in a science competition at the University of Paris."
source: gcdt_bio_bernoulli

*74.* 永远 不　　　要　　试图接近　　或者跟 野生动物　　　进行互动
never NEG should　attempt approach or with wild animal conduct interact
"never try to approach or interact with wild animals."
source: gcdt_whow_hiking

These also include cases with 把(BA) or 被(BEI).

*75.* 那 就　　把手机　　　放在 其它房间　　或者直接关机 。
then just BA phone **put-in other room　or　just turn-off** .
"Then put your phone in another room　or just turn it off."
source: gcdt_whow_procrastinating

**Copula-less coordinated nominal or adjectival phrases are not segmented when conjoined with other copula-ed propositions.**

Examples:

*76.* 生　于 荷兰　　　格罗宁根 ，著名　　数学家 ，　　　　约翰·伯努利之子 ，‖
**born in Netherlands Groningen ,　famous mathematician, John-Bernoulli 's child, ‖**
为 伯努利家族代表人物之一 。　　　　　　　　‖
COP Bernoulli family representative person one-of ." 　　‖
"Born in Groningen, Holland, famous mathematician, son of John Bernoulli, is one of the representatives of the Bernoulli family."
source: gcdt_bio_bernoulli

*77.* 字　　　　宜仲 ，‖ 生　于 天津 ，　　江苏 阳湖人 ，　　　语言学家 ，‖
style-named Yizhong, ‖ **born in Tianjin,　Jiangsu Yanghu-nese,　Linguist, ‖**
精研　　　　　　　北方话　　　与 吴语方言 的 音系 。　　‖
intensively-study Northern-dialect and　Wu dialect DE phonology . " ‖
"Named Yizhong, was born in Tianjin, Jiangsu Yanghu-nese, linguist, studied the phonology of northern dialect and Wu dialect."
source: gcdt_bio_chao

## 1.2.6 Existential clauses

Simple clause 有 (you)-constructions are formed by Locative NP + Existential Verb + Object NP. Moreover, the complement of 有 can also be a clause. In this case, the locative NP, existential you, and object clause together form one EDU.

*78.* 只要　　　　路上　　　　有 其他 人　　同行
as-long-as on-the-road EXIST other people travel-together
"As long as there are other people on the road."
source: gcdt_whow_hiking

**79.** 抖音 中　　　，有 **42.1%** 的 视频 是　　关于 普通 人
douyin within , EXIST　　　DE video COP about normal people
在 疫情　　　期间 的 抗疫　　　生活 。
at epidemic　　time DE anti-epidemic life .
"In Douyin, 42.1% of the videos are about ordinary people's anti-epidemic life during the epidemic."
source: gcdt_academic_peoples

# 1.3 Compare & Contrast

## 1.3.1 Tokenization matters

when 还有 is one token, it is a CC between nominals, not an existential verb, so it does not create a new EDU.

*80.* 要　　随身　携带　急救箱 ,　　还有 手机 。
should with-you carry first-aid-kit , **and cell-phone .**
"You should carry a first aid kit and your cell phone ."
source: gcdt_whow_hiking

Faithfulness to main-subordinating clause distinction in syntax and nuclearity-satellite distinction in RST is more important than creating extra same units.

*81.* 前往 帕劳 的　|| 旅客 ，　||
go-to Palau DE || traveler , ||
在 结束 *5* 天 或 *7* 天　的　　行程 后 ||　返回　台湾
at finish 5 day or 7 day DE itinerary after || go-back Taiwan
"Travelers to Palau returned to Taiwan after finishing their 5-day or 7-day itinerary."
source: gcdt_news_bubble

82. 中国 作为 世界 第二　　　 大 的 电影 市场 ，||好莱坞 的 制片厂 老板 希望 ||
China as world second large DE movie market , || Hollywood DE studio boss hope ||
巩固　　　　 电影 在 中国 市场　　 的 前景
consolidate movie at China market DE prospect
"China as the second largest film market in the world, bosses of Hollywood studios hope to consolidate the prospects of films in the Chinese market,"
source: gcdt_news_five

## 1.3.2 The part-of-speech of some tricky tokens

Part-of-speechs are not trivial for these tokens listed below:

**Prepositions**
由 by
以 as/by
为(wèi) for
截至 till
靠 by
作为 when expressing identity or property
比起 compared to
沿着 along with
借由 by
相对于 relative to

**Localizers**

以来 up until

**Adverbs**
特别是 especially
尤其是 especially
例如 for example
一起 together
看起来 seems

**Verbs**
利用 make use of
为(wéi) COP
相比 compared to
像是 seems like

伴随 go along with
作为 when expressing regarding sth/sb as (i.e., can be replaced by 当作)
一样 the same as (e.g., in 像...一样)
那样 the same as (e.g., in 像...那样)

# 2    Relation Annotation

In this section, we present the guidelines and examples for relation anonotation.
For each example, we lay out the texts one EDU per line, followed by a screenshot of the subtree from rstweb.
For each EDU, we give its index at the beginning of the line, as well as append an automatic English translation to the end, led by a double-slash symbol //.

## 2.0    Some annotation principles

### 2.0.1 Relation marking for relative clauses

One of the most significant differences between English and Chinese regarding the structure of an RST tree is the excess amount of **combinations of same-unit + elaboration-attribute** used to structure relative clauses in Chinese. The only two attributional relations that modify part of a clause, usually a noun phrase, are elaboration-attribute and purpose-attribute. Most commonly, they are *elaboration-attribute*.

Another difference from English is that these elaboration-attribute relations are most frequently prenominal. This is because **relative clauses in Chinese are prenominal.** There is not much previous research addressing prenominal relative clauses in RST. Our decision agrees with Shinmori et al. 2003 (see Figure 5-6 on page 7), which uses Elaboration for prenominal relative clauses in Japanese.

In the following example, we observe EDU_284, "that ends the conversation first," breaks the main clause "remember to be … the person." Following previous RST-DT guidelines, we create a same-unit + elaboration-attribute structure for DU_283-285.

83. EDU_283    8 记得 做 // 8 Remember to be
    **EDU_284**    **先 结束 对话 的 // that ends the conversation first**
    EDU_285    那 个 人 。// the person .
    source: gcdt_whow_flirt

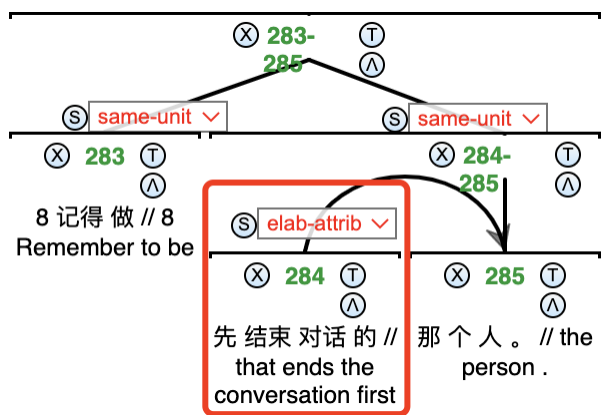

Moreover, multiple relative clauses can be coordinated to modify the same noun phrase, such as the EDU_46 and EDU_47 that modify EDU_48 below.

84. ***EDU_46*** 劳工部 报告 的 *// that the Department of Labor reported*

***EDU_47*** 上 个 星期 首 次 申请 失业 补贴 的 *// who filed for unemployment benefits for the first time last week*

*EDU_49* 人数 出人意外 地 减少 了 *2.1万 人 。// the number dropped unexpectedly by 21,000 .*

source: gcdt_news_estate

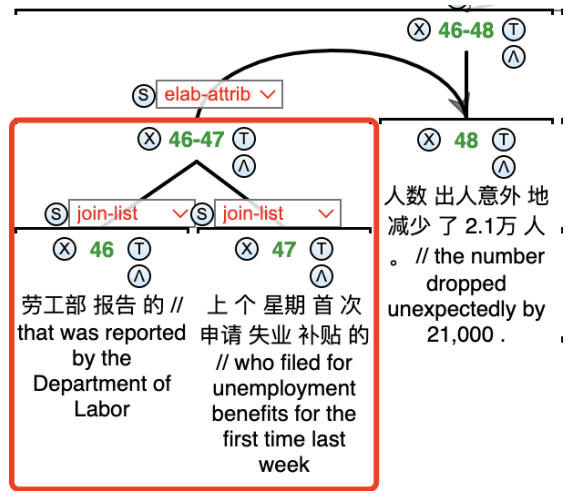

Purpose-attribute is the other attributive relation. In the following example, the stone piles are to guide the correct paths for hikers. In this case, we choose the label **purpose-attribute.**

85. EDU_235 堆石 界标 是 // Rockfill landmark is

EDU_236 用以 给 远足者们 指引 正确 道路 的 **// that is used to guide hikers on the right path**

EDU_237 石堆 。**// stone pile.**

source: gcdt_whow_hiking

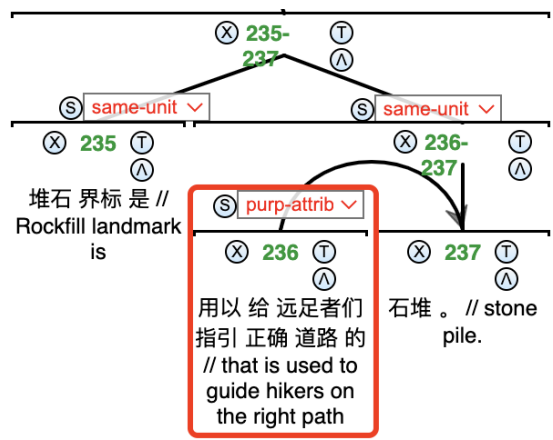

## 2.0.2 Attachment order of partial modification

In cases where the subject and object of an EDU are modified and separated by two other EDUs, we attach the subject modifier higher than the object one, based on the syntactic hierarchy that the subject governs the object.

In the following example, the relative clause modifying the subject "买家 // Buyers" is attached higher than the parenthetical "( Microsoft Corp )," which modifies the object "微软 // Microsoft."

86. **EDU_31**　　　有意 收购 **TikTok** 在 美国 、 加拿大 、 新西兰 和 澳大利亚 业务 的 **//
interested in acquiring TikTok's operations in the US , Canada, New Zealand and
Australia ||**
　　　　**EDU_32**　　　买家 包括 微软 // Buyers include Microsoft **||**
　　　　**EDU_33**　　　（ **Microsoft Corp )** 、 **// (Microsoft Corp), ||**
　　　　source: gcdt_news_tiktok

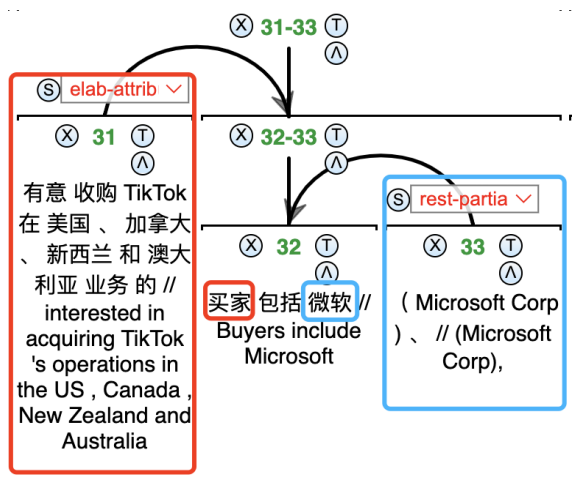

Similarly, modifier of "赞助商 // Sponsor" is higher than "参展商 // exhibitor" in the following example.

87. **EDU_5**　　　为 本次 大会 四 大 **// For the four major conferences**
　　　　**EDU_6**　　　（ 台湾 、 亚洲 、 冲刺 、 总 成绩 ） **// ( Taiwan , Asia , sprint , total score
)**
　　　　**EDU_7**　　　领骑衫 设计 的 **// jersey design**
　　　　**EDU_8**　　　赞助商 ， 同时 也 是 台北 国际 自行车展 的 参展商 // Sponsor and
exhibitor at the Taipei International Bicycle Show
　　　　**EDU_9**　　　一 德懋 国际 ， **// - Demao International,**
　　　　source: gcdt_interview_cycle

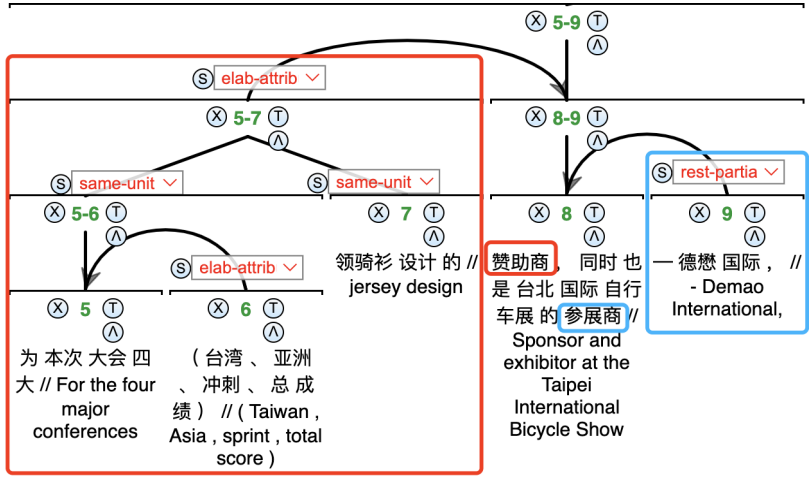

The following example shows two levels of *same-unit + elaboration-attribute* relations. At the lower level, EDU_110 "has reopened" modifies "the main library"; whereas the combination of 12 public and 5 main libraries "include Kowloon, Ping Shan Tin Shui Wai, etc." as in EDU_112.

88. EDU_109     上述 12 个 公共 图书馆 和 5 个 // The 12 public libraries mentioned above and 5

**EDU_110**     已 重新 开放 的 // **has reopened**

EDU_111     主要 图书馆 // main library

**EDU_112**     （包括 九龙 ，坪山 天水围 ，沙田 ，荃湾 ，屯门 公共 图书馆 ）// **(Including Kowloon, Ping Shan Tin Shui Wai, Shatin, Tsuen Wan, Tuen Mun Public Library)**

EDU_113     的 学生 学习室 将 在 同 一 天 恢复 服务 。 // The student study rooms will resume service on the same day.

source: gcdt_news_kangle

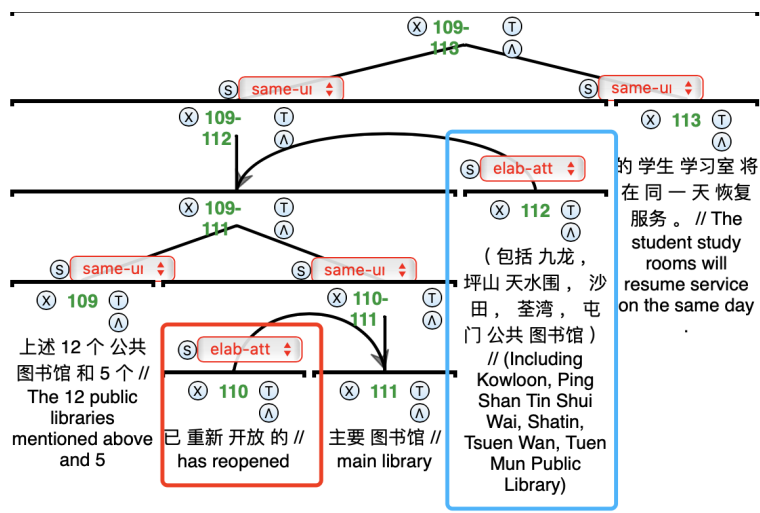

## 2.0.3 Implicit coordination

In Chinese, implicit verb phrase coordination is joint, and they are annotated as conjunctions syntactically in the Chinese Treebank.

In this corpus, unless there is a significant nucleus-satellite imbalance between these implicitly coordinated verb phrases, they form a joint-list multinuclear relation, as in the following example.

89. **EDU_253**　　皇帝 并且 下旨 **// the emperor decrees**
    **EDU_254**　　要求 **// requires**
    EDU_255　　建造 // put up
    EDU_256　　连接 奥克兰 和 旧金山 的 // that connects Oakland and San Francisco
    EDU_257　　吊桥 ， // drawbridge,
    source: gcdt_bio_emperor

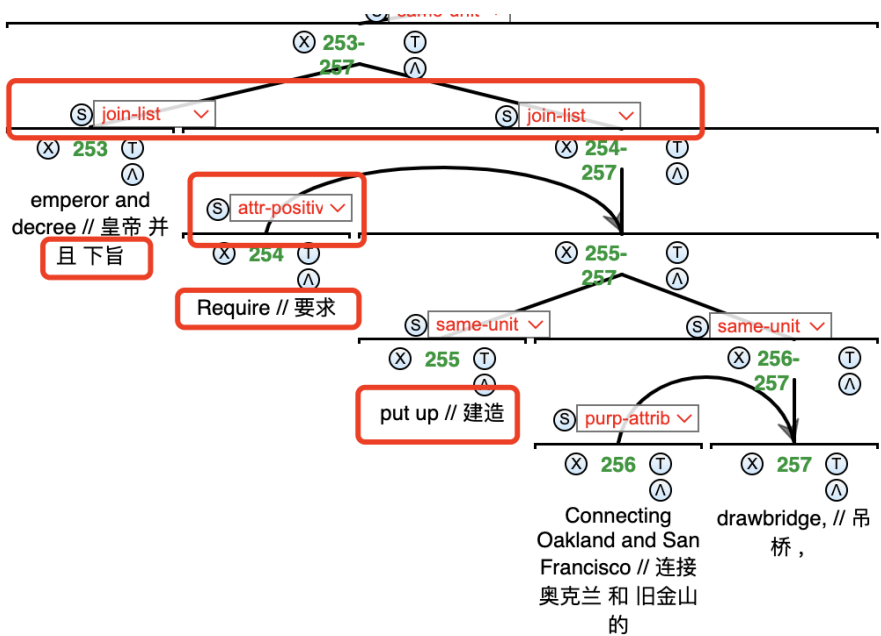

In contrast, in the following example, the content of the saying event is more important than how the saying is addressed (by quoting a traveler). Thus, the verb phrases before 说 "say" are considered satellites.

90. EDU_58　　中央社 // Central News Agency
    EDU_59　　引述 旅游 业者 的 话 **// To quote a traveler**
    EDU_60　　说 ， **// explain ，**
    EDU_61　　帕劳 // Palau
    EDU_62　　受限于 机场 ， // limited by the airport,
    EDU_63　　只 能 起降 100 座 以上 的 单走道 飞机 ， // Can only take off and land single-aisle aircraft with more than 100 seats,
    source: gcdt_news_bubble

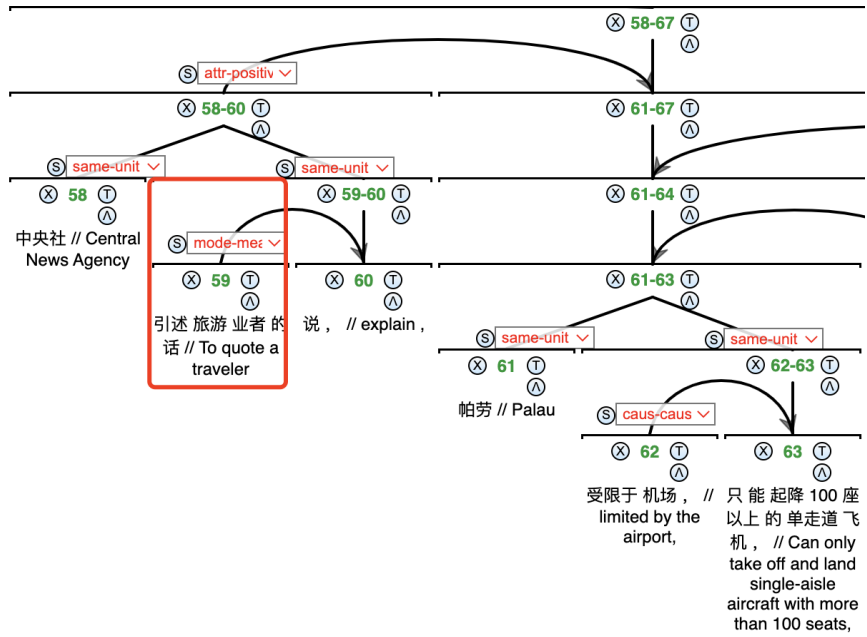

## 2.1 Nucleus-Satellite Relations

### 2.1.1 adversative-antithesis

**adversative-antithesis: the Reader finds the Nucleus more credible than the Satellite.**
For example, in the following example, "after Tao gives birth to the heaven and earth" (EDU_150), it is not the case that "it is all done" (EDU_151), instead it continues to take care of the world (DU_152-155). Thus, EDU_151 is an antithesis of EDU_152-155.

91. EDU_150　　而 道 创生 了 天地 万物 之后 ，// And after Tao created the heaven, earth, and all things,
EDU_151　　并 非 就 大功告成，**// It's not all done,**
EDU_152　　它 还 继续 生养 万物 ，// And it continues to beget all things,
EDU_153　　运化 万物 ，// transport all things,
EDU_154　　参与 万物 的 流行 变化 ，// Participate in the popular changes of all things,
EDU_155　　养育 和 辅助 万物 的 成长 发育 。// Nurturing and assisting the growth and development of all things.
source: gcdt_academic_taoist

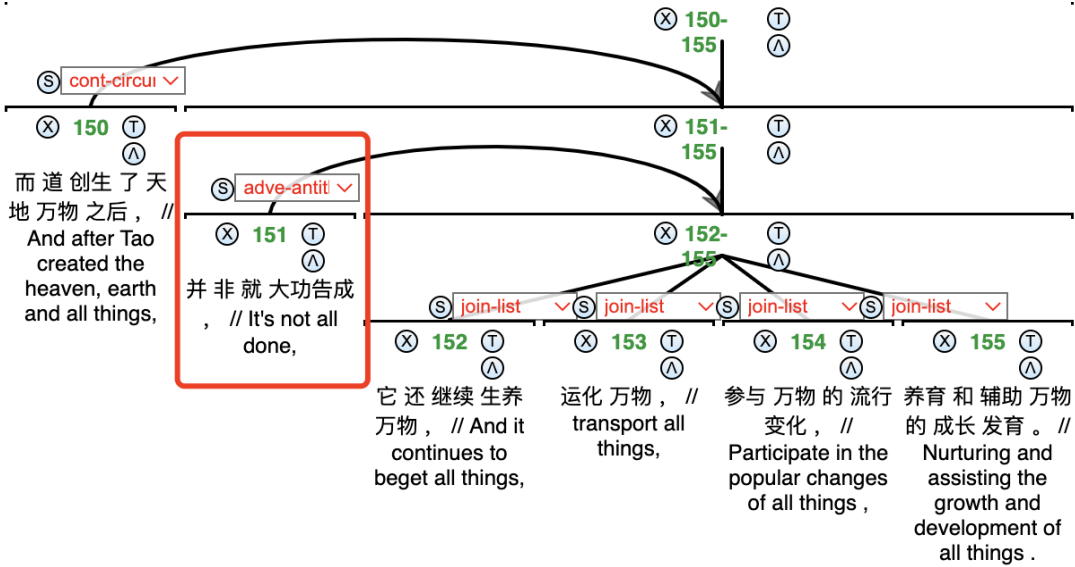

In the following example, "when it comes news," the unsurprising in EDU_65 is overridden by the "extra vigilant" in EDU_68. Thus we label DU_62-65 → DU_66-72 an adversative-antithesis.

92. **EDU_62**     我 知道 // **I know**
**EDU_63**     这些 文章 是 不 完整 的 ， // **These articles are incomplete,**
**EDU_64**     而且 似乎 是 倾斜 的 ， // **And it seems to be sloping,**
**EDU_65**     但 我 并不 感到 惊讶 。 // **But I'm not surprised.**
EDU_66     但 当 涉及 到 新闻 时 ， // But when it comes to news ,
EDU_67     管理 维基 新闻 的 // management of wiki news
EDU_68     人 需要 格外 警惕， // One needs to be extra vigilant,

source: gcdt_interview_ward

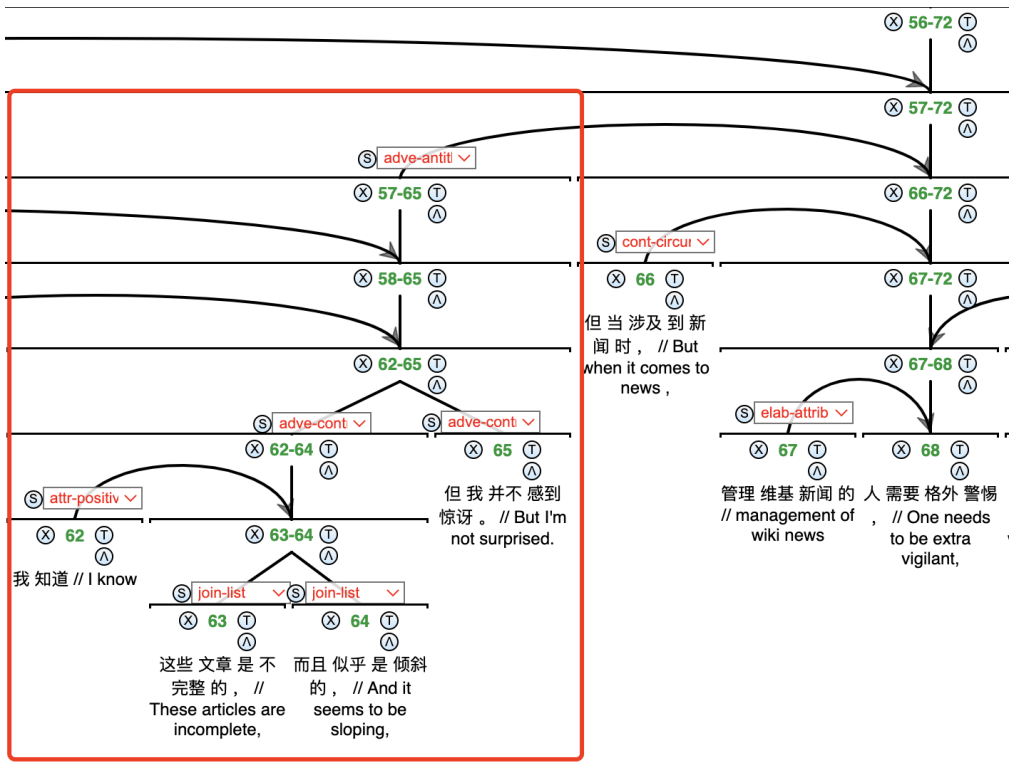

## 2.1.2 adversative-concession

**adversative-concession: the Writer admits the Satellite but still claims the Nucleus.**
Concession is the more frequent Nucleus-Satellite adversative relation in GUM. It is usually the scenario where one acknowledges the factuality of the Satellite but still stands for the Nucleus. Examples below show such preferences:

93. **EDU_82**　　传统 的 三角形 球框 也 可以 用来 摆 九 球 ， **// The traditional triangular ball frame can also be used to place nine balls,**
    EDU_83　　但是 球 之间 的 空隙 比较 大 。 // But the gap between the balls is relatively large.
    source: gcdt_whow_pool

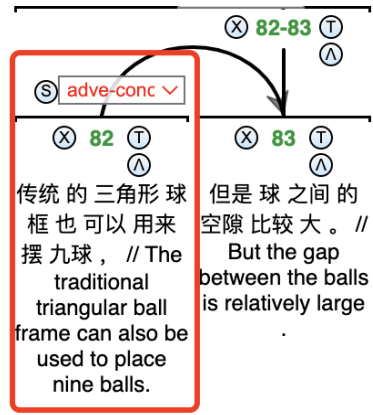

94. **EDU_29**　　虽然 仍然 是 三 镜头 设计 ， **// Although it is still a three-lens design,**
    EDU_30　　但 每 颗 镜头 都 有 明显 的 进步 。 // But each lens is a marked improvement .
    source: gcdt_news_apple

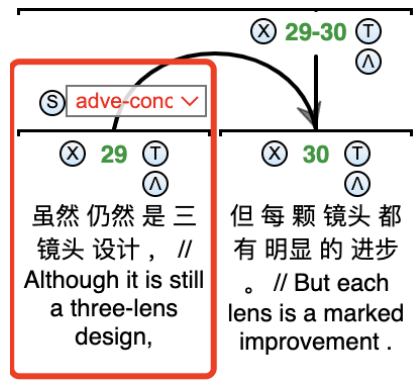

### 2.1.3 attribution-negative

**attribution-negative: the Satellite negates the source of information to the Nucleus.**
Attribution relations are essential to RST, where one addresses the content of the information more than its source. In other words, who said it is less important than what is said.
Like English GUM, we differentiate between a negative versus a positive source of information.
An **attribution-negative** is when the source of information is negated, like in the following example.

**95.** 其他 人 不 知道 **// other people don't know**

拜伦 是 双性恋 。 ” // Byron is bisexual. "

source: gcdt_bio_byron

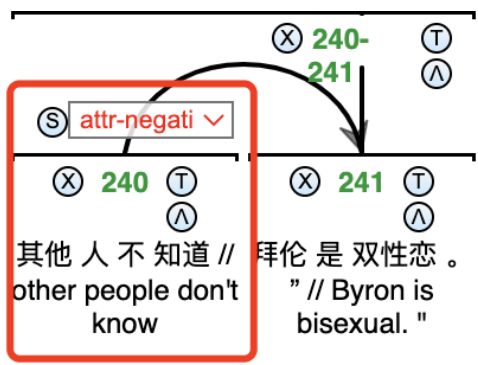

### 2.1.4 attribution-positive

**attribution-positive: the Satellite provides a positive source of information to the Nucleus.**
**See Section 1.1.4 for the list of attribution verbs.**
On the other hand, we have more frequently attribution-positive as in the following example.

96. **EDU_247**　　　有些 现代 的 医学家 认为 **// Some modern medical scientists think that**
　　EDU_248　　　这 是 小儿 麻痹症 的 结果 ， // This is the result of polio,
　　**EDU_249**　　　也 有些 学者 认为 **// Some scholars also think that**
　　EDU_250　　　是 骨骼 的 发育不良 // is skeletal dysplasia
　　EDU_251　　　[ 19 ] : pp. 3–4 。 // [19]: pp. 3–4.
　　source: gcdt_bio_byron

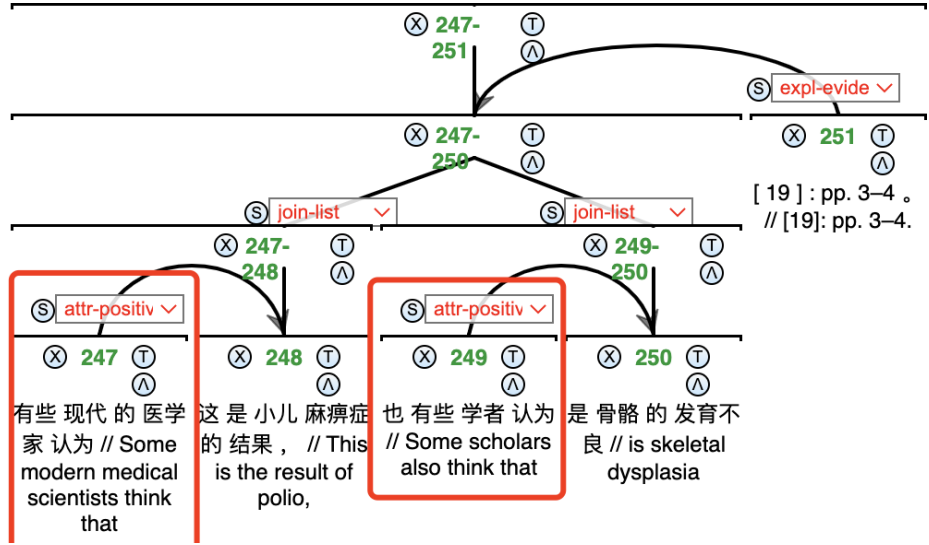

Also, note that we still label cognitive verbs with negative emotions as attribution-positive since the source of information is not negated.

**97. EDU_60** 因为 他 害怕 **// because he is afraid**

EDU_61 乌克兰 民族 运动 会 与 布尔什维克 革命 相 竞争 。 // The Ukrainian National Games competed with the Bolshevik Revolution.

source: gcdt_news_famine

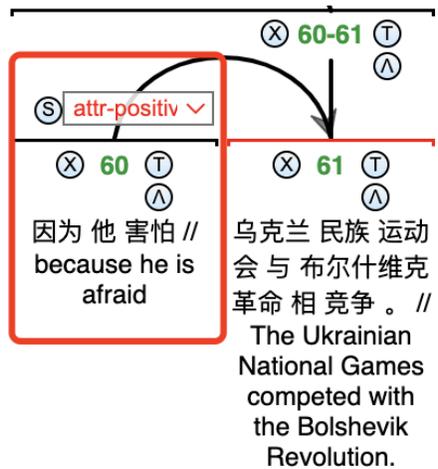

Note: the difference between *attribution-positive* and *explanation-evidence* is that the former emphasizes the **saying event,** whereas the latter only provides the source of information.

## 2.1.5 causal-cause

**causal-cause: the Satellite causes the Nucleus.**

Causal relations are predominant in RST corpora. Causal-cause labels the less prominent cause that modifies, the more prominent result.

As in the following example, "spreading the word" is more central than "not having illusion of winning."

98. EDU_86　　　在 这 一 点 上 ， 只 是 为了 帮助 宣传 ， // At this point, just to help spread the word,

    **EDU_87　　　因为 我们 并不 抱有 获胜 的 幻想 。 // Because we don't have the illusion of winning.**

    source: gcdt_interview_graaf

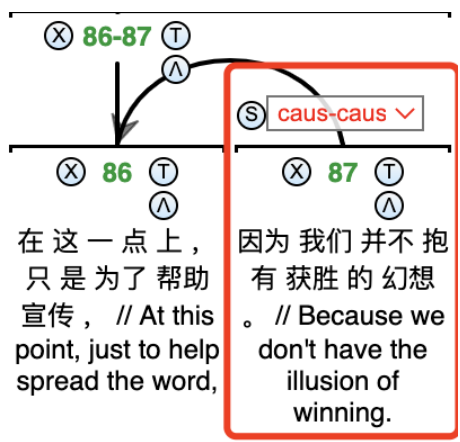

Similarly, "not being a disruptor and not winning" is the more central idea in the example below.

99. EDU_116　　　我们 并不 寻求 成为 破坏者 ， // We don't seek to be disruptors,

    EDU_117　　　也 无意 在 // also have no intention of

    EDU_118　　　导致 拜登 落选 的 // What led to Biden's defeat

    EDU_119　　　州 进行 竞选 ， // state elections,

    **EDU_120　　　因为 我们 认为 // because we think**

    **EDU_121　　　目前 最 重要 的 事情 是 结束 特朗普 总统 任期 的 畸形 状态 。 // The most important thing at the moment is to end the deformity of the Trump presidency .**

    source: gcdt_interview_graaf

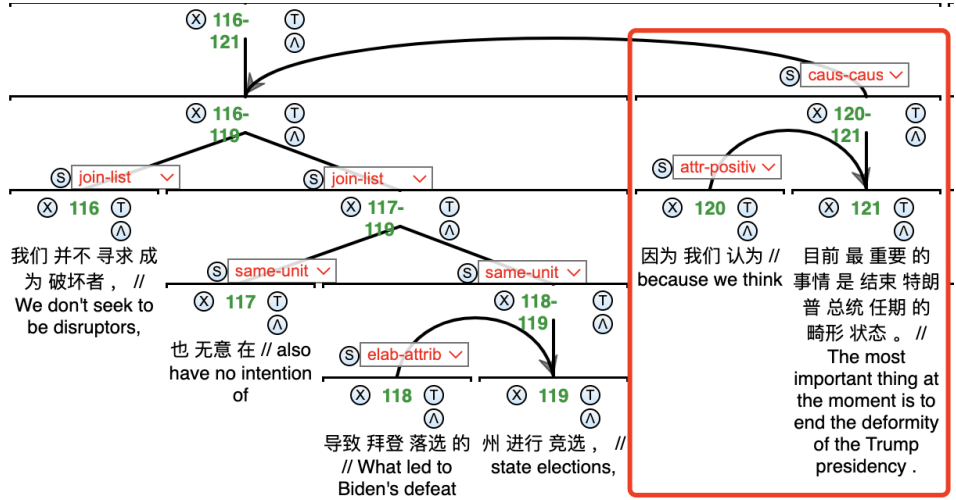

## 2.1.6 causal-result

**causal-result: the Satellite results from the Nucleus (inverse of cause).**
In these situations, the cause is more central than the result. For example, the "hiddenness" is more relevant in the context than the "difficulty to identify."

100.　　EDU_158　其 虚假 新闻 往往 隐藏 或 改编 在 真实 的 社会 热点 事件 里 // Its fake news is often hidden or adapted in real social hot events .
　　　　**EDU_159　　　而 难以 被 识别 ， // and difficult to identify,**
　　　　source: gcdt_academic_supervision

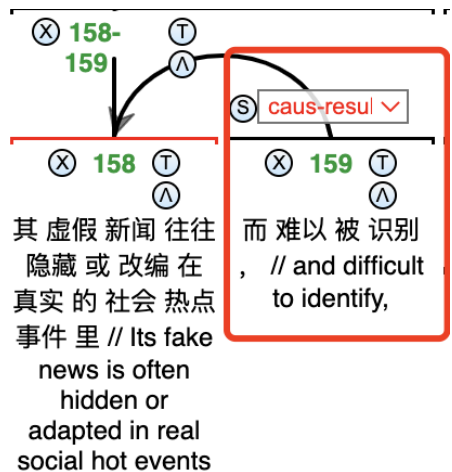

In the following example, "Byron's writing long poem" is among a sequence of events at a higher RST structure. Thus the result of his long poem is the satellite cause-result.

101.　EDU_60　1809年，面对 接踵 而至 的 攻击 和 谩骂 ， // In 1809, in the face of ensuing attacks and abuse,

EDU_61　　　拜伦 写出 长诗 《 英国 诗人 和 苏格兰 评论家 》 // Byron wrote the poem "The English Poet and Scottish Critic."

EDU_62　　　回击 攻击者 ， // hit back at the attacker,

**EDU_63　　　却 意外 地 揭开 了 积极 浪漫 主义 对抗 消极 浪漫 主义 的 序幕 ， // But unexpectedly opened the prelude of positive romanticism against negative romanticism.**

**EDU_64　　　长诗 也 使得 拜伦 在 英国 诗歌 文坛 中 初露 锋芒 。 // The long poems also made Byron first appear in the British poetry and literary world.**

EDU_65　　　[ 1 ] [ 2 ] // [ 1 ] [ 2 ]

source: gcdt_bio_byron

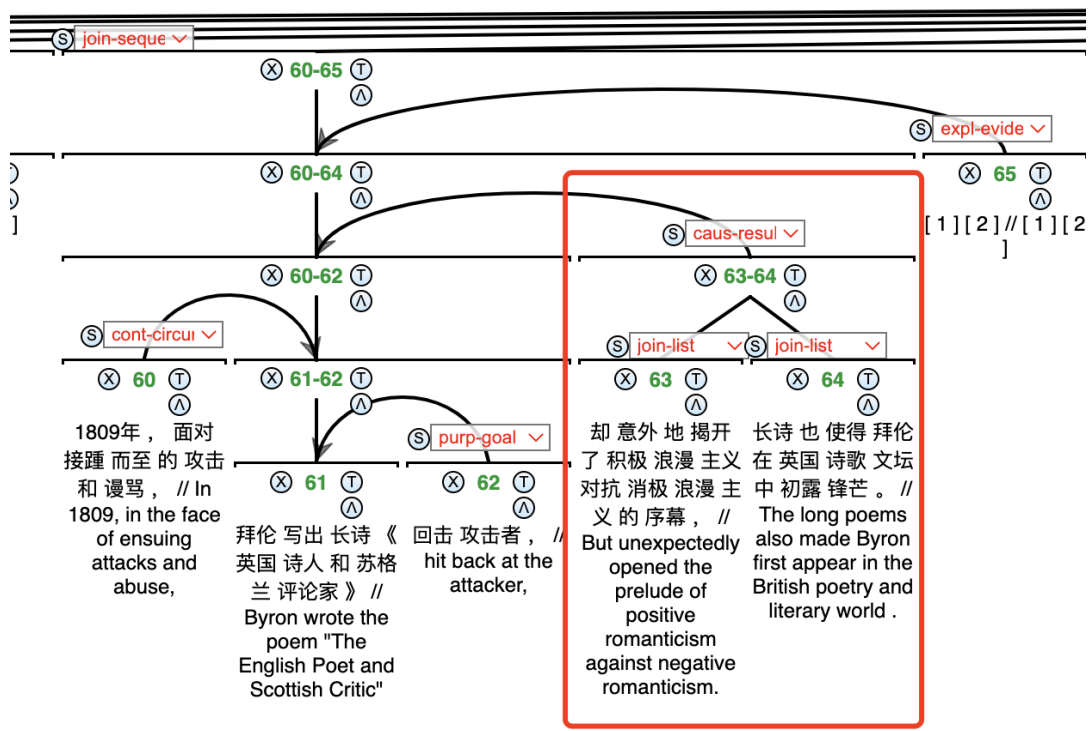

## 2.1.7 context-background

**context-background: the Reader needs to know the Satellite to understand the Nucleus.**
The Satellite provides the context for the Nucleus, and the reader needs to know the satellite to understand the nucleus.
In the following example, the equivalent of English "besides" is an excellent example of context-background.

102. **EDU_74** 除了 直接 问 人家 问题 ， **// In addition to asking people questions directly,**
EDU_75　　还 可以 给出 你 自己 的 看法 。 // You can also give your own opinion.
source: gcdt_whow_flirt

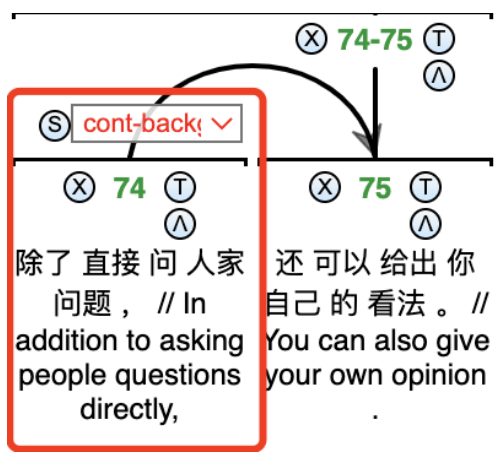

In this extended example, Edward's participation in dozens of movies and shows sets up his reputation and makes what he says credentialed.

103. **EDU_121** 好莱坞 的 韩裔 演员 爱德华·金 演出 过 几十 部 电影 和 电视剧 ， **// Hollywood actor Edward King has appeared in dozens of movies and TV shows ,**
EDU_122　　他 说 ： // He said :
EDU_123　　" 好莱坞 有 很多 机会 ， // "Hollywood has a lot of opportunities,
EDU_124　　但 总是 些 // but always
EDU_125　　只 有 一 句 对白 的 // only one line of dialogue
EDU_126　　小 角色 ， // small role,
source: gcdt_news_five

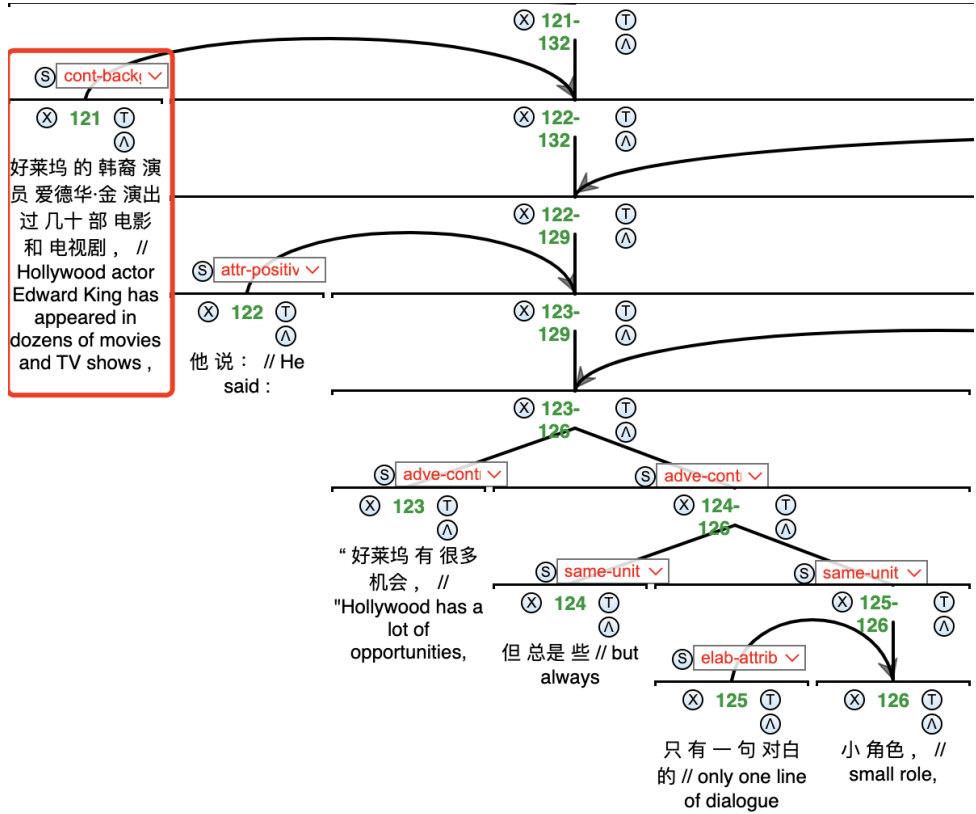

## 2.1.8 context-circumstance

**context-circumstance: the Satellite gives circumstances, e.g., time, place, of the Nucleus.**

**104.** ***EDU_53*** 她 撰写 这 本 书 时 *// While she was writing this book*
*EDU_54* 研究 了 当年 斯大林 与 乌克兰 共产党 领导人 之间 的 往来 信件 。 // *A study of letters between Stalin and the Ukrainian Communist Party leaders in those years.*
source: gcdt_news_famine

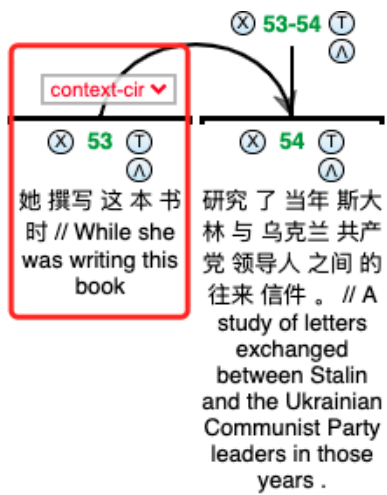

**105.** **EDU_116** 当 身处 在 一 个 宽带 或是 无线 网络 的 环境 内 ， // **When in a broadband or wireless network environment ,**
EDU_117 没 人 会 想到 任何 // no one will think of any
EDU_118 （ 好 或 坏 的 ） // (good or bad)
EDU_119 资讯流 的 出现 ， // The emergence of information flow,
source: gcdt_interview_wimax

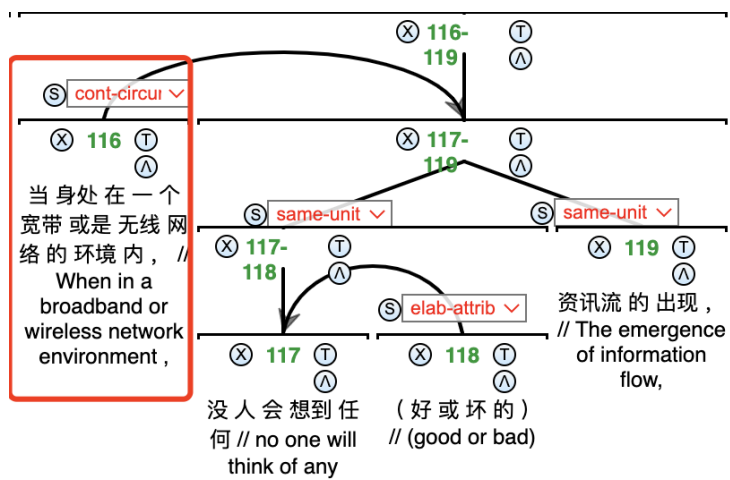

## 2.1.9 contingency-condition

**contingency-condition: the Satellite is a condition for the Nucleus to happen.**

As stated in the RST-DT manual, "the truth of the proposition associated with the nucleus is a consequence of the fulfillment of the condition in the satellite."

In the following example, 一旦 "once" sets up a hypothetical condition where mortality can be nearly 100%.

**106.** **EDU_20**  一旦 发生 疾病 ， **// Once a disease occurs,**

    EDU_21  死亡率 近 100% 。// Mortality is nearly 100%.

    source: gcdt_academic_rabies

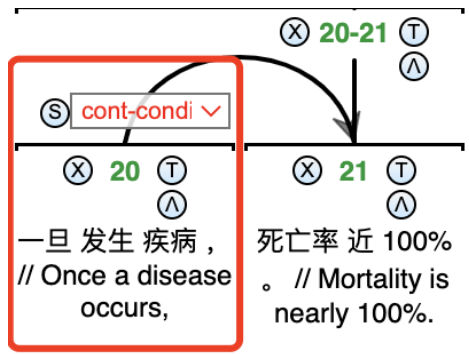

Similarly, the increase in retweets and comments is a condition for Douyin to provide traffic support in the following example.

**107.** **EDU_58**  而且 **// and**

    **EDU_59**  用户 转发 和 评论 虚假 新闻 的 **// Users retweet and comment on fake news**

    **EDU_60**  数量 短 时间 内 增多 ， **// The number increased in a short period,**

    EDU_61  抖音 会 对 该 短 视频 进行 流量 扶持 ， // Douyin will provide traffic support for this short video.

    source: gcdt_academic_supervision

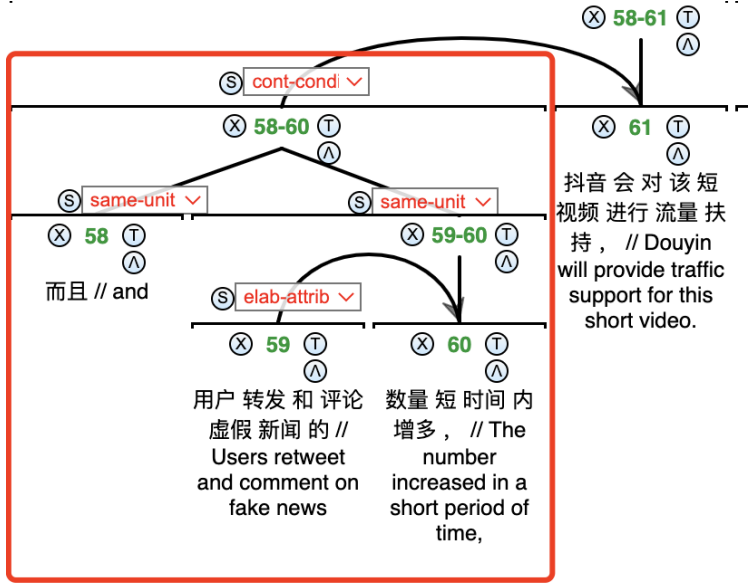

## 2.1.10 elaboration-additional

**elaboration-additional: the Satellite provides more information about the Nucleus.**
This is the most general "last resort" relation when the satellite gives more information about the nucleus. In practice, one annotates a nucleus-satellite relation as elaboration-additional only when other relations are not as suitable.

For example, the part-and-whole relation between John de Graff and the Bread and Rose Party is annotated as elaboration-additional.

108.　　EDU_2　　约翰·德·格拉夫 // John de Graaf
　　**EDU_3（面包 和 玫瑰党）// (Bread and Rose Party)**
　　source: gcdt_interview_graaf

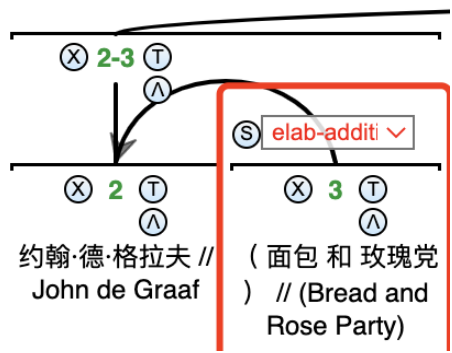

More generally, we use the elaboration-additional relation when further explanations are made on a statement. For example, the satellite in the example below explains what a "left-wing self-proclaimed socialist" means.

109.　EDU_25　该 党 持 左派 、 // The party is left-wing,

EDU_26　自称 " 社会 主义 " 的 // that self-proclaimed "socialist"

EDU_27　立场 ， // position,

**EDU_28　包括 就业权 、 带薪 休假 、 // including employment rights , paid leave ,**

**EDU_29　缩短 每周 工作 时间 、 // Shorten working hours per week,**

**EDU_30　承认 巴勒斯坦国 等 。 // Recognition of the State of Palestine etc.**

source: gcdt_interview_graaf

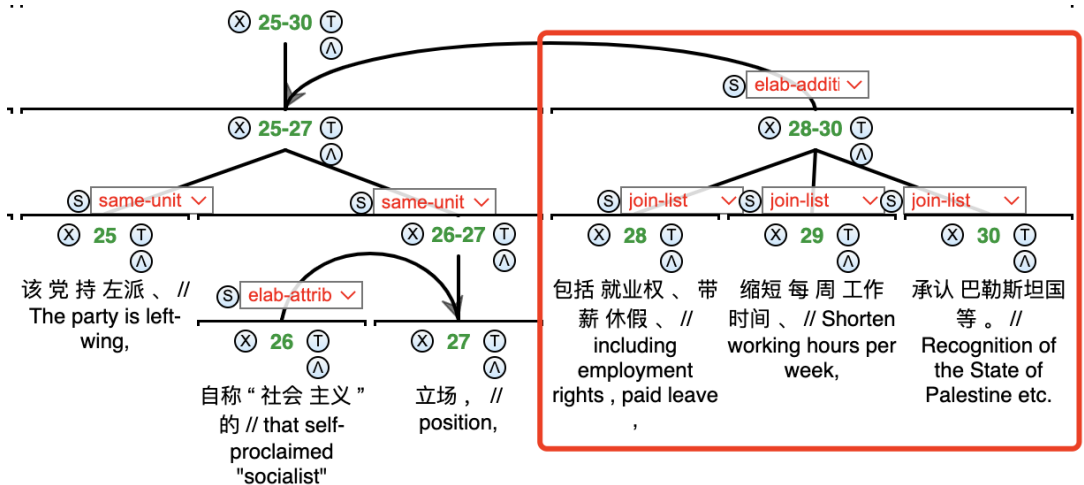

## 2.1.11 elaboration-attribute

**elaboration-attribute: the Satellite provides more information about some phrase (rather than the entire clause) in the Nucleus.**

Functionally, the satellite provides the same detail to the nucleus. The significant difference is that instead of modifying the whole clause of the nucleus, it only modifies a particular phrase, most frequently a noun phrase.

Since Chinese relative clauses are placed before the noun head, the structure of a higher *same-unit* and a lower *elaboration-attribute* is quite common in this dataset.

In the previous example, the phrase "that self-proclaimed socialist" modifies the noun phrase "position" instead of the clause (possession of a position).

110.　　EDU_25　该 党 持 左派 、 // The party is left-wing,
　　　**EDU_26　自称 " 社会 主义 " 的 // that self-proclaimed "socialist"**
　　　EDU_27　立场 ， // position,
　　source: gcdt_interview_graaf

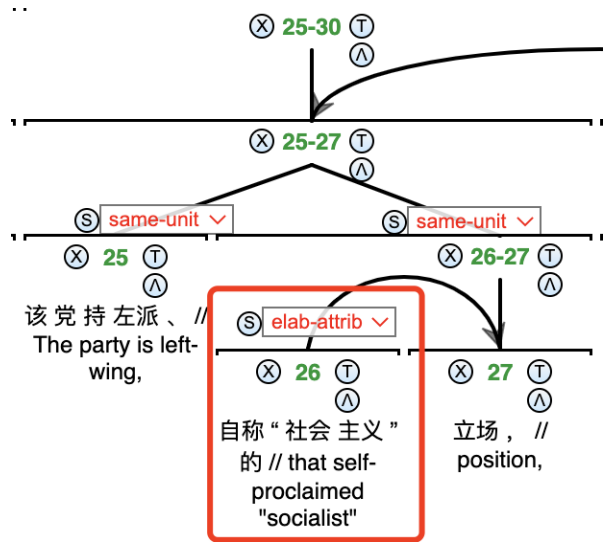

Elaboration-attribute can also label modifiers that do not surface as relative clauses. In the following example, we see three elaboration-attribute relations:

- the rabbit **that died of rabies on June 21**
- the emulsion **that was obtained from the spinal cord of a rabbit**
- the emulsion that was kept in dry air for 15 days

The first two are relative clauses, but the third is parenthetical.

111.　　EDU_72　　Grancher 说服 了 巴斯德 提供 // Grancher persuaded Pasteur to provide
　　**EDU_73　　6月 21日 死于 狂犬病 的 // died of rabies on June 21**
　　**EDU_74　　兔子 脊髓 中 获得 的 // obtained from the spinal cord of a rabbit**
　　EDU_75　　乳剂 // Emulsion
　　**EDU_76　　( 乳剂 在 干燥 的 空气 中 保存 了 15 天 ) 。 // (The emulsion is kept in dry air for 15 days).**
　　source: gcdt_news_rabies

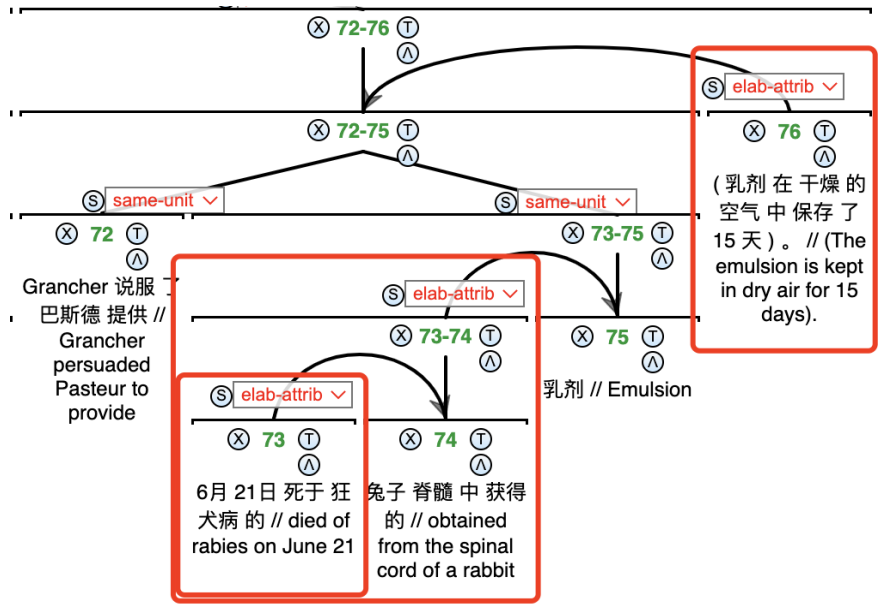

## 2.1.12 evaluation-comment

**evaluation-comment** is a Nucleus-Satellite relation in which the Satellite gives an opinion about the Nucleus (that the Reader does not need to agree with).

In the following example, EDU_177 evaluates that Zhao Yuanren, as a distinguished professor, humbly asks his students about the Anhui dialect.

112.　　EDU_176 赵元任 虚心 求教 安徽 方言 ，// Zhao Yuanren humbly learn about the Anhui dialect,

EDU_177 没有 老师 架子 。// does not behave like a teacher at all.

source: gcdt_bio_chao

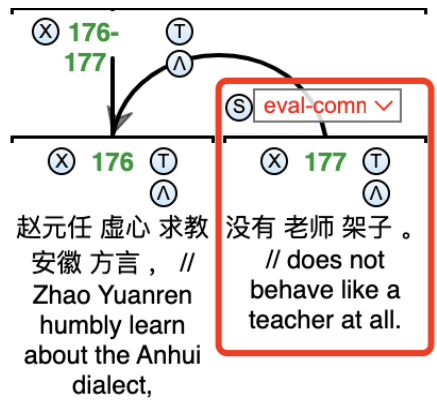

When deciding between **attribution-positive** and **evaluation-comment**, the positive sentiment is more essential than the source of information. Thus, DU_29-31 → EDU_32 is labeled **evaluation-comment**.

113.　　EDU_29 我们 // We

EDU_30（ 德懋 国际 ） // ( Demao International )

EDU_31 非常 荣幸 ，// are very honored ,

EDU_32 能 赞助 这 次 的 环台赛 。// that (we) can sponsor this Taiwan Tour .

source: gcdt_interview_cycle

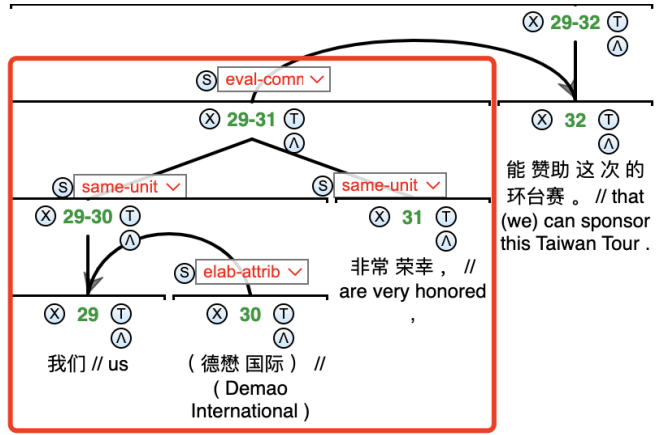

## 2.1.13 explanation-evidence

**explanation-evidence: the Satellite gives evidence that the Nucleus is true.**

One typical example of an evidence DU is the citation as below, the square-bracket citations are evidence for the preceding quotes or transliterations.

114.　EDU_54　抖音 虚假 新闻 的 裂变式 传播 数学 模型 往往 呈现 指数式 的 增长 ， // The mathematical model of the fission propagation of Douyin fake news tends to grow exponentially .

EDU_55　用户 可 对 抖音 的 虚假 新闻 进行 转发 ， // Users can retweet fake news on Douyin ,

EDU_56　成为 新 的 传播 链条 上 一 个个 节点 // Become a node on a new propagation chain

**EDU_57　[ 1 ] ，// [ 1 ] ，**

source: gcdt_evidence_supervision

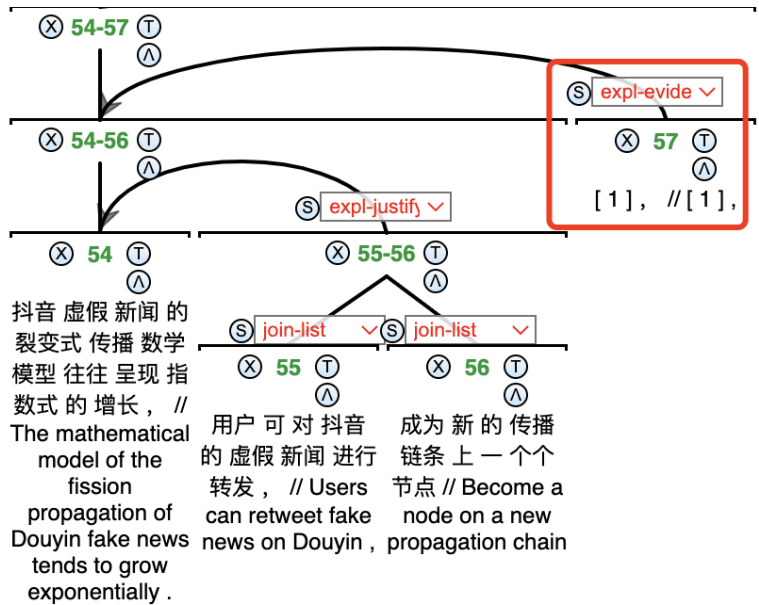

In the following example, the scenario facing fake producers is an example of problematic law enforcement.

115.　EDU_200　各级 互联网 信息 办公室 是 // Internet Information Offices at all levels are
EDU_201　　　负责 抖音 虚假 新闻 监督 管理 的 // Responsible for the supervision and management of Douyin fake news
EDU_202　　　主管 部门 ， // competent department,
EDU_203　　　但 与 公安 部门 // But with the police department
EDU_204　　　在 监管 执法 时 // in regulatory enforcement
EDU_205　　　存在 监管 执法 权力 交叉 和 双方 监管 信息 未 及时 共享 等 问题 。 // There are problems such as overlapping of regulatory and law enforcement powers and failure to share regulatory information between the two parties in a timely manner.
**EDU_206　　　尤其是 面对 虚假 新闻 制作人 是 自然人 时 // Especially when a fake news producer is a natural person**
**EDU_207　　　往往 由 公安 机关 // Often by the police**
**EDU_208　　　以 寻衅滋事 为 由 // On the grounds of picking quarrels and provoking trouble**
**EDU_209　　　处以 行政 拘留 // administrative detention**
**EDU_210　　　而 没有 // without**
**EDU_211　　　按照《 互联网 新闻 信息 服务 管理 规定 》 // According to "Internet News Information Service Management Regulations."**
**EDU_212　　　由 互联网 信息 办公室 处罚 。 // Penalized by the Internet Information Office.**
source: gcdt_academic_supervision

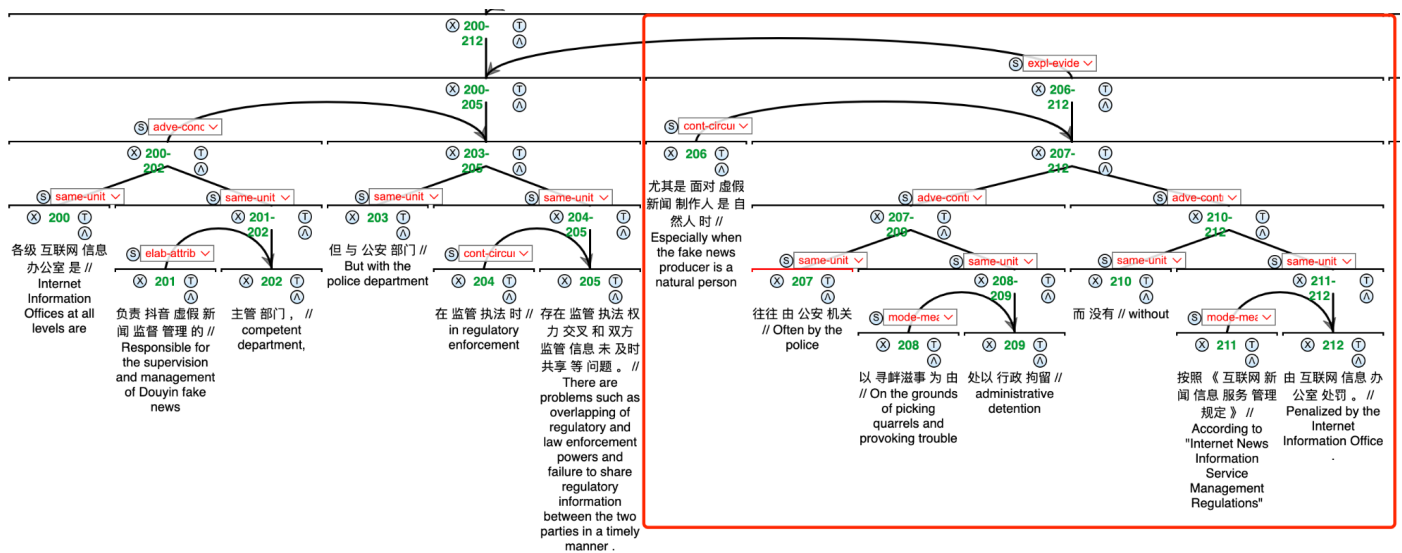

## 2.1.14 explanation-justify

**explanation-justify: the Satellite justifies why the Writer can say the Nucleus.**

The satellite explanation-justify gives further explanation to the reader why the author states the nucleus.

In the following example, "giving me an extra birthday present" justifies why the speaker does not want correct his wrong birthday on Wikipedia.

116.  EDU_36  嗯 ，我 的 资料 有 一 点 错误 ， // Well, there is a little error in my information,

EDU_37  我 的 出生 日期 是 1969年 10月 24日 ， // My date of birth is October 24, 1969,

EDU_38  不过 可以 不 用 改 . // But you don't have to change it.

**EDU_39  就 当 是 多 送 我 一 份 生日 礼物 吧 。 // Just give me an extra birthday present.**

source: gcdt_interview_keyman

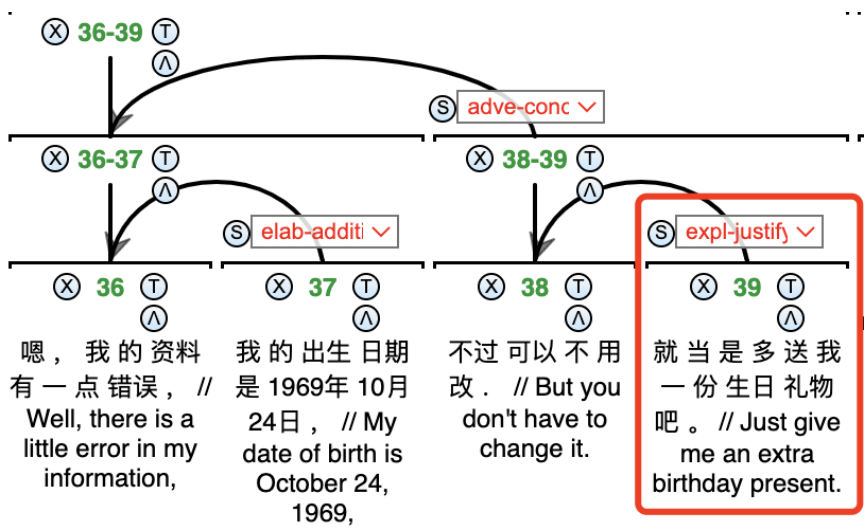

In Chinese, 鉴于 is a word that means "based on…" or "given that…," usually referring to a prescribed rule or principle. Here is an example of two consecutive 鉴于 explanation-justify examples.

117.   **EDU_174 ——— " 鉴于 ， 华盛顿市 有 一 群 人 自称 国会 议员 ， // — "Whereas, there is a group of people in Washington who call themselves members of Congress, ||**

EDU_175      这 违反 了 // this violates  ||

EDU_176      10月 12日 发表 宣布 // Published on October 12th  ||

EDU_177      国会 被 废除 的 // Congress abolished  ||

EDU_178      皇家 法令 ； // royal decree;  ||

**EDU_179      鉴于 ， 朕 有 必要 严密 地 遵照 帝国 的 旨令 ； // Whereas, it is necessary for me to abide by the decrees of the Empire strictly;  ||**

EDU_180      现在 ， 因此朕命令 和 指示 军队 的 最高 司令官 ， 陆军 少将 史考特， // Now, therefore, I order and instruct the supreme commander of the army, Major General Scott,

EDU_181      收到 朕 的 旨令 后 ， // After receiving my order,  ||

EDU_182      立刻 以 适当 的 力量 清洗 国会 大厅 。 " // Immediately cleansethe halls of Congress with appropriate force . "  ||

[ 7 ] // [ 7 ]  ||

source: gcdt_bio_emperor

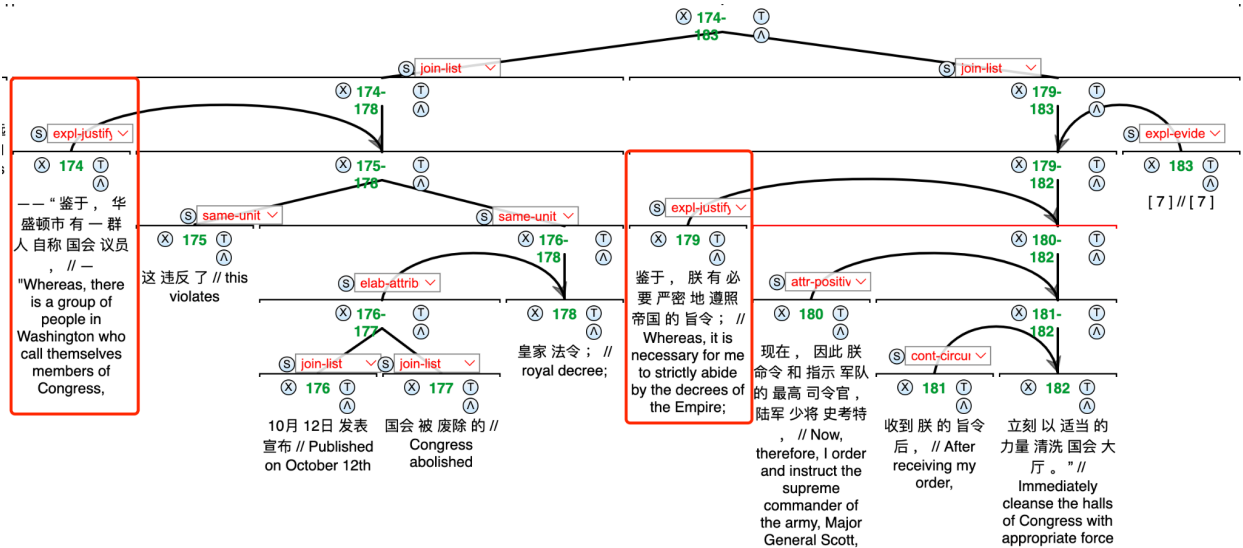

## 2.1.15 explanation-motivation

**explanation-motivation: the Satellite motivates the Reader to do the Nucleus.**
explanation-motivation rarely occurs in the written or formal text where the writer tends not to interact directly with the reader. In this corpus, organization-motivation occurs more frequently in the wikihow genre.

In this short example, asking whether the readers want to DIY their glowstick motivates them to continue reading this wikihow article.

118. **EDU_13**　还是 很 想 制作 荧光棒 吗 ？ // **Still want to make glow sticks?**
　　EDU_14　　那 就 继续 阅读 吧 。 // Then read on.
　　source: gcdt_whow_glowstick

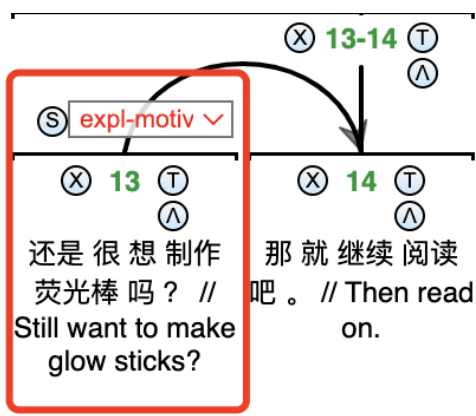

In the following example, the fact that "rats are usually tempted by food" motivates the readers to hide food from rats.

119. **EDU_18**　老鼠 跑到 屋子 里 来 一般 都 是 受到 食物 的 诱惑 。 // **Rats are usually tempted by food when they come into the house.**
　　**EDU_19**　如果 家里 什么 吃 的 都 没有 ， // **If there is nothing to eat at home,**
　　**EDU_20**　老鼠 自然 也 就 不 怎么 呆 的 下去 了 。 // **The mice naturally didn't stay much longer.**
　　EDU_21　把 食物 全 都 放在 密封 的 容器 里 ，或者是 // Put all food in airtight containers, or
　　EDU_22　老鼠 够 不 到 的 // out of reach of mice
　　EDU_23　地方 。// place .
　　EDU_24　[ 1 ] // [ 1 ]
　　source: gcdt_whow_mice

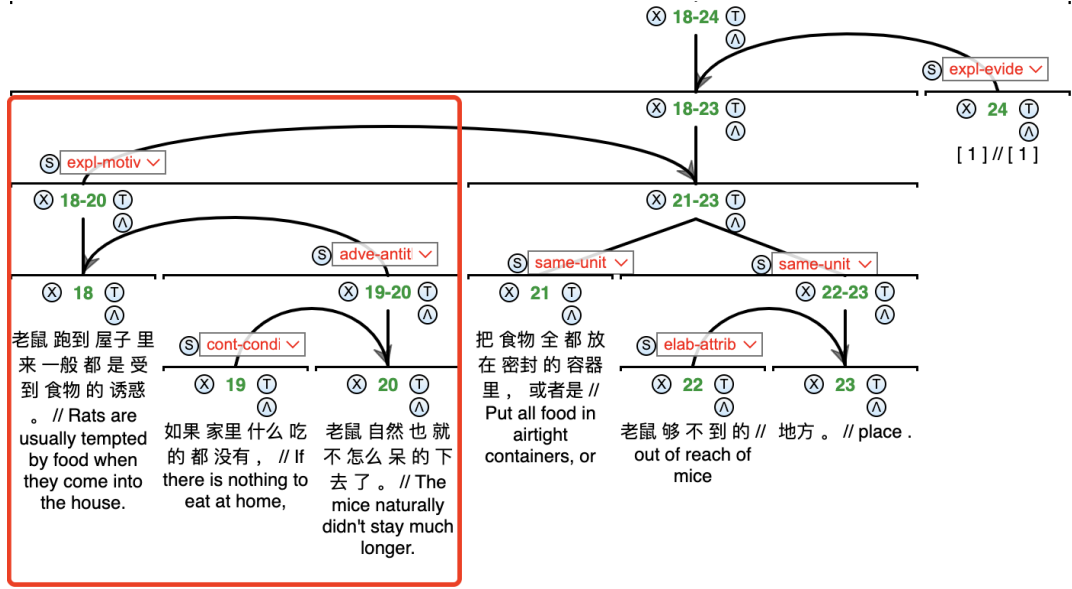

## 2.1.16 mode-manner

**mode-manner: the Satellite gives the manner of how the Nucleus happened.**

Based on Carlson 2003, "A manner satellite explains how something is done. (It sometimes also expresses some sort of similarity/comparison.) The satellite answers the question "in what manner?" or "in what way?".

**A MANNER relation is less "goal-oriented" than a MEANS relation, and often is more of a description of the style of an action."**

120.　　**EDU_22**　　和 线下 世界 一样 ，**// Like the offline world,**
　　　　EDU_23　　　　调情 的 第一 步 永远 是 打破 沉默 ，// The first step in flirting is constantly breaking the silence,
　　　　EDU_24　　　　和 对方 展开 交流 。// Communicate with each other.
　　　　source: gcdt_whow_flirt

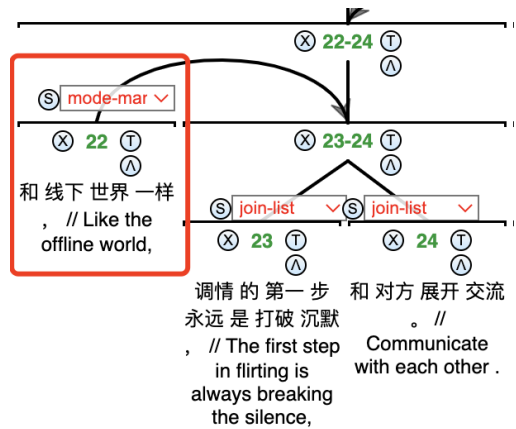

The following example, "taking a pseudonym" specifies how Higuchi publishes works.

121.　EDU_45　樋口 于 1891年 跟随 朝日 新闻 的 记者 半井桃水 // Higuchi following
Asahi Shimbun reporter Moomizu Banai in 1891
EDU_46　　　　学习 写作 技巧 ， // learn writing skills,
EDU_47　　　　并 在 同年 秋天 // and in the fall of the same year
**EDU_48　　　　取 " 一叶 " 为 笔名 // Take "Yi Ye" as a pseudonym**
EDU_49　　　　发表 作品 ， // published works,
source: gcdt_bio_higuchi

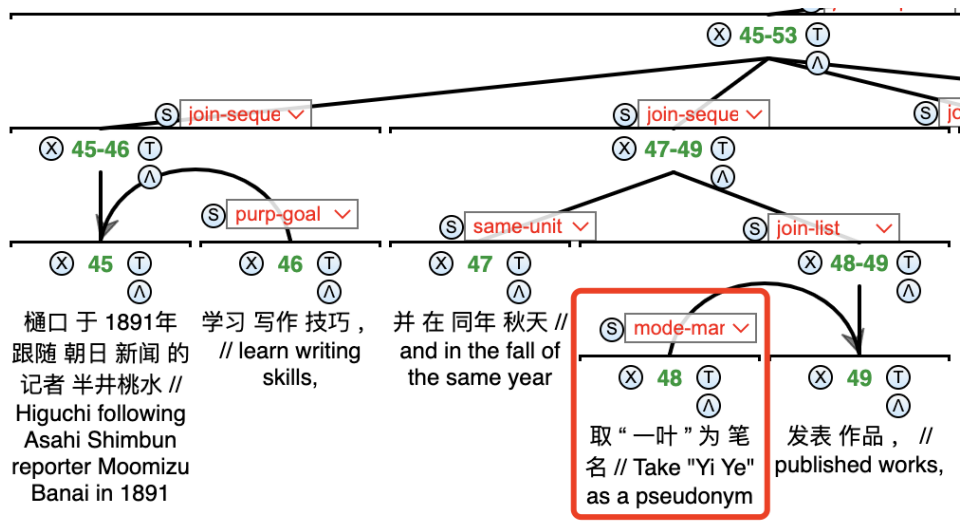

## 2.1.17 mode-means

**mode-means: the Satellite indicates means by which the Nucleus happened.**
According to Carlson 2003, "a means satellite specifies a **method, mechanism, instrument, channel or conduit for accomplishing some goal.** It should tell you how something was or is to be accomplished. In other words, the satellite answers a "by which means" or "how" question that can be assigned to the nucleus. **It is often indicated by the preposition by."**
In Chinese, means can be paraphrased as 用...的方式 (by means of …)
 in other words,  different from *manner*, *means* is when you cannot accomplish the nucleus without the method mentioned in the satellite.

122.　EDU_3　　你 打算 // are you going to
**EDU_4　　　　利用 简讯 、 微信 、 Whatsapp 等 即时 通讯 工具 // Use instant
messaging tools such as SMS, WeChat, Whatsapp**
EDU_5　　　　撩撩 妹 、 // flirting girl,
EDU_6　　　　调调 情 ， // flirt,
source: gcdt_whow_flirt

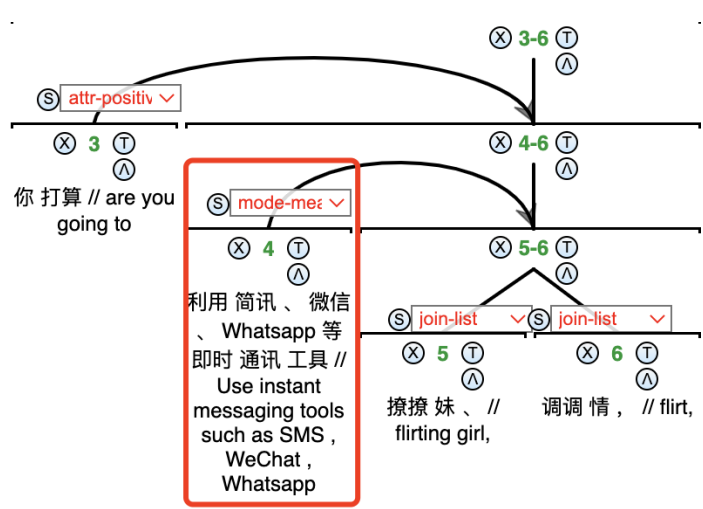

Similarly, "injecting vaccines" is the way to cure rabies.

123.   EDU_113   所以，狂犬病 // So, rabies
       **EDU_114**   **靠 接种 疫苗 // by injecting vaccines**
       EDU_115   是 可 预防 的 ， // is preventable,
       source: gcdt_academic_rabies

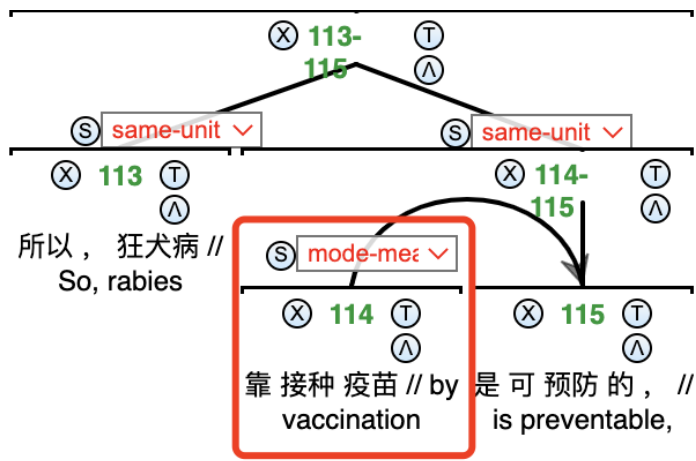

## 2.1.18 organization-heading

**organization-heading: the Satellite is graphically arranged to prepare for the Nucleus.**

Graphical traits can easily distinguish organization-heading satellites. They are usually headings, and there is a line break between the heading and the main content. These include document titles, sections, and subsection headings.

In the following example, "family background" and "life" are two section titles.

**124.   EDU_21   家世 // family background**

EDU_22     六世祖 赵翼 是 乾隆 二十六年 辛巳 恩科 进士 。 // The sixth ancestor, Zhao Yi, was a jinshi of Xin Si Enke in the twenty-sixth year of Qianlong's reign.

**EDU_23     生平 // life**

EDU_24     1892年 生于 直隶省 天津 ， // Born in 1892 in Tianjin, Zhili Province,

EDU_25     10 岁 前 随 做官 的 祖父 赵执治 辗转 居于 直隶省 各 地 ， // Zhao Zhizhi, his grandfather who became an official before the age of 10, lived in various parts of Zhili Province.

source: gcdt_bio_chao

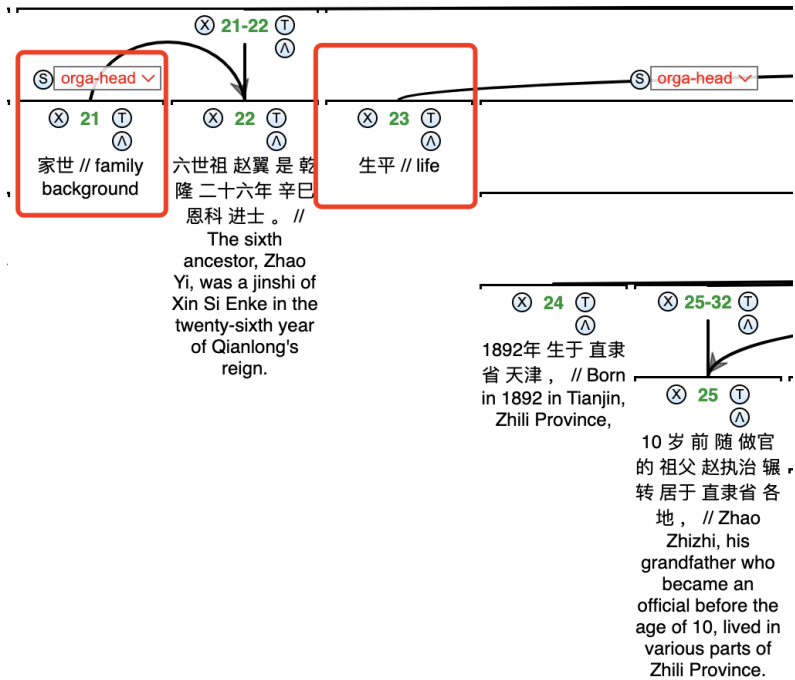

Similar parallelism of subsections are shown with different apple products in the following example.

**125.    EDU_141  iPhone 13 // iPhone 13**

EDU_142      售价 从 799 美元 // Priced from $799

EDU_143      （约 22,132 新台币 ， // (approximately NT$22,132,

EDU_144      台湾 官网 售价 25,900 元） // Taiwan official website price 25,900 yuan)

EDU_145      起跳 。// Take off.

**EDU_146      iPhone 13 Pro // iPhone 13 Pro**

EDU_147      售价 从 999 美元 // Priced from $999

EDU_148      （约 27,672 新台币 ， // (approximately NT$27,672,

EDU_149      台湾 官网 售价 32,900 元） // Taiwan official website price 32,900 yuan)

EDU_150      起跳 。// Take off.

source: gcdt_news_apple

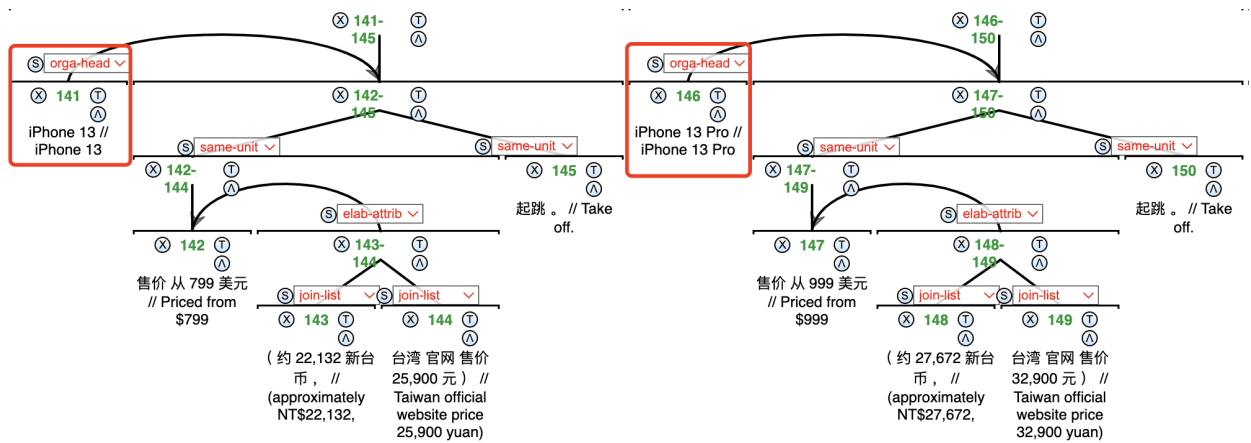

## 2.1.19 organization-phatic

**organization-phatic: the Satellite holds the floor for the Nucleus, with no semantic value.**
This label is usually applied to language disfluencies within a text.
In English, "see" and "you know" are examples of such phatic expressions. In Chinese, such examples include "这么说" (saying this way), "啥" (what), etc.

**126.    EDU_103  可以 说 // It can be said**

EDU_104      是 // it is

EDU_105      近来 改变 最 大 的 // that changed the most recently

EDU_106      iPad mini 系列 产品 。// iPad mini series products.

source: gcdt_news_apple

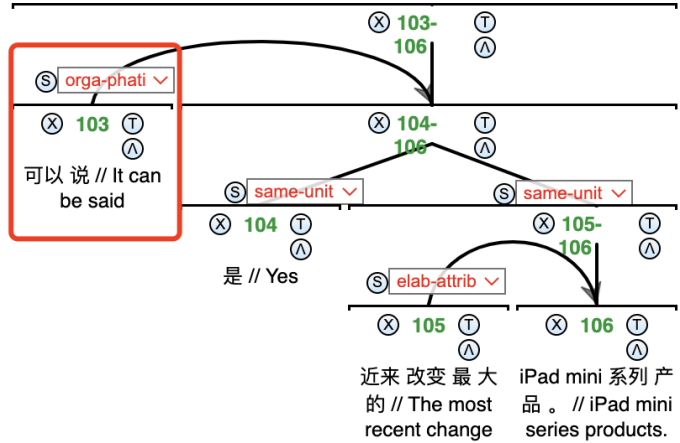

Organization-phatic is also used for self-corrections in speech. In the following example, 毕竟 (nevertheless) replaces 因为 (because), a more smooth connective between "they are wild animals" and "do not interact with them."

127.　EDU_294　永远 不 要 试图 接近 或者 跟 野生 动物 进行 互动 ，// Never try to approach or interact with wild animals,
**EDU_295**　　因为 // because
EDU_296　　　—— 毕竟 它们 是 野生 的 啊 。// - After all, they are wild.
source: gcdt_whow_hiking

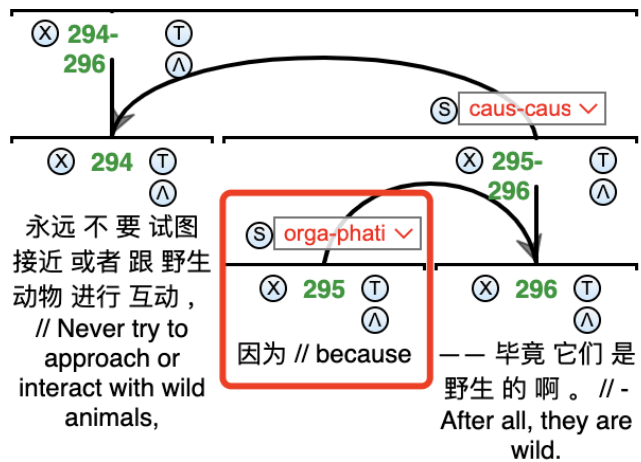

## 2.1.20 organization-preparation

**organization-preparation: the Satellite prepares the Reader for the Nucleus.**

Compared to context-background, organization-preparation contributes minimal information and simply serves the purpose of bridging discourse sections with a document.

We label those graphically not distinguished headers organization-preparation.

In the following example, "method 1" is an organization-preparation for "how to place eight balls" since there is no graphical disfluency between the two. Whereas they as a whole is the header of the following section so DU_19-20 functions as organization-heading.

128. **EDU_19** 方法 **1** **// method 1**
    EDU_20      8球 的 摆放 方法 // How to place 8 balls
    source: gcdt_whow_pool

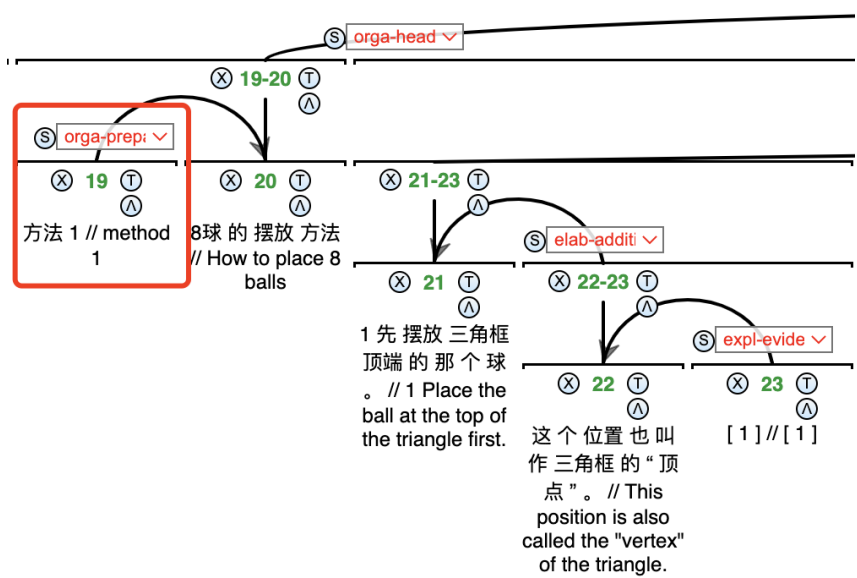

organization-preparation can also be found within the primary texts. For example, this relation is used to label the beginning of a document, section, or paragraph that continues from the preceding one.

In the following example, the targets of the genocide prepare for Ikhlov's claim that the intention is to eliminate rich peasants.

129.　EDU_81　伊赫洛夫 说 ： // Ikhlov says:

EDU_82　" 尽管 大 饥荒 中 饿死 的 以 乌克兰人 和 哈萨克人 居多 ， // **"Although most of the people who starved to death in the Great Famine were Ukrainians and Kazakhs,**

EDU_83　　但 那 场 民族 灭绝 行动 不 仅仅 针对 单一 民族 ， // **But that genocide wasn't just for a single people.**

EDU_84　　也 同样 针对 某 个 社会 阶层 ， // **also for a certain social class,**

EDU_85　　那 就 是 有意 // That is intentional

EDU_86　　利用 大 饥荒 // Take advantage of the Great Famine

EDU_87　　来 彻底 消灭 那些 比较 富裕 ， 很 独立 自主 的 农村 地区 中 的 农民 。 " // To completely eliminate the peasants in the more affluent and independent rural areas. "

source: gcdt_news_famine

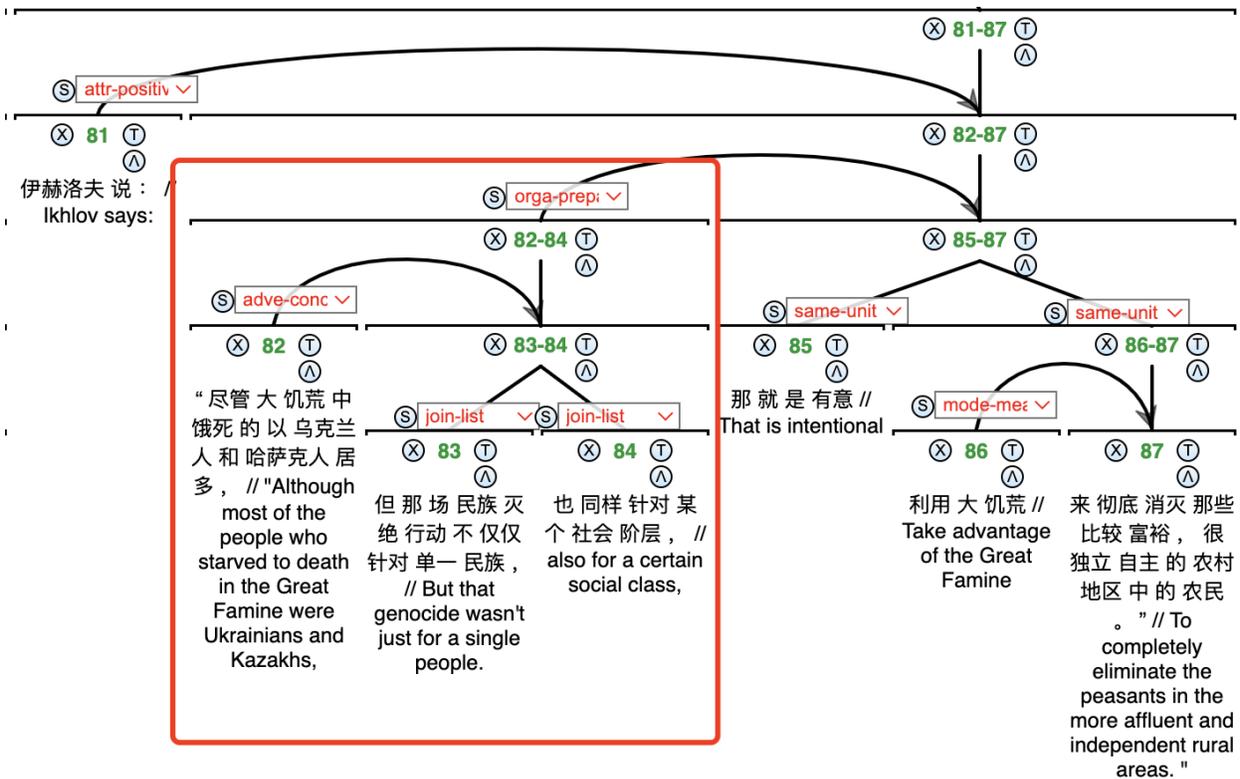

## 2.1.21 purpose-attribute

**purpose-attribute: only a part of the Nucleus (a phrase rather than the entire clause) occurs in order for the Satellite to happen.**

*Purpose-attribute* is the nominal-modifier counterpart of *purpose-goal,* just like *elaboration-attribute* for *elaboration-additional.*

In the following example, "the Greek War of Independence" is to resist Ottoman slavery.

130.    EDU_137  加入 到 了 // joined in

**EDU_138    反抗 奥斯曼 奴役 的 // against Ottoman slavery**

EDU_139        希腊 独立 战争 ， // Greek War of Independence,

source: gcdt_bio_byron

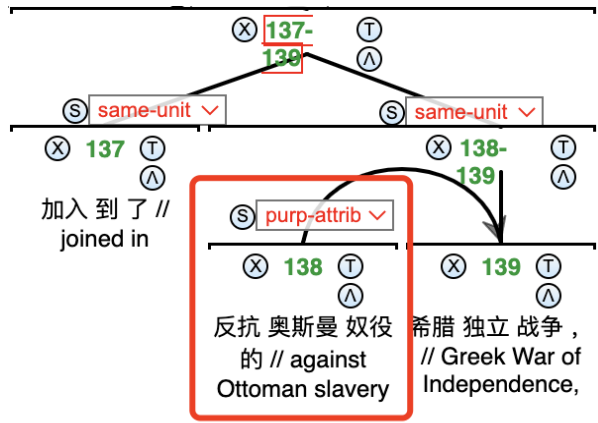

Similar to *elaboration-attribute*, there are also a few occurrences of *purpose-attribute* that are not relative clauses. In the following example, only the "extra clothing" is for preventing weather change.

131.    EDU_118  指南针 以及 地图 、 手电筒 、 火柴 或是 打火机 ， 以及 额外 的 衣物 // compass and map , flashlight , matches or lighters , and extra clothing

EDU_119        （ 以防 天气 突变 。 ） // (Just in case the weather changes suddenly.)

source: gcdt_whow_hiking

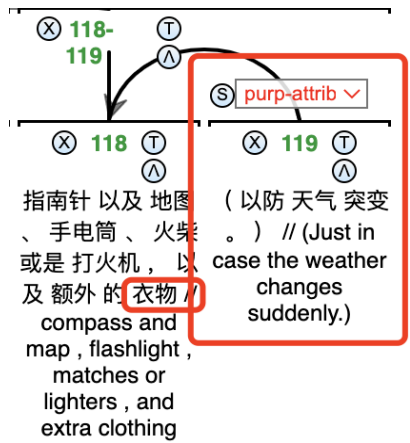

## 2.1.22 purpose-goal

**purpose-goal: the Nucleus occurs in order for the Satellite to happen.**
In the following example, the "$25" is for punishment.

132. EDU_228 并 须 支付 二十五 美元 入 皇家 财库 // and pay twenty-five dollars into the royal treasury
EDU_229 作为 惩罚 。 // as punishment .
source: gcdt_bio_emperor

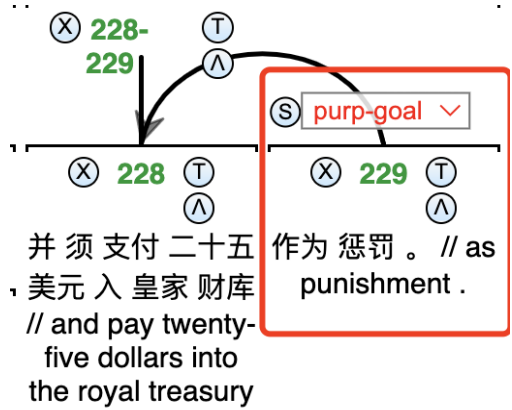

In the following example, "setting up a cordon" means reminding the spectators not to enter the area, not to hinder their rights, etc.

133. EDU_33 应当 有 设置 警戒线 ， // There should be a cordon,
**EDU_34** 提示 观赛者 // **Tips for spectators**
**EDU_35** 不 能 进入 比赛 区域 ， // **can not enter the competition area,**
**EDU_36** 妨碍 // **hinder**
**EDU_37** 参赛者 比赛 的 // **contestants of the competition**
**EDU_37** 权益 ， // **rights,**
**EDU_39** 否则 会 导致 // **Otherwise, it will cause**
**EDU_40** 比赛 不公 的 // **The game is unfair**
**EDU_41** 情况 ， // **Condition ,**
source: gcdt_interview_game

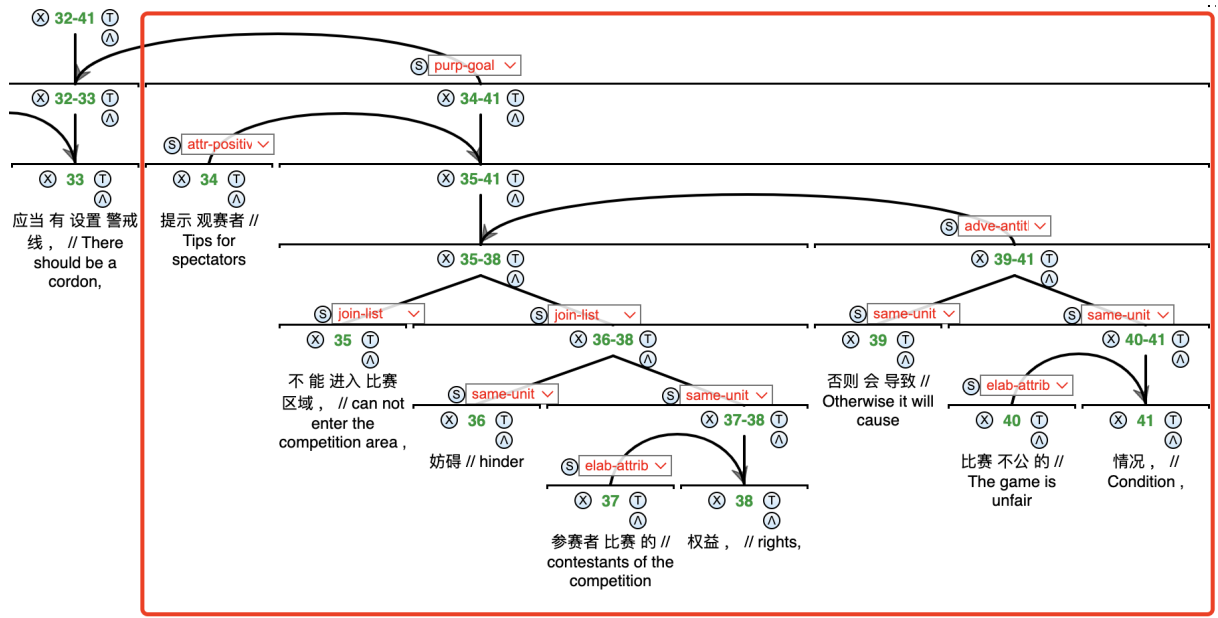

## 2.1.23 restatement-partial

**restatement-partial: the Satellite reiterates part of the Nucleus.**

If the relation is a complete repetition, please use the multinuclear relation:
restatement-repetition.

In the following example, "two balls are both solid or strip/half" is a repetition of "two balls of the
same kind" in EDU_46.

134. EDU_46 这么一来 ， 底边 两 角 的 两 个 球 就 变成 同样 花色 的 了 ， // In this
way, the two balls at the bottom corners will be of the same suit.

**EDU_47** 也 就 是 两 个 都 是 实色球 或 半色球 。 **// That is, both are
solid-colored or half-colored balls.**

source: gcdt_whow_pool

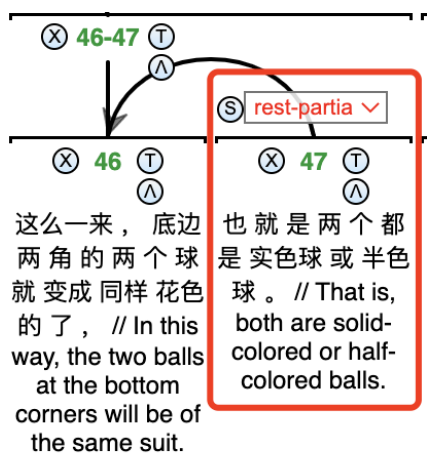

The following example includes two occurrences of *restatement-partial*: "14,700 USD" is a repetition of "108,000 SEK," and "not full but not starving" is a repetition of "minimum standard for not starving."

135.    EDU_86    筹款 带来 了 10.8万 瑞典 克朗 // Fundraising brought in 108,000 SEK
        **EDU_87    （编者 注 ： // (Editor's note:**
        **EDU_88    约 14700 美元 ） ， // about $14,700),**
        EDU_89    足够 购买 300万 张 选票 ， // enough to buy 3 million votes,
        EDU_90    这 是 某 种 // this is some kind of
        EDU_91    至少 我 们 不 会 挨饿 的 // At least we won't starve
        EDU_92    最低 标准 。 // Minimum Standards．
        **EDU_93    我们 没有 吃饱 ， // We are not full,**
        **EDU_94    但 也 没有 挨饿 。 // But also not starving.**
        source: gcdt_interview_falkvinge

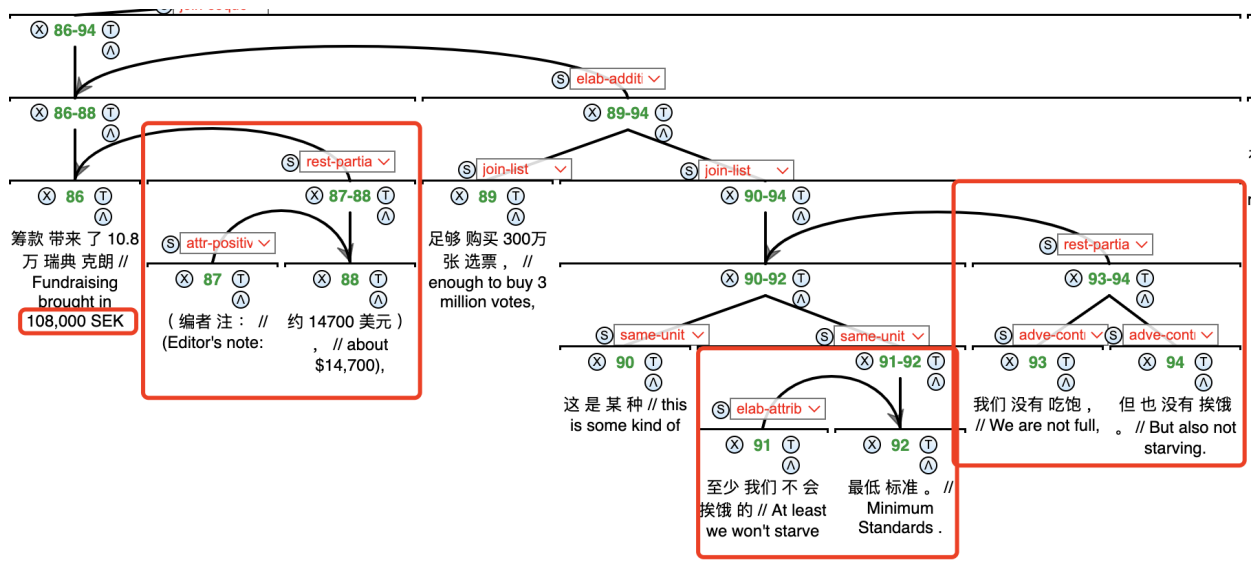

Note: **we draw the following distinctions between *restatement* versus *elaboration.***
In general, restatement does not provide additional knowledge and is interpreted as equivalent given the context (as well as world knowledge).

The followings are *restatements:*
- "Today" ← "(May 11)"
- Synonym of the same entity in the language
- when the latter part can be conducted from the former, "the number went from 50 to 40" ← "it decreased by 10."

On the contrary, the followings are *elaborations*:
- The same phrases in different languages (i.e., translations) are not restatements; e.g., "Apple" ← "( German: Apfel )."

## 2.1.24 topic-question

**topic-question: the Satellite requests the information in the Nucleus.**
In the following example, the question satellite asks for what "you will see during this time."

136.  **EDU_48**  这 期间 你 会 看到 什么 ？ **// What will you see during this time?**
      EDU_49      荧光 。 // Fluorescence.
      source: gcdt_whow_glowstick

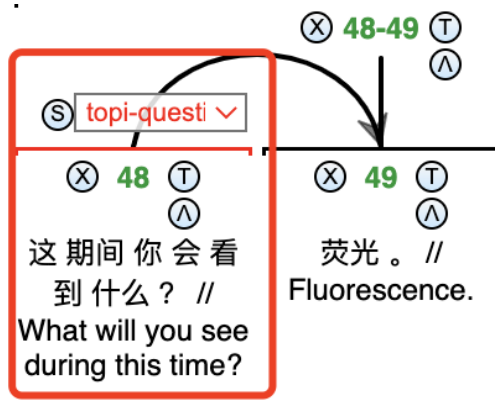

In the following example, we see two question-answer pairs. DU_146-147 asks for the interviewee's experience, and EDU_148 is a clarification question regarding the "Tour Taiwan Tournament."

137.  **EDU_146** 对于 今年 让 环台赛 、 自行车展 、 体育 用品展 三合一 的 **// For this year's trip to make the Tour of Taiwan, the bicycle show and the sporting goods show**
      **EDU_147**      看法 ？ **// view?**
      **EDU_148**      环台赛 ？ **// Tour Taiwan Tournament?**
      EDU_149      是 一 项 自由车 的 竞技 。 // It's a free bike competition.
      EDU_150      这 种 整合 不仅 恰到好处 ， // This integration is not only just right,
      EDU_151      甚至 对于 会展 的 人气 或 买气 ， 都 有 一定 程度的 提升 。 // Even the popularity or buying interest of the exhibition has increased to a certain extent.
      source: gcdt_interview_ideal

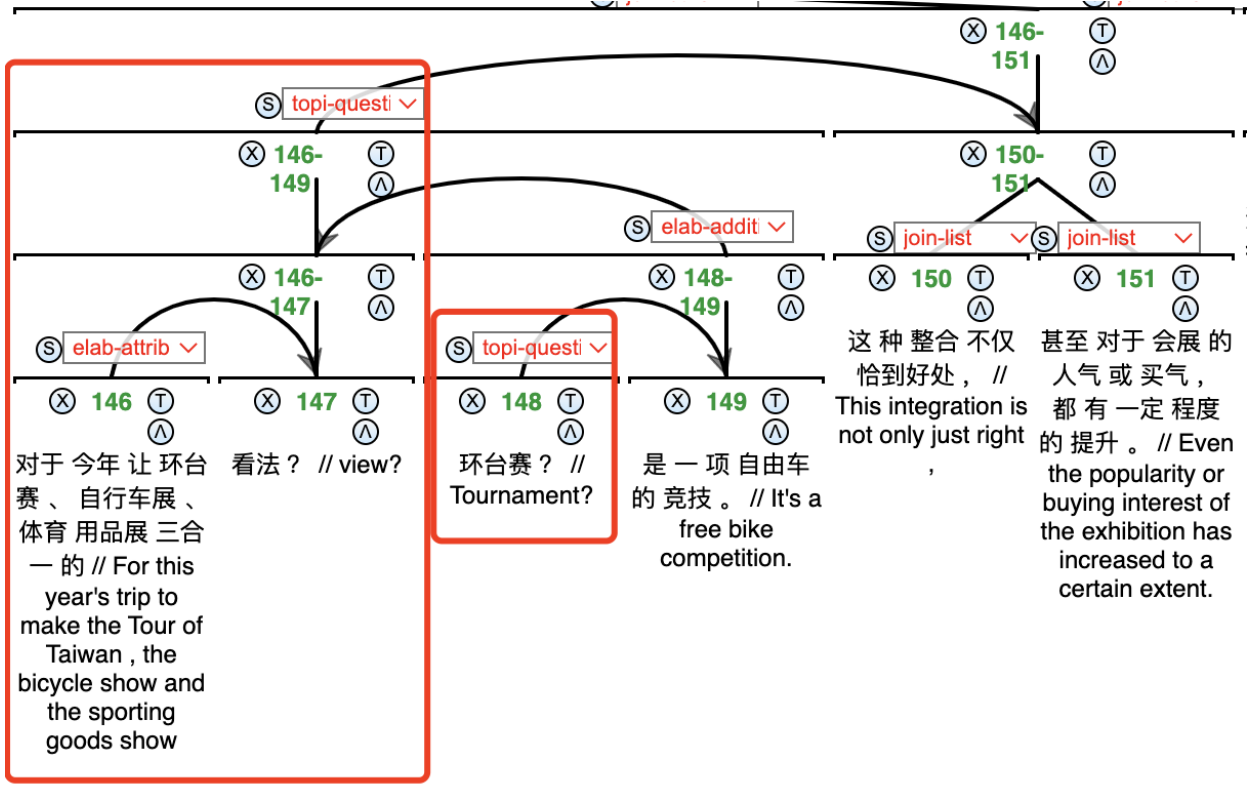

## 2.1.25 topic-solutionhood

**topic-solutionhood: the Nucleus is the answer to a problem in the Satellite.**
In other words, the Satellite poses a problem, and the nucleus presents a solution.
In the following example, "planning" is the solution to "preventing attacks."

**138.**   EDU_71   资讯 安全 ， 有 无意 与 恶意 的 攻击者 ， 要 怎么 去 阻止 ， //
**Information security, how to stop unintentional and malicious attackers,**
EDU_72     就 要 先 一 步 地 进行 规划 ， // It is necessary to plan ahead.
EDU_73     才 能 适当 地 完善 其 防护 措施 ， // In order to properly improve its
protective measures,
source: gcdt_interview_wimax

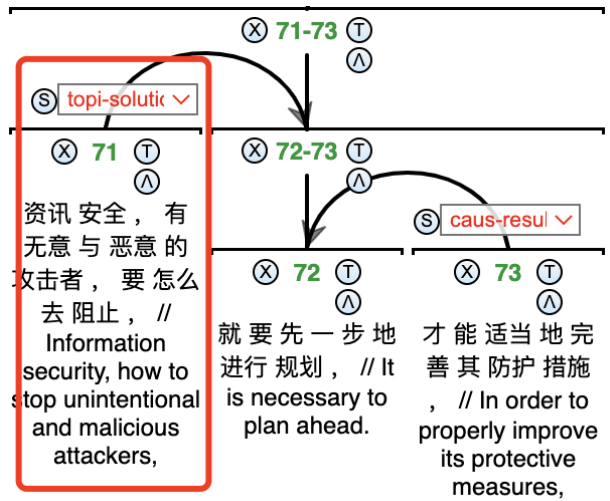

topic-solutionhood is a relatively infrequent one in Chinese. Similarly, *elaboration-additional* occurs ~45 times more frequently than *topic-solutionhood* in GUM.

## 2.2 Multinuclear relations

### 2.2.1 adversative-contrast

**adversative-contrast: the Writer presents similar units with contrast.**
Compared to *adversative-concession or adversative-antithesis*, *adversative-contrast* is multinuclear meaning that the contrastive units are equally important.

In the following example, "being contagious" and "unclear susceptibility" is adversarial to each other but of equal importance.

**139.    EDU_56**   这 证明 了 **// this proves**
**EDU_57**         该 疾病 具有 传染性 ， **// The disease is contagious ,**
**EDU_58**         但 物种 敏感性 还 不 清楚 ， **// But species susceptibility is not yet known ,**
source: gcdt_academic_rabies

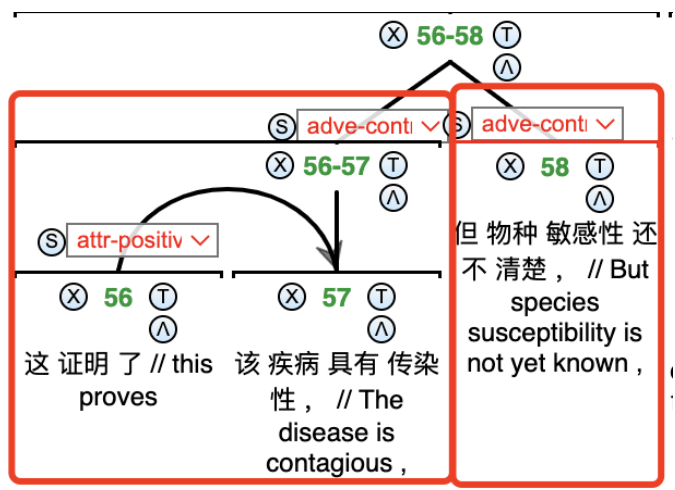

In the following example, we see a contrast between two larger DUs, DU_118-127 "DEV that has less severe" versus DU_128-129 "DEV that is not as immunogenic to prevent rabies."

**140.    EDU_118** 严重 反应 较 少 的 **// less severe reaction**
**EDU_119**       另 一 个 疫苗 是 鸭胚 疫苗 **// Another vaccine is the duck embryo vaccine**
**EDU_120**       **( duck embryo vaccine // (duck embryo vaccine**
**EDU_121**       简称 **DEV )** ， **// DEV for short),**
**EDU_122**       该 疫苗 **// the vaccine**
**EDU_123**       通过 **// pass**
**EDU_124**       在 受孕 鸭蛋 里 传播 的 **// Spread in the egg of a pregnant duck**
**EDU_125**       病毒 **// Virus**
**EDU_126**       制备 **// preparation**

**EDU_127**     [ 7 ] [ 8 ] 。 // **[7][8].**
**EDU_128**     然而，**DEV** 不如 脑 组织 疫苗 的 免疫 原性 强 ， // **However, DEV is not as immunogenic as brain tissue vaccines,**
**EDU_129**     并不 总是 能 预防 狂犬病 。 // **Rabies is not always prevented.**
source: gcdt_academic_rabies

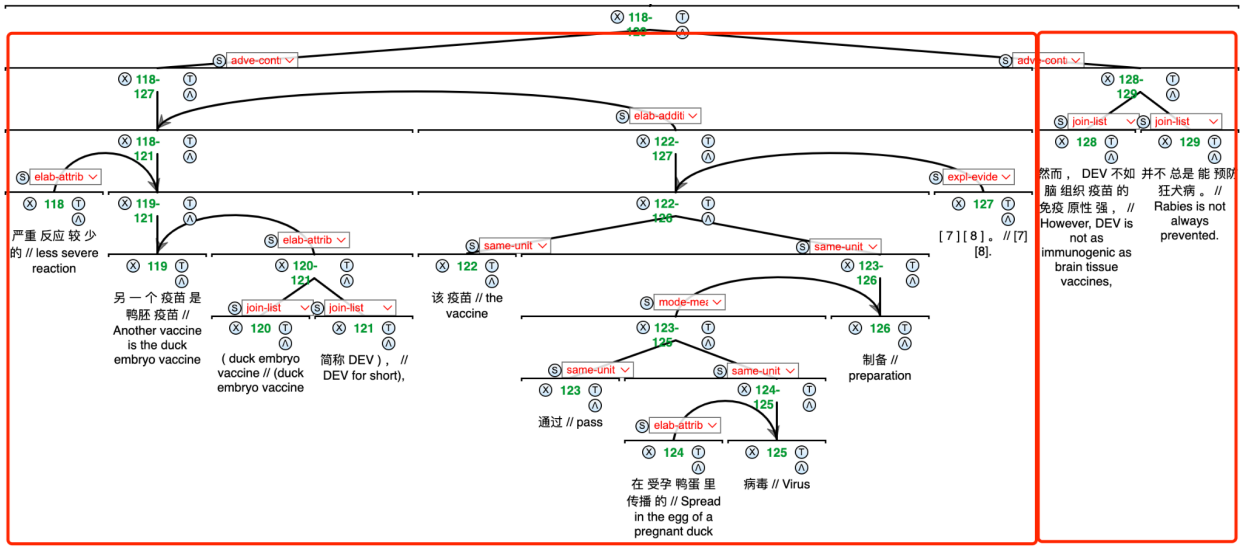

## 2.2.2 joint-disjunction

**joint-disjunction: the Writer presents a set of alternatives.**
Different from *joint-list*, the set of alternatives are in complementary distribution where I should choose one among them.

In the following example, the hiking trail is either a back-and-forth trail or a loop. Thus, the hiking distance is either twice the distance from the start to the end of the distance of the loop.

141.    EDU_200  —— 这 个 标志 通常 能 表示 出 路名 以及 // - This sign usually indicates the name of the way out and
     **EDU_201**     到 尽头 的 // **to the end**
     **EDU_202**     距离 // **distance**
     **EDU_203**     （ 或是 环路 的 长度 。 ） // **(Or the length of the loop.)**
     source: gcdt_whow_hiking

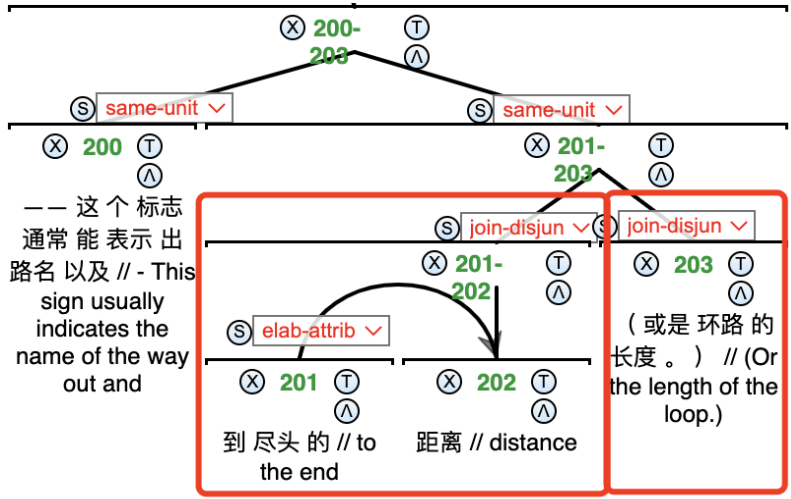

The extended example presents two alternatives: "go to your desired place according to road sign" or "according to map and rockfill landmarks if there is no sign."

142. **EDU_223** 请 **// Please**
    **EDU_224** 依照 **// according to**
    **EDU_225** 路标 上 标 的 **// marked on a road sign**
    **EDU_226** 路名 **// road name**
    **EDU_227** 前往 **// go to**
    **EDU_228** 你 要 去 的 **// you are going**
    **EDU_229** 地方 。 **// place .**
    **EDU_230** 如果 没有 路标 **// if there are no road signs**
    **EDU_231** （ 这 是 比较 罕见 的 情况 ） ， **// (this is a relatively rare case),**
    **EDU_232** 在 地图 上 查 一下 ， **// Check it out on the map,**
    **EDU_233** 或是 找找 周围 有 没有 堆石 界标 。 **// Or look around for rockfill landmarks.**
    source: gcdt_whow_hiking

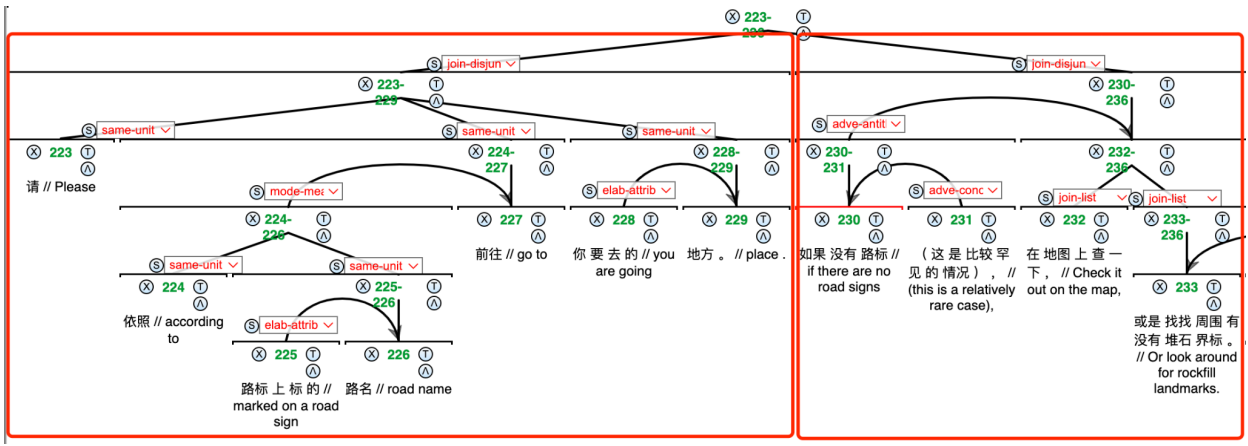

### 2.2.3 joint-list

**joint-list: the Writer presents coordinated and similar units.**
The following scenarios are typical instances of *joint-list*:
- listed coordinating conjunctions within a sentence
- enumerations such as:
    - Method 1, Method 2
    - Section numbers: 2.1, 2.2

In the following example, "freeing," "abandoning and transcending," and "starting" is the significance of the change in Marx's way of thinking.

143.　　EDU_11　　这 一 论断 表明 // This assertion shows that
　　　　EDU_12　　马克思 的 思维 方式 由 实体性 转变 为 关系性 ， // Marx 's way of thinking changed from substantive to relational ,
**　　　　EDU_13　　摆脱 了 旧 唯物 主义 的 影响 ， // Freed from the influence of old materialism,**
**　　　　EDU_14　　扬弃 和 超越 了 以往 哲学家 对 人 与 社会 关系 孤立式 的 理解 方式 ， // Abandoning and transcending the previous philosophers' isolated understanding of the relationship between man and society,**
**　　　　EDU_15　　开始 以 崭新 的 姿态 面向 人类 社会 。 // Begin to face human society with a new attitude.**
　　　　source: gcdt_academic_socialized

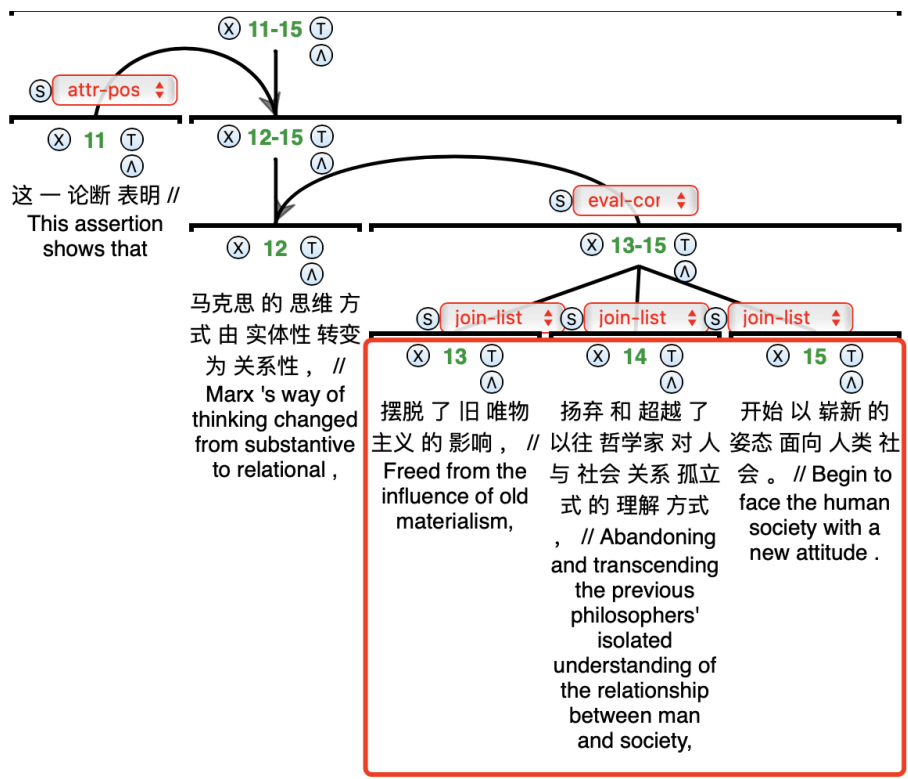

In the following example, we see three groups of *joint-list* coordinations at different levels:

- between DU_42-46 "socialized human being revealing" and DU_47-52 "ideal of comunity answering … and providing …."
- among EDU_43 "taking people's …", EDU_44 "economy's …" and EDU_45 "ideas'..."
- between DU_47-51 "answering" and EDU_52 "providing."

**144.** **EDU_42** " 社会化 的 人类 " 概念 就 像 一 把 哲学 钥匙 ， **// The concept of "socialized human being" is like a philosophical key,**

**EDU_43** 以 " 人 的 社会化 " 为 逻辑 支点 ， **// Taking "people's socialization" as the logical fulcrum,**

**EDU_44** 以 " 经济 的 社会化 " 为 思想 内核 ， **// Taking " socialization of the economy " as the core of thought,**

**EDU_45** 以 " 观念 的 社会化 " 为 价值 诉求 ， **// Taking "the socialization of ideas" as the value appeal,**

**EDU_46** 揭示 了 人类 社会 发展 的 必然 趋势 和 追求 目标 。 **// It reveals the inevitable trend and pursuit of human society development.**

**EDU_47** 人类 命运 共同体 思想 作为 新 时代 重大 的 理论 创新 成果 ， **// The idea of a community with a shared future for mankind is a major theoretical innovation in the new era.**

**EDU_48** 以 " 社会化 的 人类 " 概念 作为 思想 来源 ， **// Using the concept of " socialized human beings " as a source of thought,**

**EDU_49** 向 全 世界 解答 了 **// Answered to the world**

**EDU_50** 人类 命运 共同体 何以 可行 的 **// How is a community with a shared future for mankind feasible**

**EDU_51** 问题 ， **// question ,**

**EDU_52** 为 人类 命运 的 总体 发展 提供 了 全新 的 价值 理念 。 **// It provides a new value concept for the overall development of human destiny.**

source: gcdt_academic_socialized

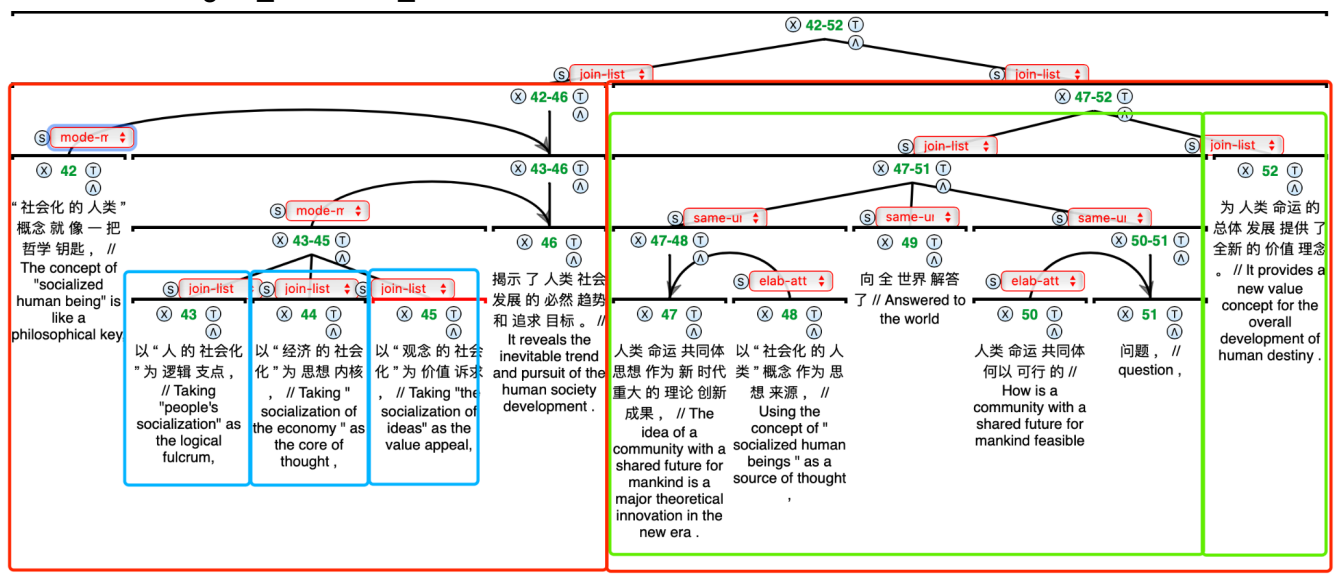

## 2.2.4 joint-sequence

**joint-sequence: the Writer presents EDUs of chronological sequence.**

Coordinating conjunctions that follow each other in a time sequence. For example:

- coordinations signaled by date or time
- section titles ordered by a person's growing to death or an event's preparation to completion

The following example DU_16-18 shows symptoms of rabies from the beginning to death.

145.   EDU_14   狂犬病 影响 大脑 和 脊髓 // Rabies affects the brain and spinal cord
   EDU_15   （中枢 神经 系统），// ( Central Nervous System ) ,
   **EDU_16   初始 症状 类似 流感 、 发烧 、 头痛 ， // Initial symptoms are flu-like, fever, headache,**
   **EDU_17   但是 感染 可以 快速 发展 到 幻觉 、 瘫痪 ， // But infections can progress quickly to hallucinations , paralysis ,**
   **EDU_18   并 最终 死亡 ， // and eventually died,**
   source: gcdt_academic_rabies

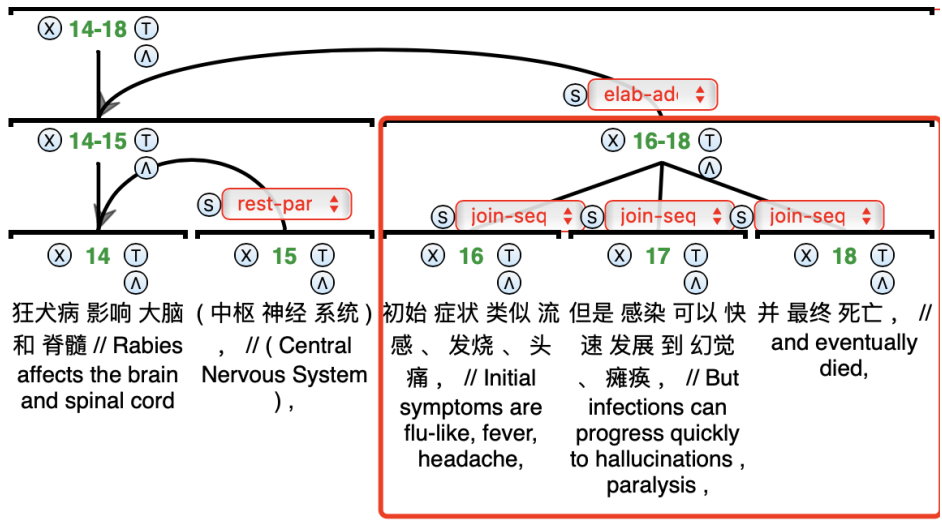

In the following example, we see three groups of *joint-sequence*:

- Among "Brighton High School," "Bachelor in Suffolk University," and "Master in Boston University"
- between "moving to Boston" and "studying at Suffolk University."
- between "transferring to Boston University" and "obtaining master's degree."

146.   EDU_32   她 中学 就读于 布莱顿 高中 ，// She attended Brighton High School in secondary school,

EDU_33   2004年 毕业 后 ，// After graduating in 2004,

EDU_34   穆雷 搬到 波士顿 // Murray moves to Boston

EDU_35   就读 萨福克 大学 的 心理学系 // Studied psychology at Suffolk University

EDU_36   [ 8 ] 。// [ 8 ] .

EDU_37   在 获得 理学士 学位 后 ，// After graduating with a Bachelor of Science degree,

EDU_38   她 转读 波士顿 大学 的 体育 心理学系 // She transferred to Boston University's sports psychology department.

EDU_39   并 获得 硕士 学位 // and obtained a master's degree

EDU_40   [ 9 ] 。// [ 9 ] .

source: gcdt_bio_marbles

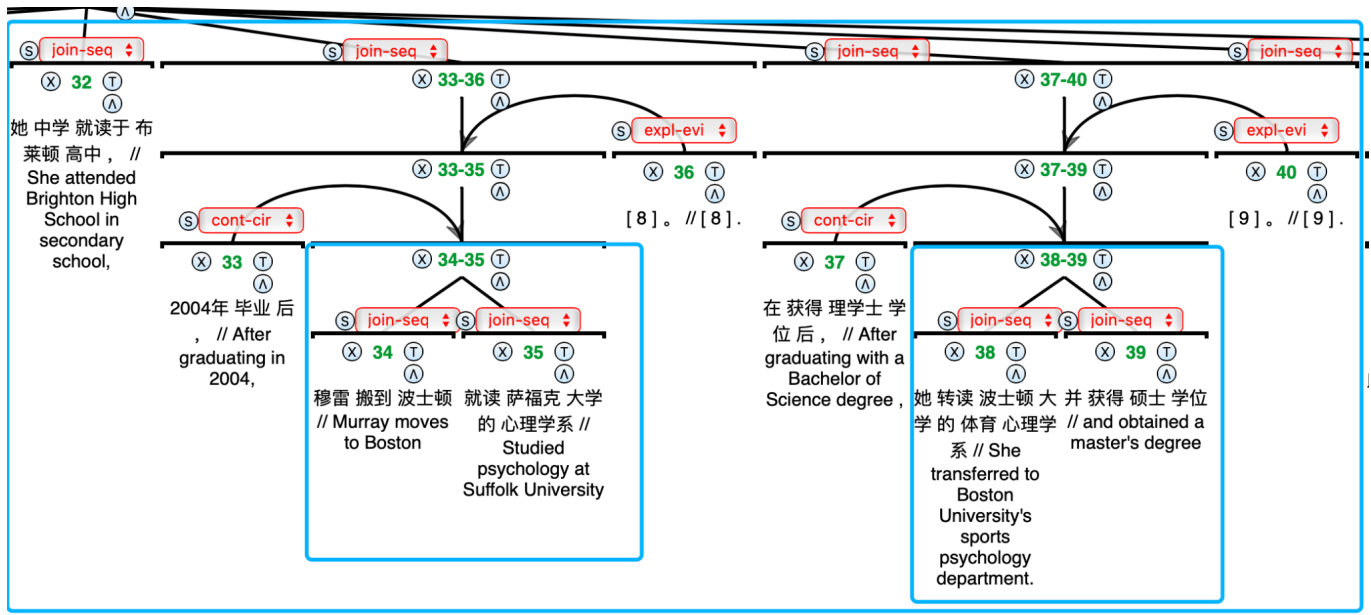

## 2.2.5 joint-other

**joint-other: the Writer presents unlike units with no other relation.**
This is the last resort for multinuclear relations. We use this joint-other label when conjoined elements are neither listed nor of sequential order. For example:
- section and subsection heads without enumeration or time sequence

*Joint-other* usually occurs between larger discourse units. Due to the length of this larger DUs, it is difficult to present a complete example. The following is a snippet in a sequence of *joint-other* DUs, and here, we show a *joint-other* relation between DU_14-21 and DU_22-30 (only the nucleus part in the snippet).

147.  **EDU_14**　　狂犬病 影响 大脑 和 脊髓 **// Rabies affects the brain and spinal cord**

　　**EDU_15**　　（ 中枢 神经 系统 ） ， **// ( Central Nervous System ) ,**

　　**EDU_16**　　初始 症状 类似 流感 、 发烧 、 头痛 ， **// Initial symptoms are flu-like, fever, headache,**

　　**EDU_17**　　但是 感染 可以 快速 发展 到 幻觉 、 瘫痪 ， **// But infections can progress quickly to hallucinations , paralysis ,**

　　**EDU_18**　　并 最终 死亡 ， **// and eventually died,**

　　**EDU_19**　　目前 没有 有效 的 疗法 。 **// There is currently no effective treatment .**

　　**EDU_20**　　一旦 发生 疾病 ， **// Once a disease occurs,**

　　**EDU_21**　　死亡率 近 **100%** 。 **// Mortality is nearly 100%.**

　　**EDU_22**　　狂犬病 在 **150** 多 个 国家 和 地区 都 有 存在 。 **// Rabies is present in more than 150 countries and territories .**

　　**EDU_23**　　据 世卫 组织 估计 ， **// According to WHO estimates,**

　　**EDU_24**　　每 年 有 近 **55,000** 人 死于 狂犬病 。 **// Nearly 55,000 people die from rabies each year.**

source: gcdt_bio_marbles

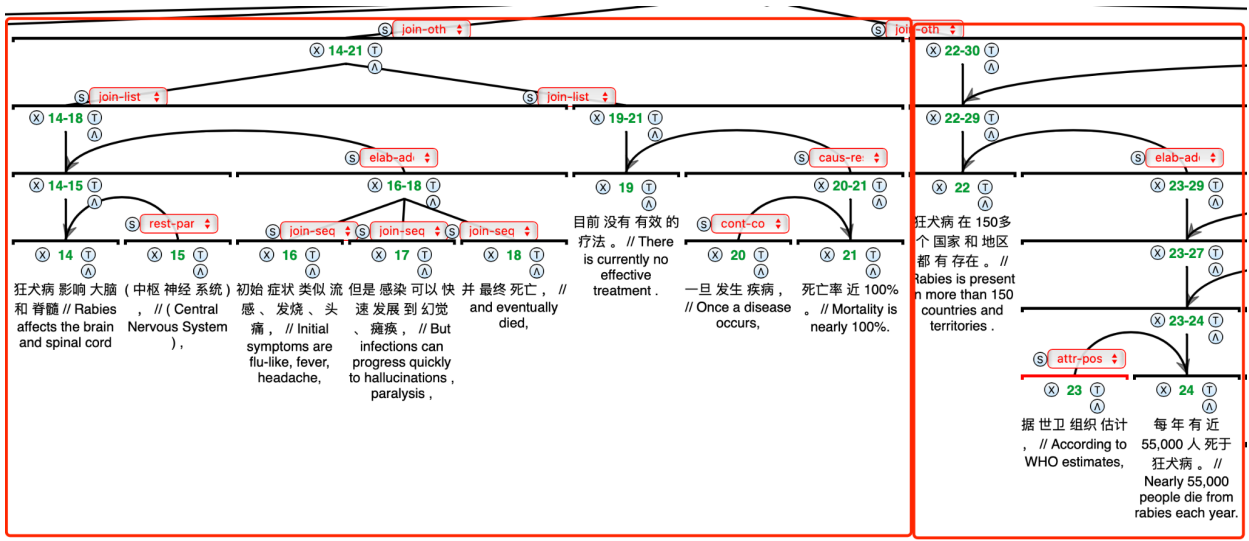

## 2.2.6 restatement-repetition

**restatement-repetition: the Writer presents equivalent or redundant units.**

In the following example, EDU_81 is a modern Chinese paraphrase of the ancient Chinese quote EDU_82.

148.   **EDU_81**   第一，由 精气 直接 生出 万事 万物 。// **First, all things are born directly from the essence.**

   **EDU_82**   " 精微者 天地 之 始 也 。" // **"The subtle is the beginning of heaven and earth."**

   EDU_83        (《 鹖冠子·泰录 》 ) // ( "Heguanzi·Tailu")

   source: gcdt_academic_taoist

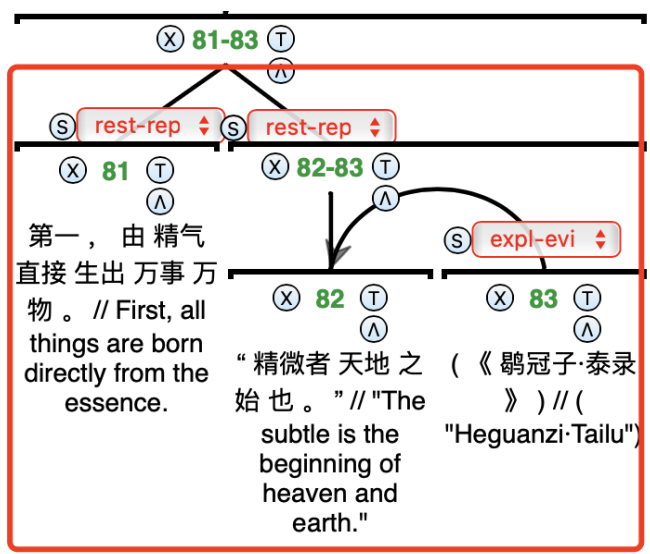

The following example shows a repetition of "rare to see People Daily's comments" between DU_141-143 and EDU_144.

149.   EDU_140  2.2.2. 很少 回复 大众 评论 // 2.2.2. Few replies to public comments

   **EDU_141**        在 人民 日报 的 评论区 ， 很少 看到 人民 日报 的 评论 回应 。 // **In the comments section of the people's daily, it is rare to see the people's daily commentary response.**

   **EDU_142**        很多 视频 的 第一 条 热评 来源于 各 地区 的 消防 公安 政务号 ， // **The first hot comment of many videos comes from the fire, public security and government affairs accounts of various regions.**

   **EDU_143**        普通 大众 的 评论 大部分 都 在 每 条 视频 的 第二 条 评论 之后 。 // **Most of the comments from the general public come after the second comment on each video.**

   **EDU_144**        评论区 很 难 见到 人民 日报 的 评论 回复 。 // **It's hard to see replies to comments from People's Daily in the comment area.**

   source: gcdt_academic_peoples

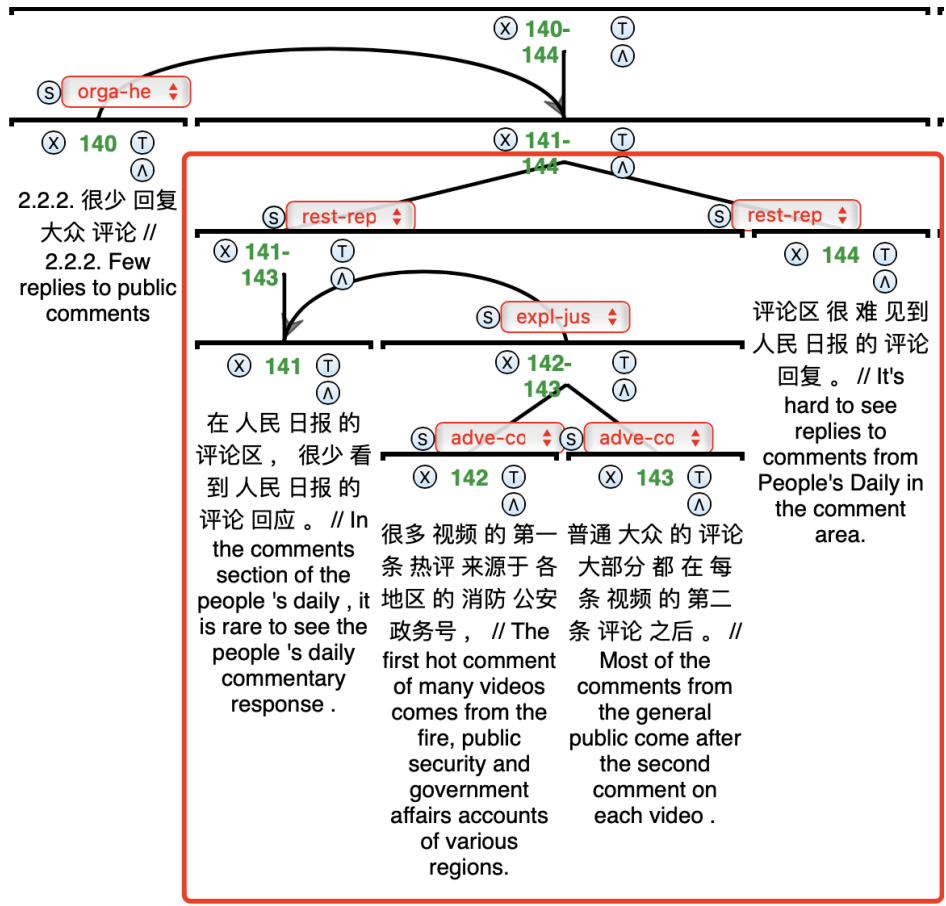

## 2.3 same-unit

**same-unit: this is a technical device for interrupted EDUs.**

In the following example, we see a single EDU broken up by five parenthetical abbreviations, forming a series of *elaboration-attribute + same-unit* relations.

150.    EDU_37    这些 基因 编码 核 蛋白 // These genes encode nuclear proteins
       EDU_38    （ N ）、 // (N),
       EDU_39    磷 蛋白质 // phosphoprotein
       EDU_40    （ P ）、 // (P),
       EDU_41    基质 蛋白 // matrix protein
       EDU_42    （ M ）、 // (M),
       EDU_43    糖蛋白 // glycoprotein
       EDU_44    （ G ）// ( G )
       EDU_45    和 病毒 RNA 聚合 酶 // and viral RNA polymerase
       EDU_46    （ L ）、 // (L),
      source: gcdt_academic_rabies

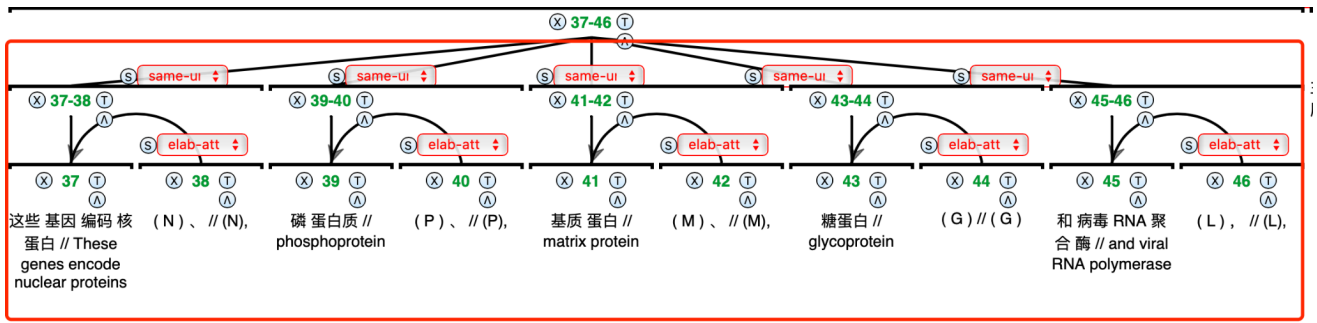

Sometimes, due to the structure of modifications, it is more reasonable to have hierarchical *same-unit* groups than a one-level *same-unit* grouping.

In the following example, we see two levels of same-unit due to the *attribution-positive* to DU_70-72 and parenthetical repetitions of EDU_70.

151. EDU_68　这 说法 // this statement ||
EDU_69　和 罗拔·高云 （ 1923年 ） 声称 // and Robert Gowan (1923) claimed ||
EDU_70　" 那 个 皇帝 诺顿 一世 拥有 希伯来 // "The Emperor Norton I had the Hebrew ||
EDU_71　（ 犹太 ） // (Jewish) ||
EDU_72　血统 " // pedigree" ||
EDU_73　相 吻合 。 // match. ||
source: gcdt_bio_emperor

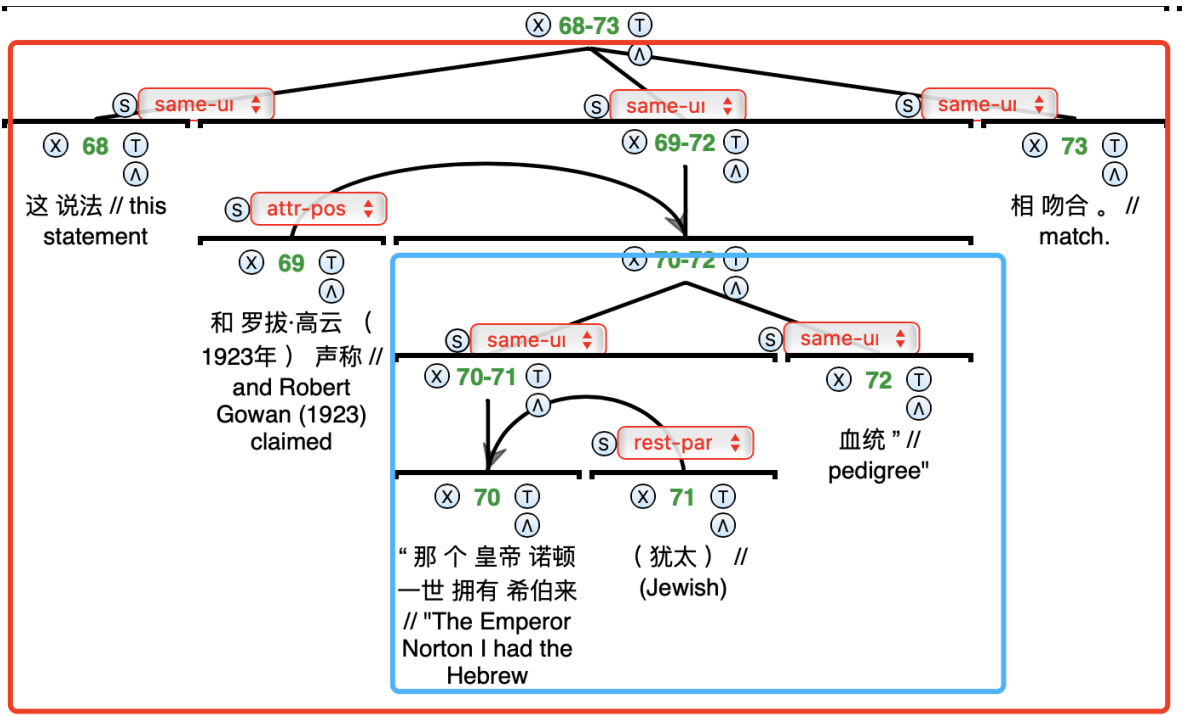

# 3 Deciding between possible relations

## 3.1 NS vs. SN: heading versus content

This conflict is particularly frequent in the how-to guide (whow) genre.
Compare the following examples:

When the content is a sequence of actions or items that make up the heading together, then the nucleus is the sequence of actions;

152.　EDU_58　3 准备 好 播种 的 容器 和 土壤 。 // 3 Prepare the container and soil for sowing.

EDU_59　　在 方盆 或是 单独 的 容器 里 填上 同等 分量 的 蛭石 、 珍珠岩 和 泥炭土 。 // Fill a square pot or separate container with equal parts vermiculite, perlite, and peat.

EDU_60　　轻压 土壤 // lightly press the soil

EDU_61　　来 排除 空气 。 // to remove air.

EDU_62　　把 土壤 用 水 打湿 ， // Wet the soil with water,

EDU_63　　为 种子 发芽 // germinate for seeds

EDU_64　　准备 好 合适 的 环境 。 // Get the right environment ready.

source: gcdt_whow_basil

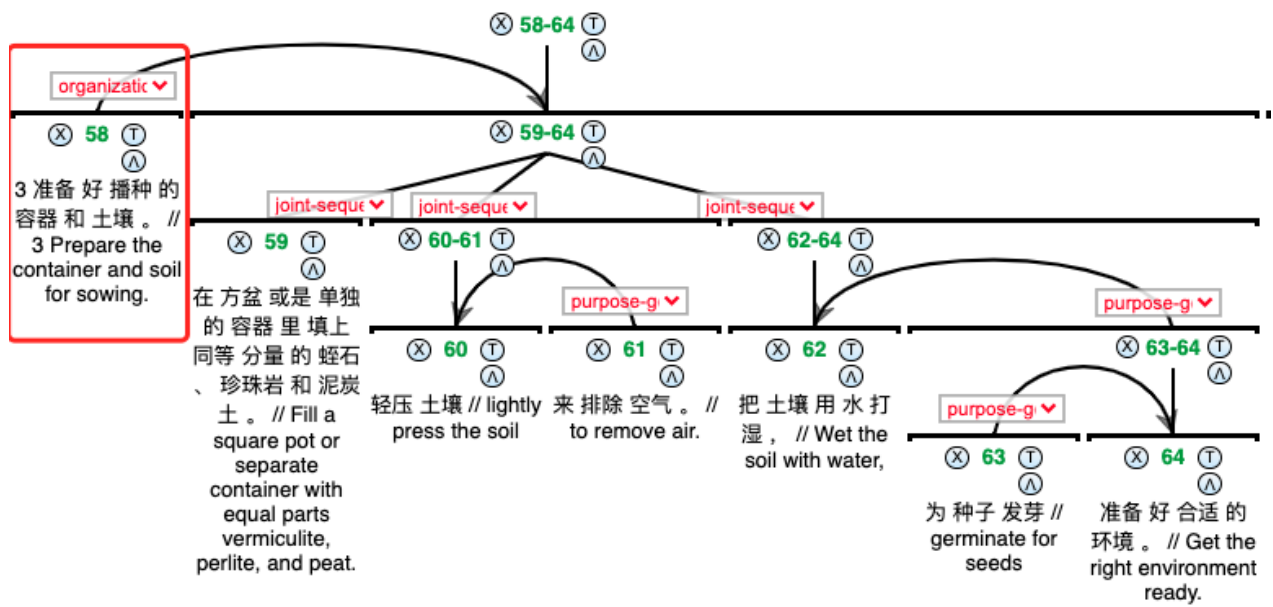

However, when the content only gives further details or explanations of the heading but does not repeat it anywhere, the heading is the nucleus.

In the following example, "enjoying fresh basil" is not conveyed explicitly in the content.

153.　　EDU_176　2 享受 新鲜 的 罗勒 。 // 2 Enjoy the fresh basil.
EDU_177　　　把 罗勒 洗 干净 后 ， // After washing the basil,
EDU_178　　　就 可以 用 它们 做 绿酱 ， // You can use them to make green sauce,
EDU_179　　　或是 用 西红柿 和 马苏里 芝士 做成 番茄 芝士 沙拉 。 // Or make a tomato cheese salad with tomatoes and mozzarella.
source: gcdt_whow_basil

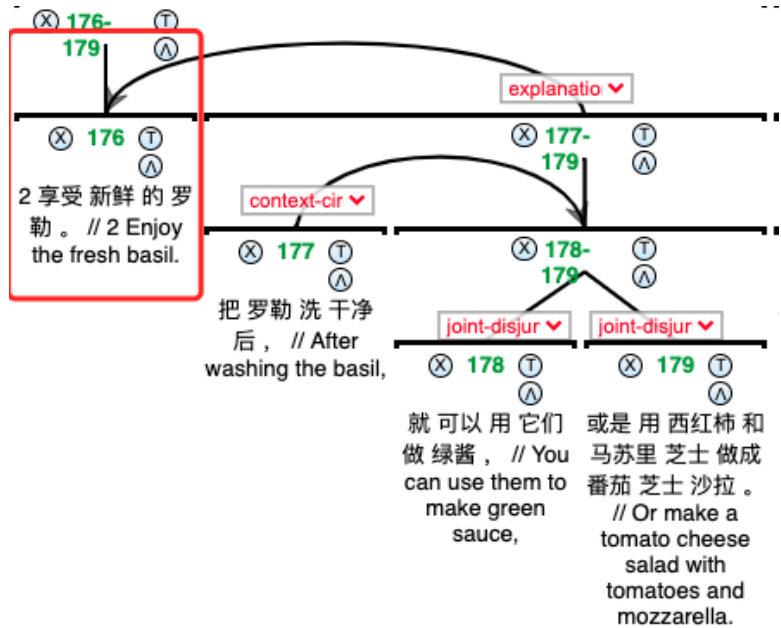

## 3.2 Motivation versus cause

Motivation is when the situation in the satellite motivates the reader to make actions in the nucleus.

In the following example, the disadvantages of quiet rooms or songs with lyrics do not cause listening to white noise or pure music. Instead, the writer tries to inform the reader that he/she should listen to white noises based on the drawbacks of a quiet room or listening to songs.

154.   **EDU_248**  很多 人 很 难 在 完全 安静 的 房间 工作 // **A lot of people have a hard time working in a completely quiet room**
**EDU_249**      或者 集中 注意力，// **or focus,**
**EDU_250**      但是 如果 听 // **But if you listen**
**EDU_251**      有 歌词 的 // **with lyrics**
**EDU_252**      音乐 ，// **music ,**
**EDU_253**      你 很 容易 // **you are easy**
**EDU_254**      因为 歌词 // **because of the lyrics**
**EDU_255**      而 分心 。// **And distracted.**
EDU_256      试试 听 白 噪音 或 纯 音乐 。// Try listening to white noise or pure music.
source: gcdt_whow_procrastination

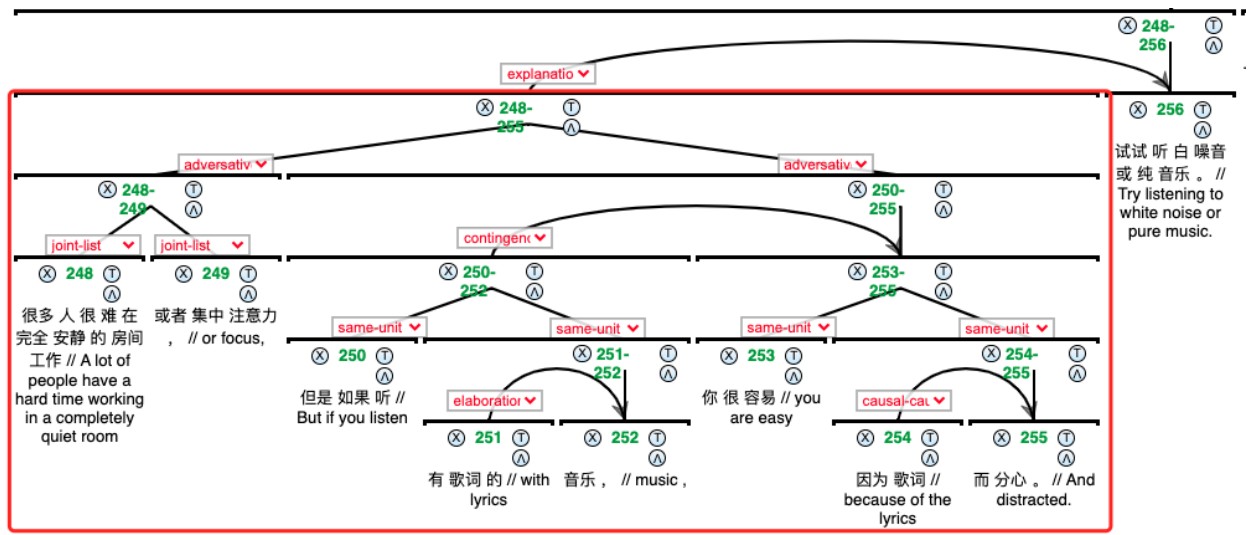

In contrast, *causal-cause* is annotated when the satellite is the direct cause of the nucleus. In the previous example, we see a *causal-cause* subtree DU_253-255 where lyrics cause distraction.

155.   EDU_253  你 很 容易 // you are easy
**EDU_254**      因为 歌词 // **because of the lyrics**
EDU_255      而 分心 。// And distracted.
source: gcdt_whow_procrastination

## 3.3 recursive question-answer pairs

Even though higher-level discourse structures have a relatively higher frequency of establishing multinuclear relations, in genres such as interviews, we observe evidence from follow-up questions that a hierarchical discourse structure is preferred over a flat one.

The following English example illustrates a chain of follow-up questions from food, to chickpea, to legume, recursively providing more detailed information to previously mentioned entities.

156.  EDU_1   What food do you like ?
      EDU_2   I like chickpeas.
      EDU_3   What is chickpea?
      EDU_4   Chickpea is an annual legume
      EDU_5   consumed in Mediterranean and Middle Eastern cuisines.
      EDU_6   What is a legume in Chinese?
      EDU_7   It is 荚果.

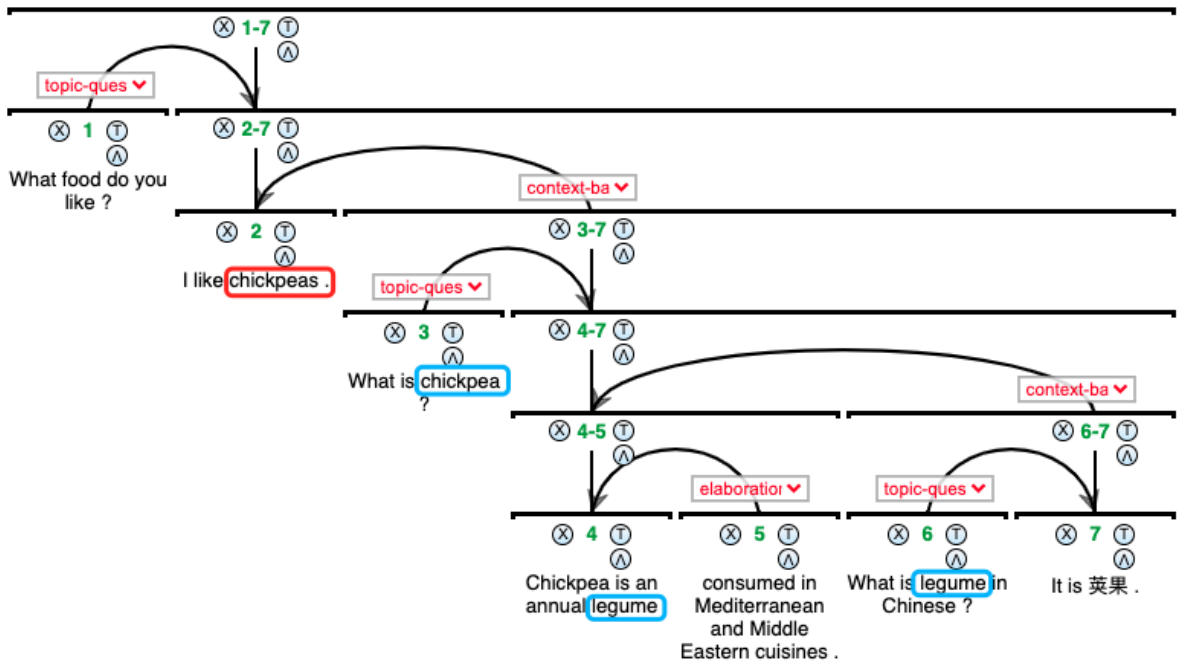

The following example is a shortened version of *gcdt_interview_game*, where we see branching out of question-answer pairs:
- game quality → cordon → process-hindering factors → human and non-human factors
- game system

157.  EDU_1      … 分别 就 比赛 品质 与 制度 两 主轴 ，探讨 比赛 与 音游 的 互动 。// … Discuss the interaction between the game and the music game in terms of the game quality and the system, respectively.

EDU_2      店家 维持 比赛 秩序 时 ，// When the store maintains the game order,

EDU_3      需要 注意 的 要点 ？// Points to pay attention to?

EDU_4      ... 应当 有 设置 警戒线 ... // ... There should be a cordon ...

EDU_5      假使 店家 的 比赛 空间 不利于 警戒线 的 架设 ，// If the store's competition space is not conducive to the erection of the cordon,

EDU_6      举办 比赛 的 店家 应当 如何 应变 ？// How should the store holding the competition respond?

EDU_7      ... 仍 会 有 一定 几率，妨碍 比赛 的 正常 进行 。// ... There will still be a certain chance that it will hinder the regular progress of the game.

EDU_8      妨碍 选手 正常 比赛 的 // obstructing the player's normal game

EDU_9      主要 人为 因素 ？// Major human factors?

EDU_10      .....

EDU_11      非人为 因素 ？// non-human factors?

EDU_12      ......

EDU_13      ... 赛制 ... // ... competition system ...

EDU_14      针对 这 部分 的 看法 ？// Opinions on this part?

EDU_15      ......

source: sampled EDUs from gcdt_interview_game

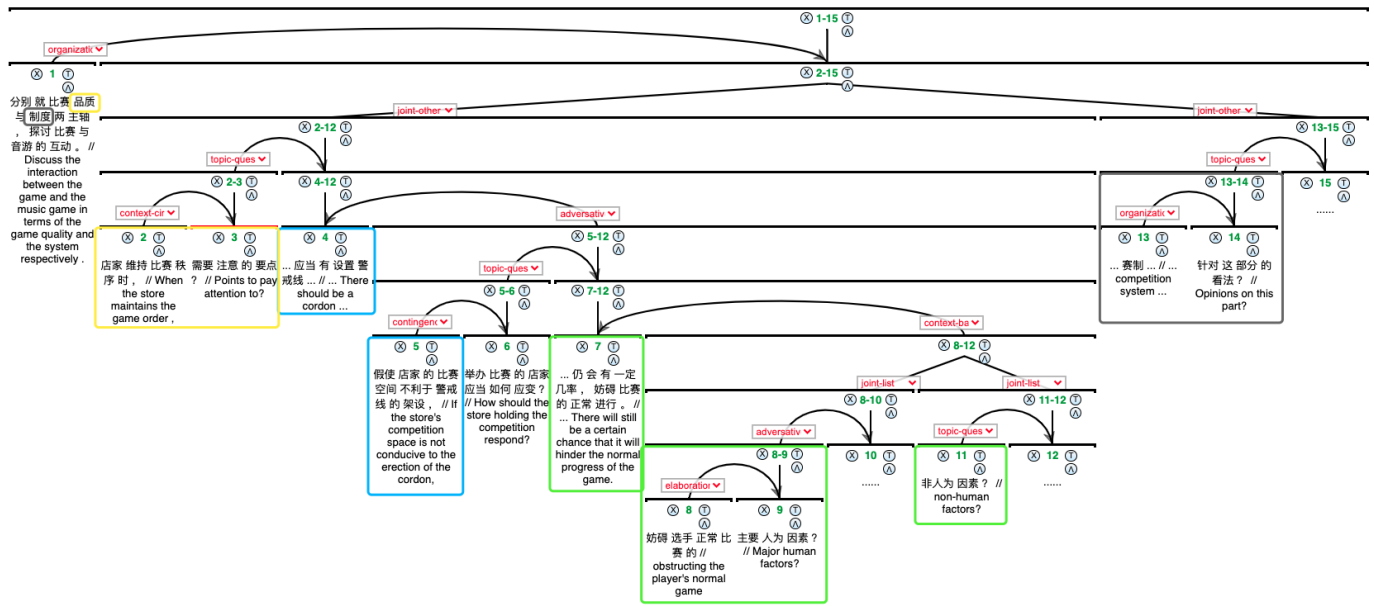

# References

Lynn Carlson and Daniel Marcu. 2001. Discourse tagging reference manual. ISI Technical Report ISI-TR545, 54(2001):56.

William C. Mann and Sandra A. Thompson. 1988. Rhetorical structure theory: Toward a functional theory of text organization. Text-Interdisciplinary Journal for the Study of Discourse, 8(3):243–281.

Mitchell P. Marcus, Beatrice Santorini, and Mary Ann Marcinkiewicz. 1993. Building a large annotated corpus of English: The Penn Treebank. Computational Linguistics, 19(2):313–330.

Fei Xia. 2000a. The Part-of-Speech Guidelines for the Penn Chinese Treebank (3.0).

Fei Xia. 2000b. The segmentation guidelines for the Penn Chinese Treebank (3.0).

Amir Zeldes. 2016. rstWeb - a browser-based annotation interface for Rhetorical Structure Theory and discourse relations. In Proceedings of the 2016 Conference of the North American Chapter of the Association for Computational Linguistics: Demonstrations, pages 1–5, San Diego, California. Association for Computational Linguistics.

Amir Zeldes. 2017. The GUM corpus: Creating multilayer resources in the classroom. Language Resources and Evaluation, 51(3):581–612.